\documentclass[10pt,twocolumn,letterpaper]{article}
\usepackage{times}
\usepackage{epsfig}
\usepackage{graphicx}
\usepackage{amsmath}
\usepackage{amssymb}
\usepackage[dvipsnames]{xcolor}

\usepackage{url}            
\usepackage{booktabs}       
\usepackage{rotating}
\usepackage{amsfonts}       
\usepackage{nicefrac}       
\usepackage{microtype}      
\usepackage{graphicx}
\usepackage{amsmath,amssymb} 
\usepackage{bbm}
\usepackage{array}
\usepackage{multirow}
\usepackage{bbding} 
\usepackage{paralist}
\usepackage{xspace}
\usepackage{subcaption}
\usepackage{xcolor} 
\usepackage{mathtools}
\usepackage{comment}
\usepackage{pbox}
\usepackage[numbers,sort,compress]{natbib}
\usepackage{makecell}

\usepackage[pagebackref=true,breaklinks=true,letterpaper=true,colorlinks,bookmarks=false]{hyperref}
\usepackage{cleveref}

\newcolumntype{H}[1]{>{\raggedright\let\newline\\\arraybackslash\hspace{0pt}}m{#1}}

\newcommand{\ra}[1]{\renewcommand{\arraystretch}{#1}}
\newcommand{\xhdr}[1]{\vspace{2pt}\noindent\textbf{#1}}

\newcommand{\Success}{\textsc{Success}\xspace}
\newcommand{\Progress}{\textsc{Progress}\xspace}
\newcommand{\SPL}{\textsc{SPL}\xspace}
\newcommand{\PPL}{\textsc{PPL}\xspace}

\newcommand{\aone}{$A_{O}$\xspace}
\newcommand{\atwo}{$A_{N}$\xspace}
\newcommand{\aones}{$A_{O}$'s\xspace}
\newcommand{\atwos}{$A_{N}$'s\xspace}

\newcommand{\symone}{$\Delta_1$\xspace}
\newcommand{\symtwo}{$\Delta_2$\xspace}
\newcommand{\symthree}{$\Delta_3$\xspace}
\newcommand{\symfour}{$\Delta_4$\xspace}
\newcommand{\symfive}{$\Delta_5$\xspace}
\newcommand{\symsix}{$\Delta_6$\xspace}

\newcommand{\found}{\textsc{found}\xspace}
\newcommand{\taskmulti}{multiON\xspace}
\newcommand{\task}{CoMON\xspace}

\newcommand{\mon}[1]{$#1$-ON}

\newcommand{\ContCom}{\texttt{U-Comm}\xspace}
\newcommand{\DiscCom}{\texttt{S-Comm}\xspace}
\newcommand{\RandContCom}{\texttt{Rand U-Comm}\xspace}
\newcommand{\RandDiscCom}{\texttt{Rand S-Comm}\xspace}

\newcommand{\NoMap}{\texttt{NoCom}\xspace}
\newcommand{\OracleMap}{\texttt{OracleMap}\xspace}

\newcommand{\secref}[1]{Sec\onedot~\ref{#1}}

\definecolor{deltaone}{RGB}{28, 81, 255}
\definecolor{deltatwo}{RGB}{255,127,80}
\definecolor{deltathree}{RGB}{33, 173, 65}

\newcommand{\compresslist}{%
\setlength{\itemsep}{1pt}%
\setlength{\parskip}{0pt}%
\setlength{\parsep}{0pt}%
}

\usepackage{iccv}
\usepackage{times}
\usepackage{epsfig}
\usepackage{graphicx}
\usepackage{amsmath}
\usepackage{amssymb}
\usepackage[numbers,sort,compress]{natbib}
\usepackage[normalem]{ulem}
\usepackage[accsupp]{axessibility} 

\usepackage[pagebackref=true,breaklinks=true,letterpaper=true,colorlinks,bookmarks=false]{hyperref}

\iccvfinalcopy 

\def\iccvPaperID{7990} 

\ificcvfinal\fi

\begin{document}

\title{Interpretation of Emergent Communication in\\Heterogeneous Collaborative Embodied Agents}

\author{Shivansh Patel$^{1}$\thanks{denotes equal contribution by SP, SW, and UJ}
\hspace{3em} Saim Wani$^{2}$\footnotemark[1]
\hspace{3em} Unnat Jain$^{3}$\footnotemark[1]
\hspace{3em} Alexander Schwing$^{3}$\\
Svetlana Lazebnik$^{3}$
\qquad Manolis Savva$^{1}$
\qquad Angel X. Chang$^{1}$
\\[2pt]
\small{$^{1}$Simon Fraser University \qquad $^{2}$IIT Kanpur \qquad $^{3}$UIUC}\\[2pt]
\small{\url{https://shivanshpatel35.github.io/comon}}
}

\maketitle
\ificcvfinal\thispagestyle{empty}\fi

\begin{abstract}

Communication between embodied AI agents has received increasing attention in recent years.
Despite its use, it is still unclear whether the learned communication is interpretable and grounded in perception.
To study the grounding of emergent forms of communication, we first introduce the collaborative multi-object navigation task `CoMON.'
In this task, an `oracle agent' has detailed environment information in the form of a map.
It communicates with a `navigator agent' that perceives the environment visually and is tasked to find a sequence of goals.
To succeed at the task, effective communication is essential.
CoMON hence serves as a basis to study different communication mechanisms between heterogeneous agents, that is, agents with different capabilities and roles.
We study two common communication mechanisms and analyze their communication patterns through an egocentric and spatial lens.
We show that the emergent communication can be grounded to the agent observations and the spatial structure of the 3D environment.
Video summary: \url{https://youtu.be/kLv2rxO9t0g}
\end{abstract}
\section{Introduction}
\label{sec:introduction}

Research in embodied AI agents that learn to perceive, act, and communicate within 3D environments has become popular in recent years~\cite{anderson2018evaluation,batra2020rearrangement,bisk2020experience}.
Collaboration between multiple agents has also received an increasing amount of attention.
Consequently, there has been renewed interest in studying communication mechanisms that increase the effectiveness of collaborative agents~\cite{lazaridou2020emergent}.

\begin{figure}
\includegraphics[width=\linewidth]{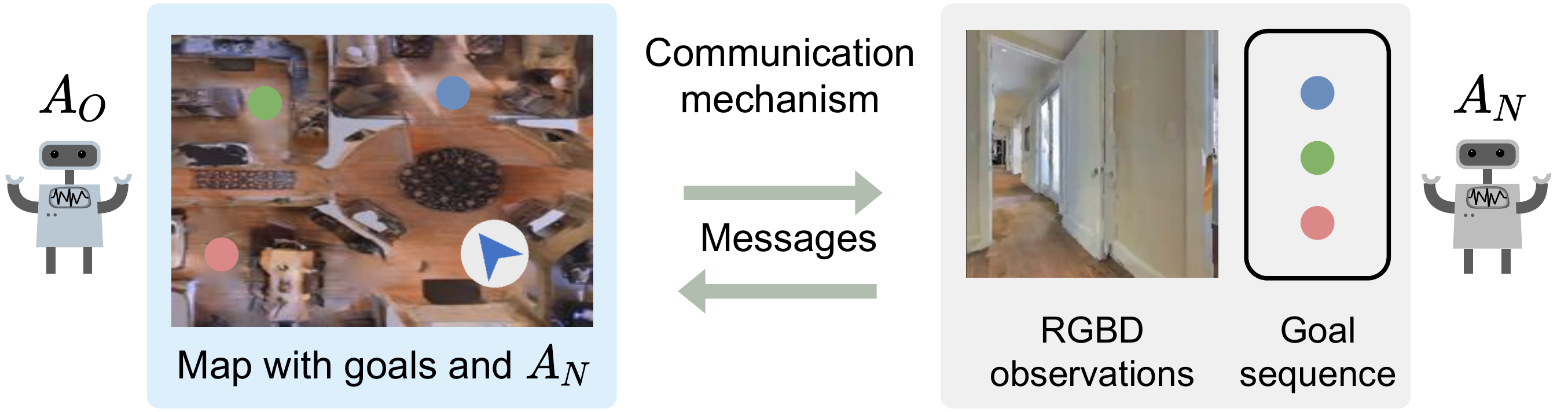}
\caption{
We propose a collaborative multi-object navigation task (\task) where an oracle agent \aone communicates with a navigator agent \atwo.
The oracle \aone possesses a global map and the navigator \atwo needs to perceive and navigate a 3D environment to find a sequence of goal objects while avoiding collisions.
Through this task, we study the impact of structured and unstructured communication mechanisms on navigation performance, and the emergence of messages grounded in egocentric perception.
}
\label{fig:task}
\end{figure}

A key goal of communication is to transmit information.
Therefore, to analyze 
communication it is common to study collaborative tasks where agents have heterogeneous abilities or asymmetric access to information~\cite{kottur2017natural,bogin2018emergence}.
A heterogeneous agent setup also corresponds to real-world scenarios such as guiding a delivery driver to our home over a phone call.
However, prior work on emergent communication has adopted simplified settings like reference games~\cite{batali1998computational,lazaridou2017multi} or agents communicating within 2D environments~\cite{bogin2018emergence}.
Work involving communication in 3D environments has focused on whether communication can lead to improved performance through cooperation in solving the task  \cite{jain2019CVPRTBONE,iqbal2019coordinated,JainWeihs2020CordialSync}, rather than detailed interpretation of the emergent communication patterns.
Despite this rich literature studying emergent communication, there has been no systematic analysis and interpretation of emergent communication in realistic 3D environments.

In this paper, we focus on interpreting emergent communication through an \emph{egocentric} and \emph{spatially grounded} analysis.
To do this, we define a collaborative multi-object navigation task (\task), which extends the recently proposed multi-object navigation (MultiON) task~\cite{wani2020multion}.
The \task task requires a pair of agents---an oracle with privileged knowledge of the environment in the form of a map, and a navigator who can perceive and navigate the environment---to communicate with each other in order for the navigator to find and reach a sequence of goal objects (see \Cref{fig:task}).
The primary role of this task is to study the emergent communication between heterogeneous agents in visually realistic 3D environments.

We conduct a rigorous comparison and interpretation of two families of communication mechanisms: \emph{unstructured} and \emph{structured} communication.
The first is commonly adopted in non-visual RL settings~\cite{FoersterNIPS2016,liu2020who2com} and corresponds to the `continuous communication' of DIAL by \citet{FoersterNIPS2016}.
The latter introduces an inductive bias by imposing a discrete message structure and has been adopted by the Embodied AI community~\cite{jain2019CVPRTBONE,JainWeihs2020CordialSync}.

We find that: 1) structured communication can achieve higher navigation performance than unstructured communication; 2) agents using structured communication come close to matching the success rate of `oracle' communication but are less efficient; 3) interpretable messages akin to `I am looking for the red goal' emerge in both communication mechanisms; and 4) both communication mechanisms lead to the emergence of egocentrically-grounded messages such as `goal is close in front of you,' and `goal is behind you.'

\section{Related work}
\label{sec:rel}

Our work is related to cooperation and coordination between multiple agents~\cite{busoniu2008comprehensive,foerster2018counterfactual,gupta2017cooperative,lauer2000algorithm,lowe2017multi,matignon2007hysteretic,omidshafiei2017deep,panait2005cooperative,foerster2017learning,liu2020pic,liu2021cooperative,weihs2021learning}. 
We discuss relevant work in emergent communication, collaborative embodied AI, and embodied navigation tasks.

\xhdr{Emergent communication.}
Work on understanding the emergence of communication through simulations has a long history.
\citet{batali1998computational} studied this by encoding simple phrases into a series of characters that need to be decoded by another agent.
\citet{STEELS2003308} studied a similar experiment with robots that had to generate a shared lexicon to perform well in a guessing game.
\citet{foerster2016riddles} showed that RL agents can learn successful communication protocols.
\citet{FoersterNIPS2016} then showed that agents can learn communication protocols in the form of messages that are sent to each other.
When the agents are allowed to communicate, interesting communication patterns emerge~\cite{bratman2010new,giles2002learning,kasai2008learning,lazaridou2017multi,melo2011querypomdp,mordatch2018emergence,sukhbaatar2016learning,bullard2020exploring,gupta2020compositionality,lowe2020interaction}.
\citet{das2017learning} propose a cooperative image guessing game between two static heterogeneous agents where the agents communicate through natural language.
\citet{mordatch2018emergence} investigate the emergence of grounded compositional language in multi-agent populations.  For a survey of emergent communication methods we refer the reader to \citet{lazaridou2020emergent}.

Our work is similar in spirit to \citet{kottur2017natural}, in that we study and analyze emergent communication patterns and what information they communicate.
Unlike that work and other work in emergent communication, we are less interested in whether compositional language emerges when using discrete symbols, but rather on whether there is consistent interpretation of messages between the two agents, and whether they correspond to available visual information.
\citet{kajic2020learning} study how agents develop interpretable communication mechanisms in grid-world navigation environments and visualize agent policies conditioned on messages.
We have a similar focus but we study continuous communication in realistic 3D environments.

\xhdr{Collaborative embodied AI tasks.}
While single agent embodied tasks have been studied in depth, there is less work on collaborative embodied agents.
\citet{das2018tarmac} develop a targeted multi-agent communication architecture where agents select which of the other agents to communicate with.
\citet{jain2019CVPRTBONE} introduce a furniture-lifting task where two agents must navigate to a furniture item.
These agents must coordinate to satisfy spatial constraints for lifting the heavy furniture.
Followup work studies a furniture-moving task where the agents relocate lifted furniture items~\cite{JainWeihs2020CordialSync,jain2021gridtopix}.
However, the agents are homogeneous and no map representation is studied in these prior works.
\citet{iqbal2019coordinated} study coordinated exploration by introducing handcrafted intrinsic rewards to incentivize agents to explore `novel' states.
Here, agents do not explicitly communicate with each other.
Our work is focused on studying a spectrum of communication mechanisms for heterogeneous agents in visually realistic indoor 3D environments.

\xhdr{Navigation tasks in Embodied AI.}
Agents capable of navigating in complex, visual, 3D environments~\cite{ammirato2017dataset,BrodeurARXIV2017,chang2017matterport,ai2thor,stanford2d3d,xia2018gibson,xia2019interactive,kempka2016vizdoom,dosovitskiy2017carla,chen2020soundspaces,AllenAct} have been studied extensively. 
\citet{anderson2018evaluation} divide embodied navigation tasks into point goal navigation (PointNav), object goal navigation (ObjectNav) and room goal navigation (RoomNav).
Pertinent to this work, ObjectNav agents are given goal cues such as an object category label or an image of the goal object~\cite{ZhuARXIV2016,yang2018visual,batra2020objectnav,chaplot2020object,chaplot2020neural,chang2020semantic,ye2021auxiliary}.
Long-horizon navigation tasks are most relevant to our work~\cite{multitargeteqa,fang2019scene,beeching2020egomap,weihs2021visual,szot2021habitat}.
Map-based navigation methods have been benchmarked on multi-object navigation (\taskmulti) \ie navigating to an ordered sequence of goal objects~\cite{wani2020multion}.
Since we study communication involving map-based memory, we extend \taskmulti to a collaborative setting.

\section{Task}
\label{sec:task}

Here, we describe the collaborative \taskmulti (\task) task, the agent observation and action spaces, and discuss alternatives to sharing information between the agents.

\xhdr{Background task (\taskmulti).}
In an episode of \taskmulti~\cite{wani2020multion}, an agent must navigate to an ordered sequence of goal objects $G$ placed within an environment.
The agent indicates discovery of a goal object by executing a \found action within a threshold distance from the goal.
The objects in $G$ are sampled from a set of $k$ unique categories.
An episode is a failure if the agent calls \found while not in the vicinity of the current goal, or if the allocated time budget is exceeded.
We use \mon{m} to denote an episode with $m$ sequential goals.

\xhdr{\task task.}
In Collaborative MultiON (\task), an episode involves two heterogeneous agents \aone and \atwo. \aone is a disembodied oracle, which cannot navigate in the environment. 
However, \aone has access to oracle information (detailed later) of the environment's state.
\atwo is an embodied navigator, which navigates and interacts with the environment. \atwo carries out a \taskmulti~\cite{wani2020multion} task.
To optimize the team's (shared) rewards, both agents must collaborate. 
For this, 
\aone and \atwo perform the task collaboratively by communicating via a limited-bandwidth channel.

\begin{figure*}
\includegraphics[width=\textwidth]{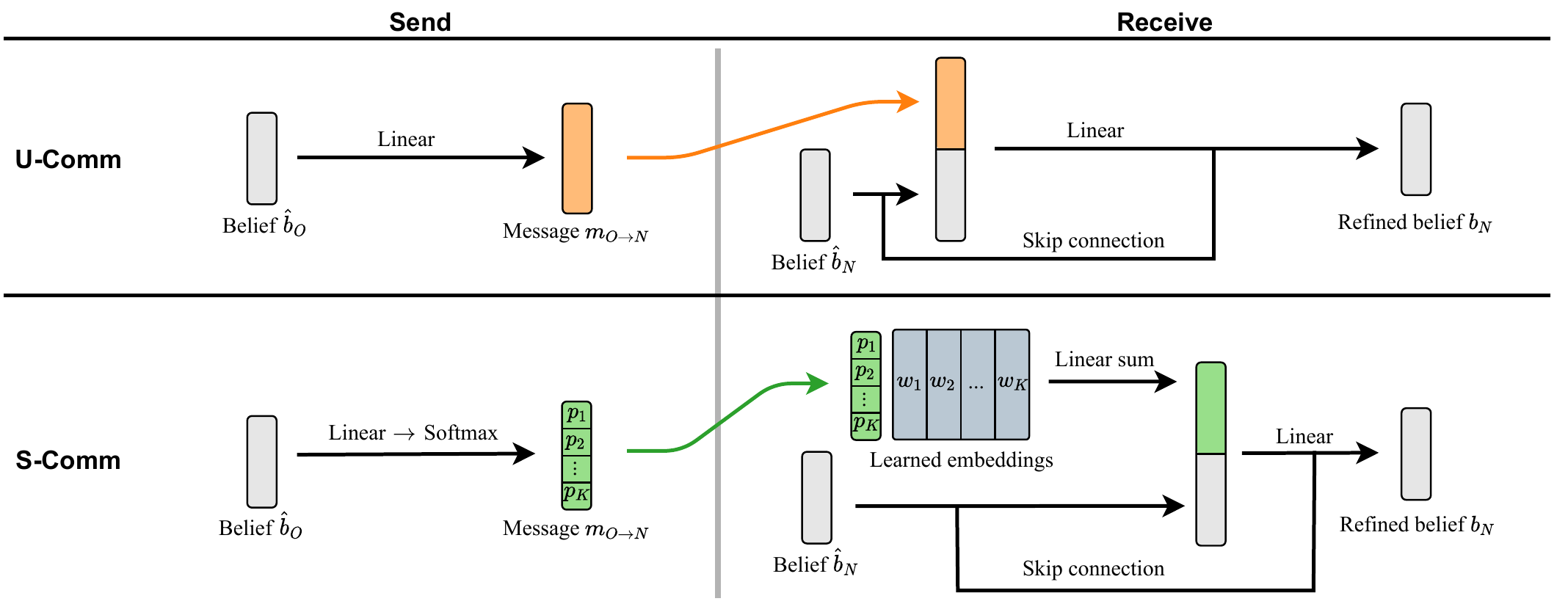}
\caption{
\textbf{Architecture of the send and receive branches for the unstructured (\ContCom) and structured (\DiscCom) communication mechanisms.}
On the sending branch, the agent creates a message by passing through a linear layer for \ContCom and by passing through a linear layer and a softmax layer for \DiscCom.
On the receiving branch for \ContCom, the message is concatenated with the belief and passed through a linear layer and skip connected to obtain the refined belief.
For \DiscCom, the message is first \textit{decoded} by linearly combining the word embeddings $w_k$ while using the probabilities $p_k$ as weights ($\sum_{k=1}^Kp_kw_k$).
The embeddings are learned for each agent and round.
}
\label{fig:comm_architecture}
\end{figure*}

\xhdr{Agent observations.}
\aone has access to a fixed top-down view of the scene along with \atwo's position and orientation.
The scene is discretized and represented as an oracle map $M$, a 3D tensor.
The first two dimensions correspond to the horizontal and vertical axes of the top-down view, and the third contains semantic information in each cell $M[i,j]$:
\begin{itemize}\compresslist
    \item \emph{Occupancy}: whether location $[i,j]$ is free space (\ie, navigable), occupied, or out of the scene bounds.
    \item \emph{Goal objects}: categorical variable denoting which goal object is located at $[i,j]$ or a `no object' indicator.
\end{itemize}

The observations of \atwo are consistent with \taskmulti~\cite{wani2020multion}, allowing architectures trained on the single-agent task to be used in our collaborative setting.
At time-step $t$, the observations of \atwo include:
\begin{itemize}\compresslist
    \item \emph{RGBD}: egocentric vision and depth frame $o_t$.
    \item \emph{Object}: categorical variable denoting the current goal object as one-hot vector $g_t$.
    \item \emph{Previous action}: agent action at previous time step as one-hot vector $a_{t-1}$.
\end{itemize}

\xhdr{Agent action space.}
At each time step, both \aone and \atwo send messages to each other.
\atwo additionally takes an environment action following the communication round.
The action space consists of four actions: \{ \textsc{Forward}, \textsc{Turn Left}, \textsc{Turn Right}, and \found \}. \textsc{Forward} takes the agent forward by $0.25$m and turns are $30^{\circ}$ each.

\xhdr{Task design alternatives.}
We note that there are other choices for how to distribute information between \aone and \atwo.
For example, the goal sequence information could be given to \aone.
This would correspond to the practical scenario of a dispatch operator communicating with a taxi driver.
However, this would lead to most information being concentrated with \aone and obviate the need for frequent two-way communication between \aone and \atwo.
Yet another setting would  hide \atwo's position and orientation on the map from \aone.
Our preliminary investigations included experiments in this setting, with no information about \atwo's position on the map being given to \aone.
We empirically observed that this was a hard learning problem, with the agents failing to acquire meaningful task performance or communication strategies.
We hypothesize that this may be partly due to a strong coupling with the independently challenging localization problem (\ie, determining \atwo's position and orientation in the map through egocentric observations from \atwo's perspective).
Since there is a rich literature for localization based on egocentric visual data (\eg, see \citet{fuentes2015visual} for a survey), we factor out this aspect allowing a deeper focus on interpretation of emergent communication.

\section{Agent models}
\label{sec:approach}

We provide an overview of our agent models by describing the communication mechanisms, the agent network architectures, the reward structure and implementation details.

\subsection{Communication mechanisms}
\label{sec:comm_mech}

We study two types of communication mechanisms: unstructured~\cite{FoersterNIPS2016,liu2020who2com} and structured~\cite{jain2019CVPRTBONE,JainWeihs2020CordialSync}.
Their key difference is that the unstructured mechanism implements free-form communication via a real-valued vector, whereas the structured communication mechanism has an inductive bias through the imposed message structure. 
\Cref{fig:comm_architecture} illustrates these two types of communication.
Each round of communication involves the two agents synchronously sending a message to each other.
The receiving agent uses the message to refine its internal representation (\ie, belief).
The same architecture is used for both agents and for each communication round.

\xhdr{Unstructured communication (\ContCom).}
The agent communicates a real-valued vector message.
For sending the message, the belief is passed through a linear layer to produce the sent message.
On the receiving side, the received message is concatenated with the belief and passed through two fully connected layers and skip connected through the belief to obtain the refined belief.

\xhdr{Structured communication (\DiscCom).}
The agent has a vocabulary of $K$ words $w_1,\dots,w_K$, implemented as learnable embeddings.
Note that the embeddings for the two rounds, and the two agents differ and are separately learned.
The sent message is a set of probabilities $p_1,\dots,p_K$ (where $\sum_{l=1}^{K} p_l = 1$) corresponding to the $K$ words.
These probabilities are obtained by passing the belief through a linear layer followed by a softmax layer.
On the receiving side, the agent decodes these incoming message probabilities by linearly combining its word embeddings using the probabilities as weights, \ie, it computes $\sum_{l=1}^{K} p_l w_l $.
Similar to the previous mechanism, this decoded message is concatenated with the belief and passed through two fully connected layers and skip connected to obtain the refined belief.
In early experiments, we tried using discrete tokens instead of a weighted sum.
To make the model differentiable, we used the Gumbel-Softmax trick but found the agents could not be trained successfully.
We hypothesize this is due to the high-dimensional input space and the numerical instability of Gumbel-Softmax~\cite{potapczynski2019invertible}.

\begin{figure*}
\centering
\includegraphics[width=\textwidth]{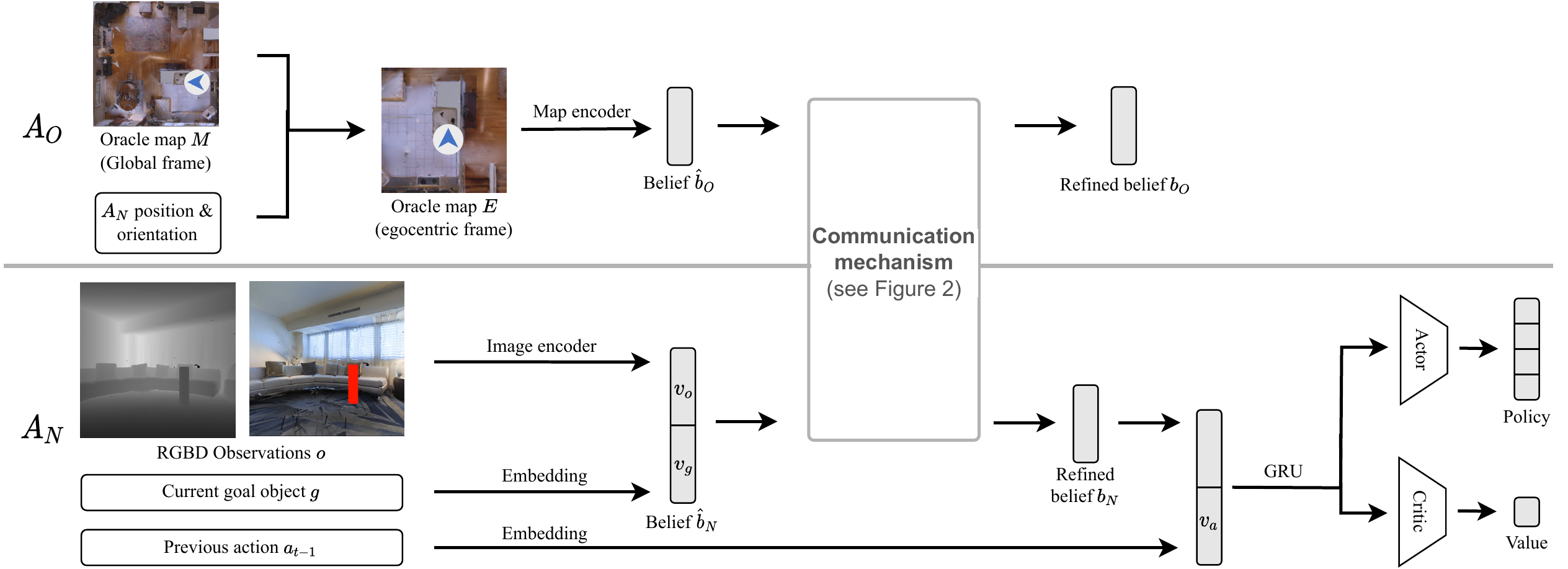}
\caption{
\textbf{Overall agent model architecture.}
\aone and \atwo process their respective inputs to get initial beliefs $\hat{b}_O$ and $\hat{b}_N$ which encode the agent's \textit{belief} about the current observation.
These are refined by a communication channel into final beliefs $b_O$ and $b_N$.
The belief $b_N$ is concatenated with the previous action, and passed through a GRU to actor and critic heads to obtain policy and value function estimates.
}
\label{fig:architecture}
\end{figure*}

\subsection{Agent network architecture}
\Cref{fig:architecture} illustrates the network architecture.
We adapt the TBONE architecture which has been shown to be successful for multi-agent embodied tasks~\cite{jain2019CVPRTBONE,JainWeihs2020CordialSync}.
For readability we drop the subscript $t$ denoting the time step.
\aone encodes the map by storing two 16-dimensional learnable embeddings for the occupancy and goal object category information at each grid location.
Since \aone has access to \atwo's position and orientation, it programmatically crops and rotates the map $M$ around \atwo's position and orientation to build an egocentric map $E$.
This implicitly encodes \atwo's position and orientation into $E$ which is then passed through a CNN and a linear layer to obtain \aone's initial belief $\hat{b}_O$.

\atwo passes its RGBD observations $o$ through a CNN and a linear layer to obtain an observation embedding $v_o$.
It also passes the object category $g$ and previous action $a_{t-1}$ through separate embedding layers to obtain a real-valued goal embedding $v_g$ and action embedding $v_a$ respectively.
$v_o$ and $v_g$ are concatenated to obtain \atwo's initial belief $\hat{b}_N$.

Both \aone and \atwo go through two rounds of communication (as detailed in~\Cref{sec:comm_mech}) to obtain their final beliefs $b_O$ and $b_N$ respectively. \atwo concatenates its final belief $b_N$ with the previous action embedding $v_a$ and passes it through a GRU to obtain a state vector $s$.
Following \citet{jain2019CVPRTBONE,JainWeihs2020CordialSync}, we use an actor-critic architecture where the state vector $s$ is passed through: i) an actor head to estimate the distribution over the action space; and ii) a critic head that outputs a value estimating the utility of the state.
$b_O$ is left unused and hence it is discarded.

\subsection{Reward structure}

We model our multi-agent setup using the centralized training and decentralized execution paradigm~\cite{lowe2017multi,foerster2018counterfactual,sunehag2017value,mahajan2019maven,son2019qtran}.
In this paradigm, a central critic estimates the value function $V(s)$ of all the agents.
Execution during testing is decentralized and each agent takes independent actions.
The agents are trained using the  navigator (\atwo) reward:
$r_t = \mathbbm{1}^{[\text{reached subgoal}]}_tr^{\text{goal}} + r^{\text{closer}}_t + r^{\text{time penalty}}$
where $\mathbbm{1}^{[\text{reached subgoal}]}_t$ is a binary indicator of finding a goal at time step $t$, $r^\text{goal}$ is the reward for finding a goal, $r^\text{closer}_t$ is the decrease in geodesic distance to the goal between the previous and the current time step, and $r^{\text{time penalty}}$ is the penalty per time step.

\subsection{Implementation details}

Following \citet{wani2020multion}, we set $r^\text{goal}$ to $3$ and $r^{\text{time penalty}}$ to $-0.01$.
We train with PPO~\cite{schulman2017proximal}, using 16 parallel threads with 4 mini-batches and 2 epochs per PPO update.
Agents are trained until 50M steps accumulate across worker threads.
The map $M$ is of dimension $300 \times 300$ and each cell corresponds to a $0.8m \times 0.8m$ patch on the ground.
See the supplement for more details.

\section{Experiments}
\label{sec:experiments}

Here, we describe the experimental setup we adopt to study both communication mechanisms.

\subsection{Agent models}

All  agent models share the  base architecture  explained in~\secref{sec:approach}. For ablations each model is adjusted as follows (see supplement for details):

\xhdr{\NoMap}~\cite{wani2020multion} is the model without agent \aone.
This represents the case where navigator \atwo can't receive help from an oracle. It hence represents the `no communication' scenario.

\xhdr{\RandContCom} represents a model using unstructured communication while the messages sent between the agents are Gaussian random vectors.
This provides a lower bound for unstructured communication.

\xhdr{\RandDiscCom} represents a model using structured communication while the messages sent between the agents are  random multinomial probability vectors.
This provides the lower bound for structured communication.

\xhdr{\ContCom} represents a model using unstructured communication  as explained in~\secref{sec:comm_mech}.

\xhdr{\DiscCom} represents a model using structured communication  as explained in~\secref{sec:comm_mech}.

\xhdr{\OracleMap}~\cite{wani2020multion} combines both \aone and \atwo into a single agent. 
Effectively, this agent has access to the map and it has to navigate in the environment without a need for communication. 
Hence, it sets an upper bound for performance.

\subsection{Datasets}

We use the \taskmulti dataset~\cite{wani2020multion} based on the AI Habitat simulator~\cite{habitat19iccv}.
This dataset contains episodes with agent starting position, orientation, and goal locations.
There are eight goal objects with identical cylindrical shape but different colors.
The episodes are generated from Matterport3D~\cite{chang2017matterport} scenes.
We follow the standard scene-based Matterport3D train/val/test split with episodes established by~\citet{wani2020multion}.
Each scene contains 50,000 episodes for the train split and 12,500 episodes for the val and test splits.
We train models for \mon{3} (3 sequential goals) and evaluate on \mon{1}, \mon{2}, \mon{3}, \mon{4} and \mon{5}.

\begin{table}
\ra{1.3}
\centering
\resizebox{\linewidth}{!}{
\begin{tabular}{@{}l rrr rrr@{}}
\toprule
 & \multicolumn{3}{c}{\Progress (\%)} & \multicolumn{3}{c}{\PPL (\%)}\\ \cmidrule(lr){2-4} \cmidrule(lr@{0mm}){5-7}
 & \mon{1} & \mon{2} & \mon{3} & \mon{1} & \mon{2} & \mon{3}\\
\midrule
 \NoMap & 56  & 39 & 26 & 35  & 26  & 16 \\
 \RandContCom  & 59 & 40 & 28 & 36 & 28 & 18 \\
 \RandDiscCom & 50 & 31 & 24 & 33 & 24 & 16 \\
 \ContCom & \textbf{87} & 77 & 63 & 60 & 51 & 39 \\
 \DiscCom & 85 & \textbf{80} & \textbf{70} & \textbf{67} & \textbf{59} & \textbf{50} \\
\midrule
 \OracleMap & 89 & 80 & 70 & 74 & 64 & 52 \\
\bottomrule
\end{tabular}
}
\vspace{1pt}
\caption{\textbf{Task performance metrics for different communication mechanisms evaluated on the} \mon{1}, \mon{2} \textbf{and} \mon{3} \textbf{tasks.}
\RandDiscCom and \DiscCom have a vocabulary size of two.
For a fair comparison, both \RandContCom and \ContCom have the same message length of two elements.
The random baselines perform poorly, and are close to the \NoMap (\ie `no communication') baseline.
Both the \ContCom and \DiscCom communication mechanisms perform much better and approach \OracleMap, with \DiscCom being mostly more successful (higher \Progress) and more efficient (higher \PPL), especially as the task becomes more challenging.
Variance in \Progress for all models in \mon{3} is less than 2\%.
}
\label{tab:results}
\end{table}

\subsection{Quantitative evaluation}

We adopt the four metrics used in \citet{wani2020multion}.
\textbf{\Success}: episode success rate;
\textbf{\Progress}: fraction of goal objects found in an episode;
\textbf{\SPL}: success weighted by path length;
\textbf{\PPL}: progress weighted by path length.

We summarize our experimental findings in \Cref{tab:results}.
We report \Progress and \PPL for 1,000 val episodes.
As expected, \OracleMap has the highest performance among all the agent models, with significant gains over \NoMap.
\RandContCom and \RandDiscCom perform close to \NoMap which shows that the learnt messages indeed contain useful information.
We observe that \DiscCom performs better than \ContCom.
The difference is more pronounced as the task difficulty increases.
\PPL decreases by 10.44\% for \mon{1}, 13.5\% for \mon{2}, and 22\% for \mon{3}.
This shows that the imposed communication structure helps learn more efficient communication strategies.
\NoMap and \OracleMap are the same as in \citet{wani2020multion} but we train for 50M steps instead of 40M steps.
To test  generalization, we also evaluate on \mon{4} (\DiscCom \Progress is 63\% \vs \ContCom 41\%) and \mon{5} (\DiscCom \Progress is 52\% \vs \ContCom 26\%).
This indicates that \DiscCom agents are better able to generalize to harder tasks (see supplement for more details).

\begin{figure}
\includegraphics[width=\linewidth]{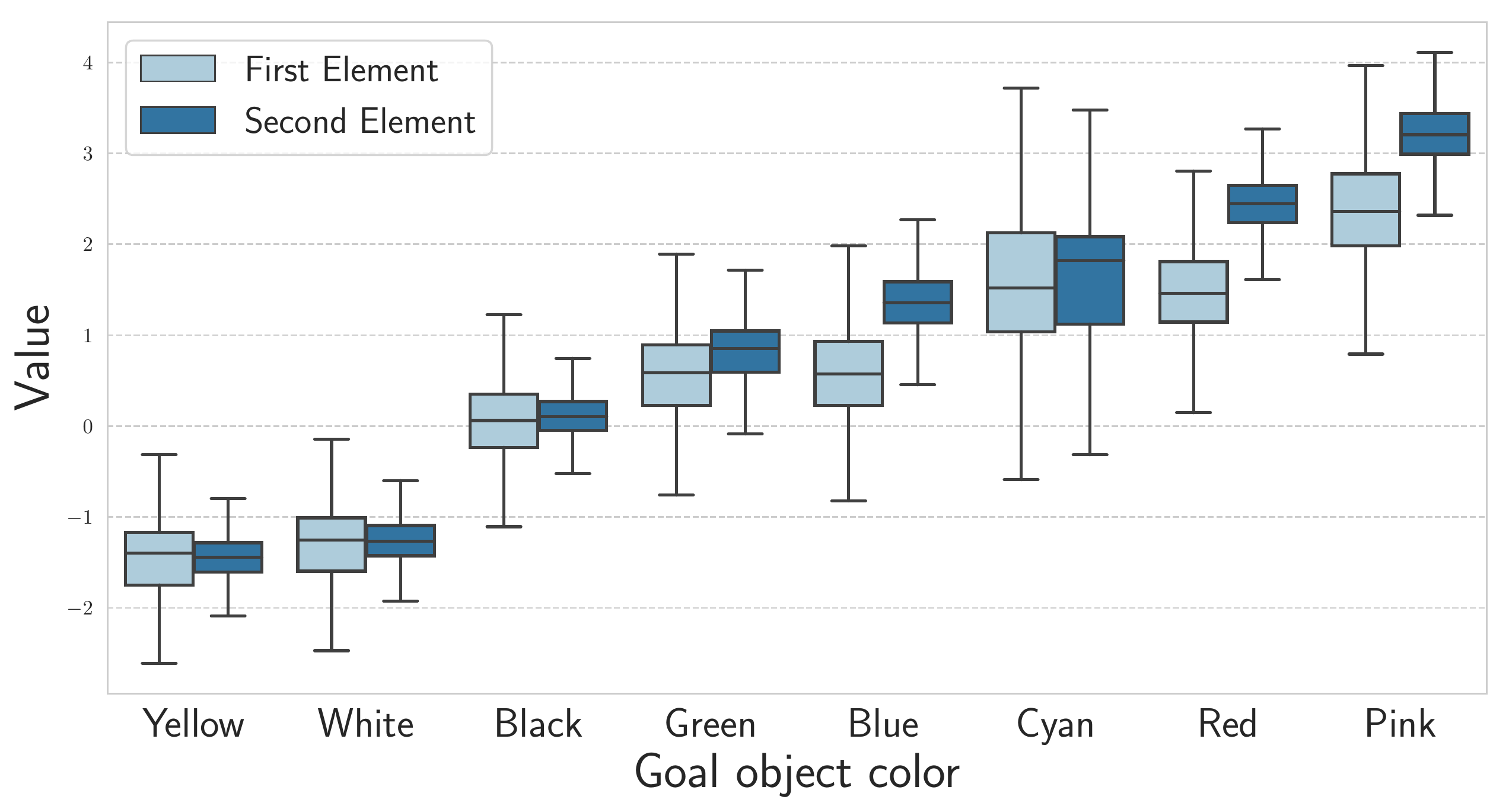}
\caption{\textbf{Value of first and second element of} $m_{N \rightarrow O}^1$\textbf{ message plotted against goal object color in \ContCom.}
Goal object colors are on the x-axis and the distribution of $m_{N \rightarrow O}^1$ values is on the y-axis.
The box plots show $0^\text{th}$, $25^\text{th}$, $50^\text{th}$, $75^\text{th}$, and $100^\text{th}$ quartiles after removing outliers.
Note that \atwo sends different messages for differently colored objects.
The ordering of the colors by average message value appears to respect color hue similarity (\eg, red and pink are close together and far from yellow and white).
}
\label{fig:cont_comm_len2_m_N_to_O}
\end{figure}

\begin{figure*}
\resizebox{\linewidth}{!}{
\newcolumntype{C}{>{\centering\arraybackslash} m{8.8cm} }
\begin{tabular}{@{}CC@{}}
{$1^{st}$ element} & {$2^{nd}$ element} \\
    \includegraphics[width=1\linewidth]{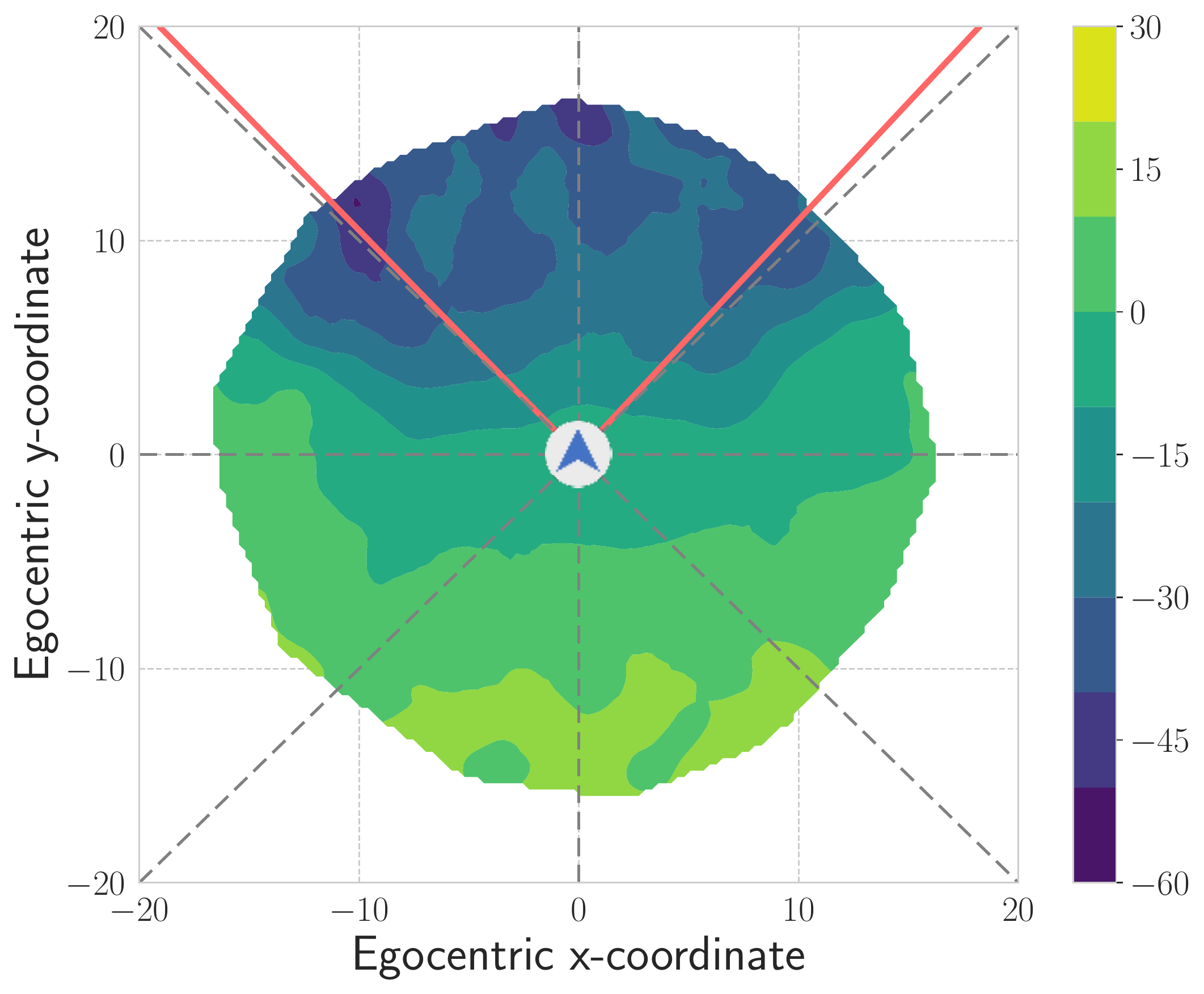}
    &
    \includegraphics[width=1\linewidth]{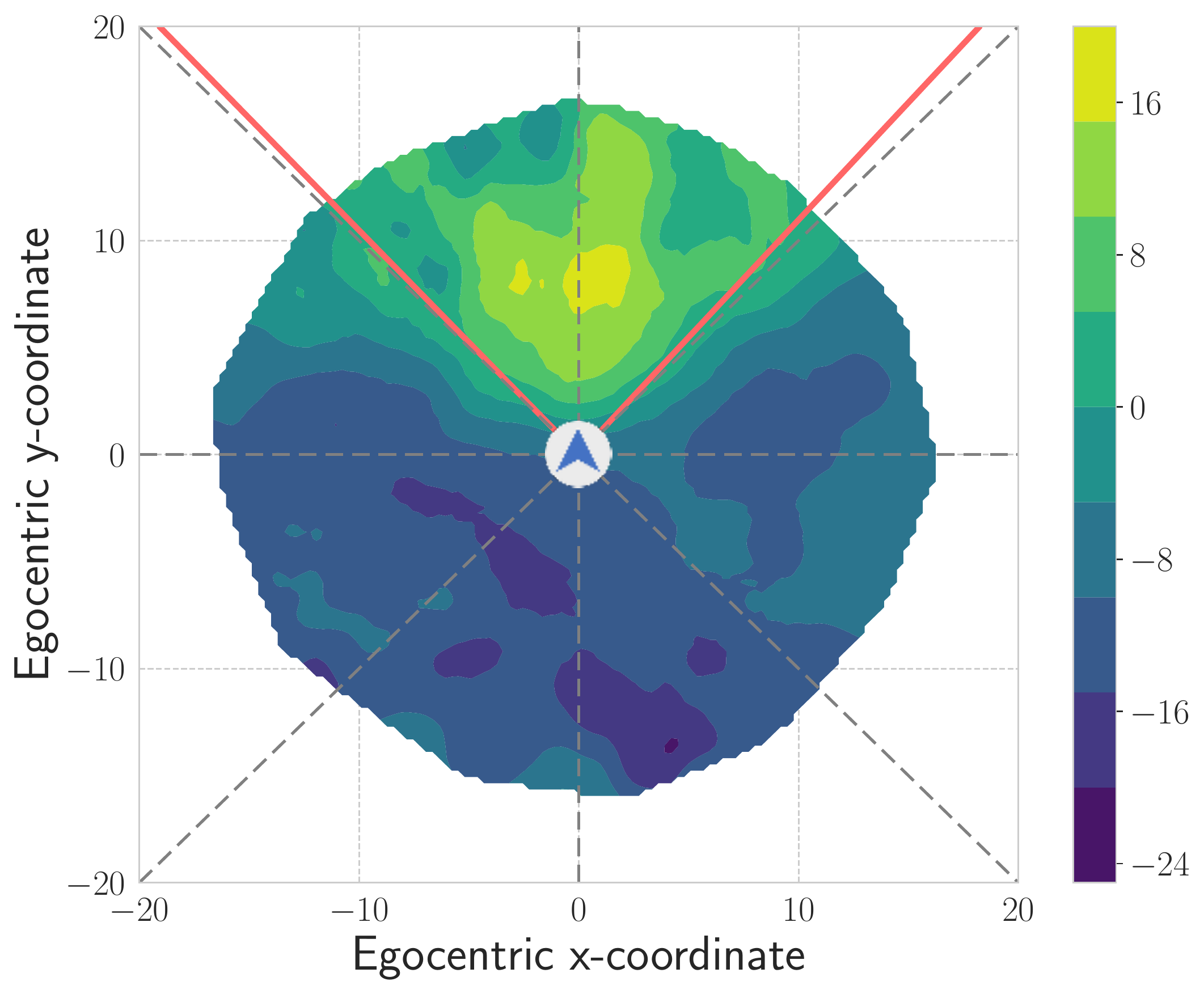}
\end{tabular}
}
\caption{
\textbf{Egocentric visualization of \ContCom communication symbol} $m_{O \rightarrow N}^2$.
The two plots visualize the value of the first and second element of the message plotted \wrt the relative coordinates of the goal object from \atwo.
The navigator agent \atwo is facing the +y axis and its field-of-view is marked with red lines.
The plot on the left corresponds to the $1^\text{st}$ dimension of the message, while the plot on the right corresponds to the $2^\text{nd}$ dimension.
The value of each dimension is indicated by the color hue.
We observe that higher values of the $1^\text{st}$ dimension correspond to `farther behind', while higher values of the $2^\text{nd}$ dimension are clustered `close and in front' of the agent.
}
\label{fig:cont_comm_len2_m_O_to_N}
\end{figure*}

\section{Communication analysis}
\label{sec:comm_analysis}

Here, we interpret the emergent communication between the agents.
We use the notation $m_{\text{sender} \rightarrow \text{receiver}}^{\text{round}}$.
Hence $m_{O \rightarrow N}^1$ denotes the message sent by \aone to \atwo for round one.
At each step, four messages are sent between the agents: $m_{O \rightarrow N}^1$, $m_{N \rightarrow O}^1$, $m_{O \rightarrow N}^2$, and $m_{N \rightarrow O}^2$.
We interpret $m_{N \rightarrow O}^1$ and $m_{O \rightarrow N}^2$ in the main paper, and discuss the interpretation of $m_{O \rightarrow N}^1$ in the supplement.
We do not interpret $m_{N \rightarrow O}^2$ as it is used to refine belief $\tilde{b}_O$ to $b_O$ which is not used anywhere.
For \ContCom, we interpret messages of length 2 and for \DiscCom, we interpret vocabulary of size 2 and 3 (see supplement for vocabulary size 3).

\subsection{\textbf{\ContCom} interpretation}

\xhdr{What does \atwo tell \aone in $m_{N \rightarrow O}^1$?}
This is the first message that \atwo sends to \aone.
We hypothesize that it is used to communicate the goal object color.
This is intuitive as \aone needs to know the goal to which \atwo must navigate.
This is similar to a human asking ``where is the green object goal located?''
\Cref{fig:cont_comm_len2_m_N_to_O} shows the distribution of the two elements of $m_{N \rightarrow O}^1$ \wrt goal object category (x-axis).
The data for the plot is collected from each step across 1,000 validation episodes.
It appears that \atwo sends different messages for different objects.
To test this hypothesis, we quantify the correlation between $m_{N \rightarrow O}^1$ and the goal object.
We fit linear probes~\cite{alain2016understanding} on $m_{N \rightarrow O}^1$ to classify goal objects.
Linear probes use linear classifiers to map input data to output and are trained using a cross-entropy loss.
We use the same data for this analysis as for \Cref{fig:cont_comm_len2_m_N_to_O}.  We split the data into train and val with a ratio of 3:1 and train the probe to predict the goal object category with $m_{N \rightarrow O}^1$ as input.
The probe achieves an accuracy of 69.7\% on the val split, supporting our hypothesis that $m_{N \rightarrow O}^1$ communicates the goal object color.

\xhdr{What does \aone tell \atwo in $m_{O \rightarrow N}^2$?}
This is the second message that \aone send to \atwo. We hypothesize that \aone uses it to communicate the relative position of the goal w.r.t. \atwo.
This is akin to a human saying ``the goal you asked for is in front of you.'' 
\Cref{fig:cont_comm_len2_m_O_to_N} shows the distribution of the two elements of $m_{O \rightarrow N}^2$ against the current object goal in the spatial reference frame defined by the position and orientation of \atwo (egocentric frame) at the environment step when the message was sent.
In the figure, the agent is facing up and the field-of-view is marked by red lines.
When the goal is in front of \atwo, \aone sends smaller values for the $1^\text{st}$ element and higher values for the $2^\text{nd}$ element of $m_{O \rightarrow N}^2$.
We observe that the emergent communication exhibits an angular pattern.
To quantify this observation, we again fit linear probes.
Given $m_{O \rightarrow N}^2$, we predict the angle of the goal object \wrt \atwo's heading direction (+y axis).
Since the plot is mostly symmetric about the y-axis, we take the absolute value of the angle from the heading direction and bin the angles into 4 bins: $[0^\circ, 45^\circ), [45^\circ, 90^\circ), [90^\circ, 135^\circ), [135^\circ,$ $180^\circ)$.
Given $m_{O \rightarrow N}^2$, our probe has to predict the bin to which the goal location would belong.
We observe a classification accuracy of 58\% (compared to chance accuracy of 25\%), providing support for our hypothesis that $m_{O \rightarrow N}^2$ conveys the egocentric relative position of the goal.

Since both \aone and \atwo send messages that are statistically dependent on their respective observations, we can conclude that they exhibit positive signaling~\cite{lowe2019pitfalls} (sending messages related to their observations or actions).

\subsection{\textbf{\DiscCom} interpretation}

In this communication mechanism, the messages exchanged between the agents consist of probabilities $p_1$ and $p_2$ for words $w_1$ and $w_2$ respectively.
In \Cref{fig:probability-distributions} we plot the distribution of $p_1$ for messages $m_{N \rightarrow O}^1$ and $m_{O \rightarrow N}^2$ on all val set episodes (note that $p_2 = 1 - p_1$, which can hence be inferred from the distribution of $p_1$).
We observe that most probabilities for vocabulary of size 2 are close to 0 or 1.
Based on this observation, for vocabulary size 2, we bin the probabilities into three classes: \symone ($p_1 < 0.2$), \symtwo ($0.2 \leq  p_1 \leq 0.8$), or \symthree ($p_1 > 0.8$).
Here, we only interpret for vocabulary size 2 and defer the interpretation for vocabulary size 3 to the supplement.

\begin{figure*}
\resizebox{\linewidth}{!}{
\newcolumntype{C}{>{\centering\arraybackslash} m{8.8cm} }
\begin{tabular}{@{}CCCC@{}}
    \includegraphics[width=1\linewidth]{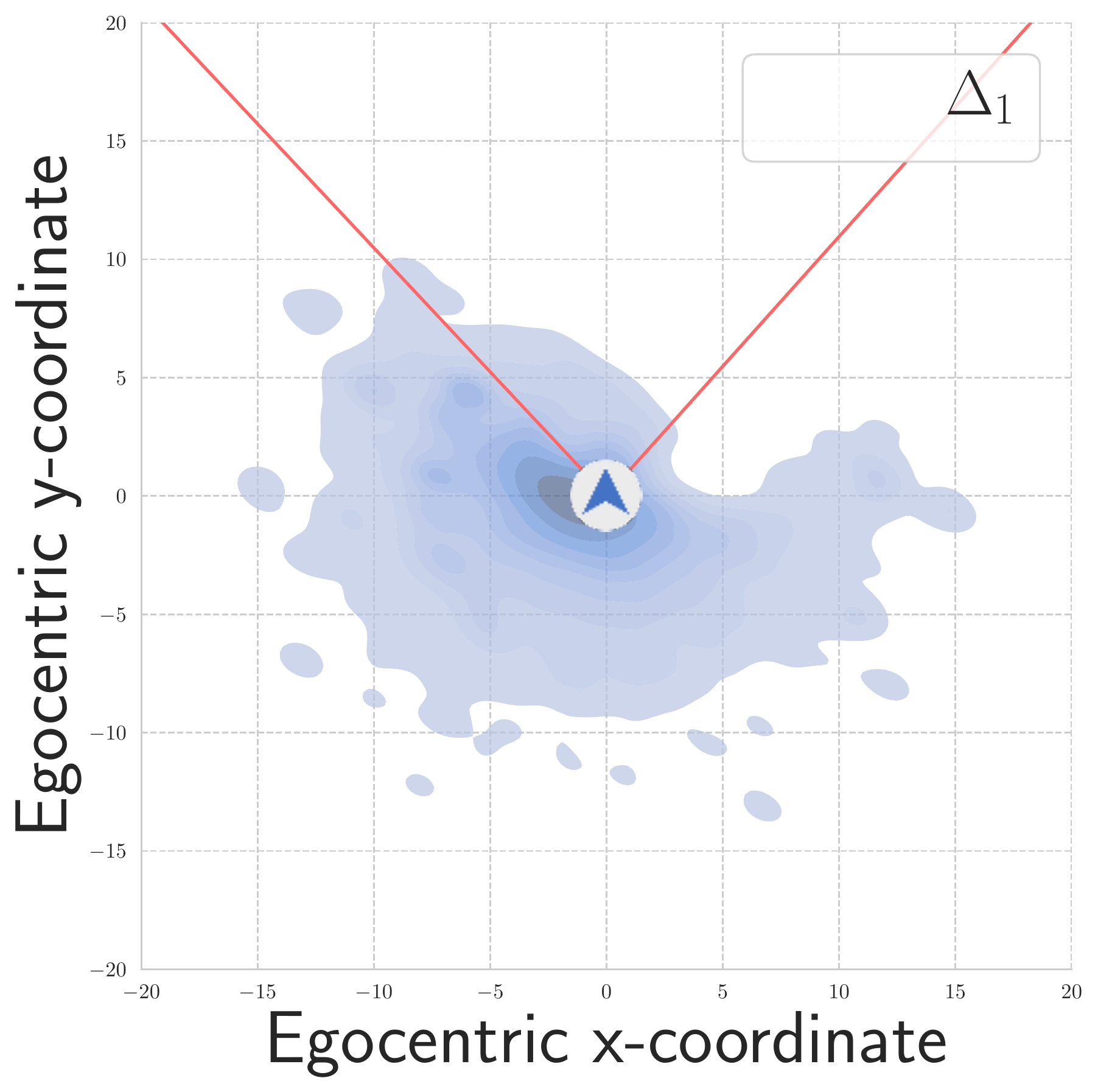} &  
    \includegraphics[width=1\linewidth]{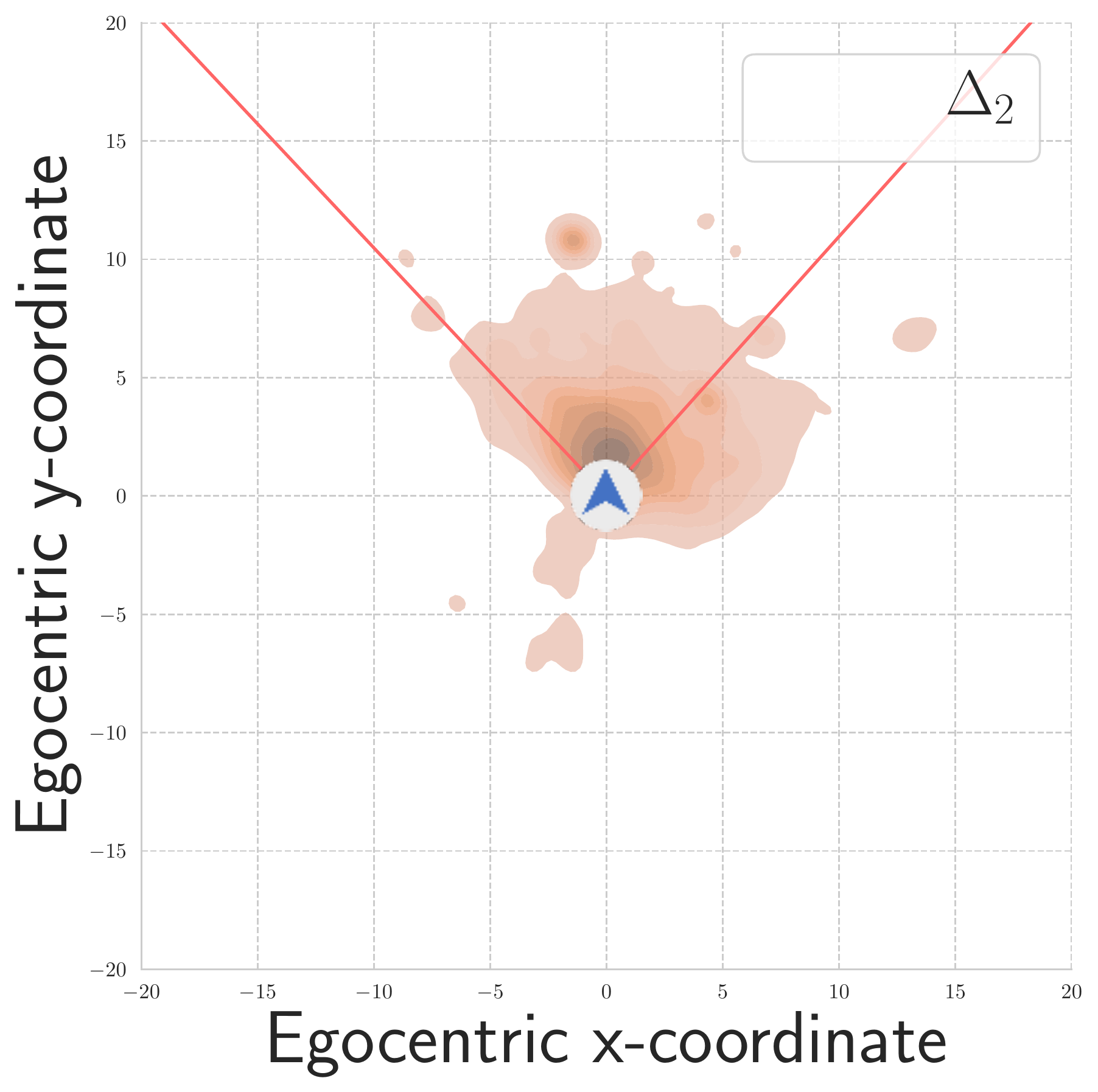} &  
    \includegraphics[width=1\linewidth]{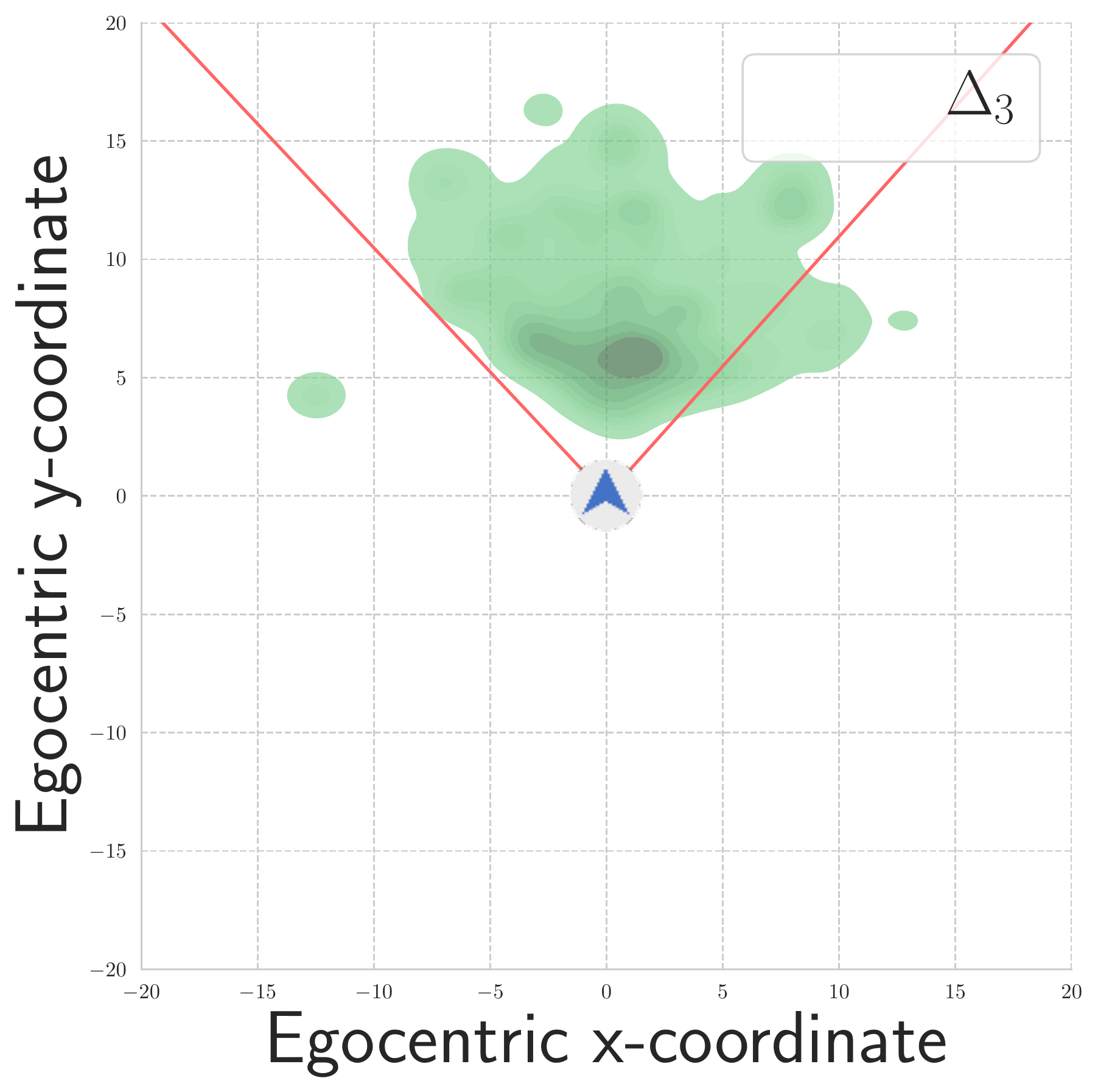} &  
    \includegraphics[width=1\linewidth]{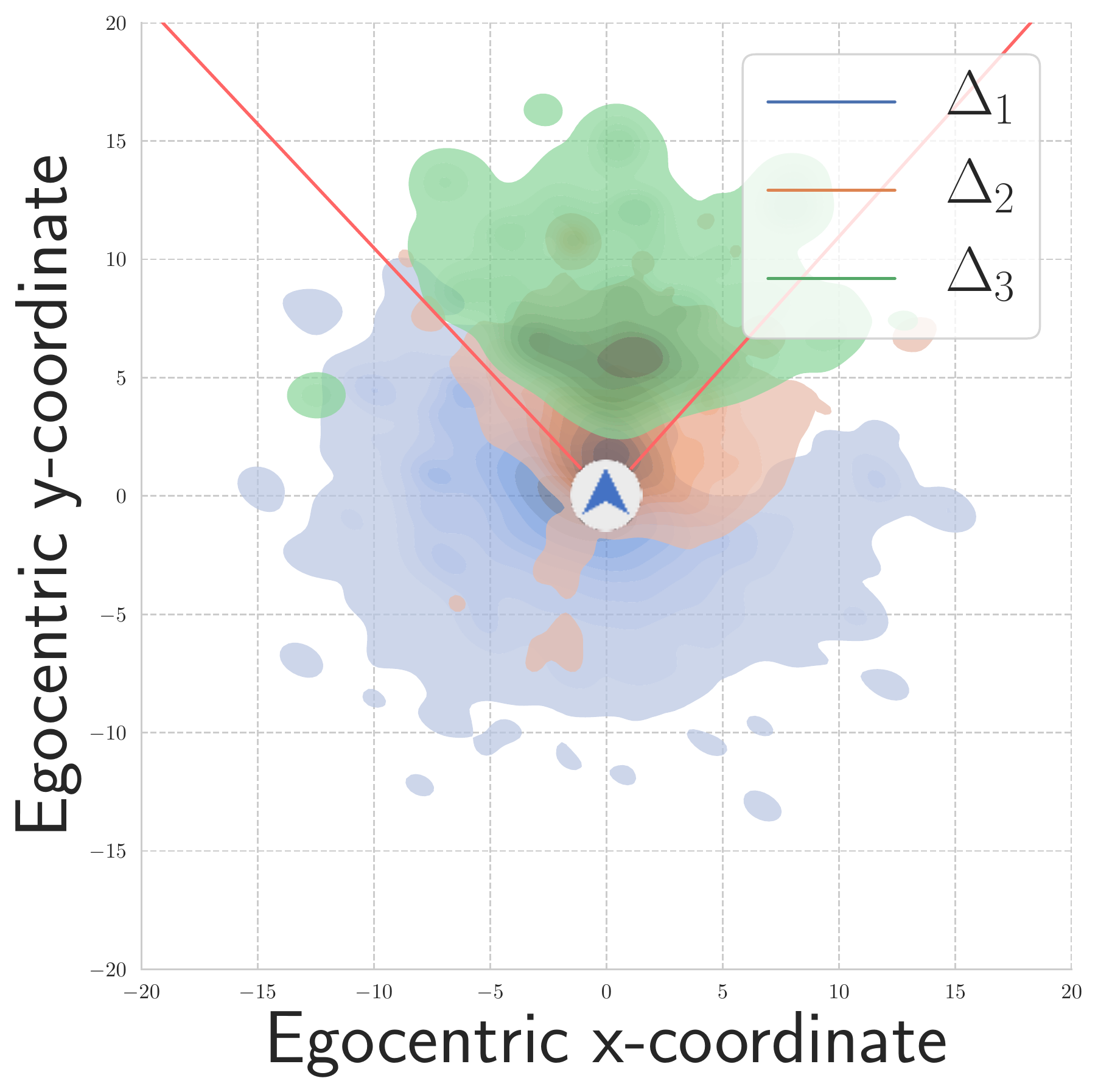}   
\end{tabular}
}
\vspace{3mm}
\caption{
\textbf{Egocentric visualization of \DiscCom communication symbol} $m_{O \rightarrow N}^2$.
The plots show the relative coordinates of the current goal object from \atwo's perspective when \aone communicates the symbol through \DiscCom with vocabulary size two.
The navigator agent (\atwo) is facing the +y axis and its field-of-view is marked with red lines.
Data points are accumulated across all validation episodes, and we plot contour lines of the bivariate density distribution.
Each data point is a message with $(x,y)$ coordinates determined from the coordinates of the current goal object in \atwo's egocentric reference frame when the message was sent.
The first three plots are for each communication symbol, and the right-most combines all symbols.
Note how each symbol represents distinct regions that are egocentrically organized around the agent: \symone captures `behind and not visible', \symtwo corresponds mostly to `close, in front', and \symthree is `farther in front'.
}
\label{fig:m21_1ON}
\end{figure*}

\begin{figure}
\centering
 \includegraphics[width=\linewidth]{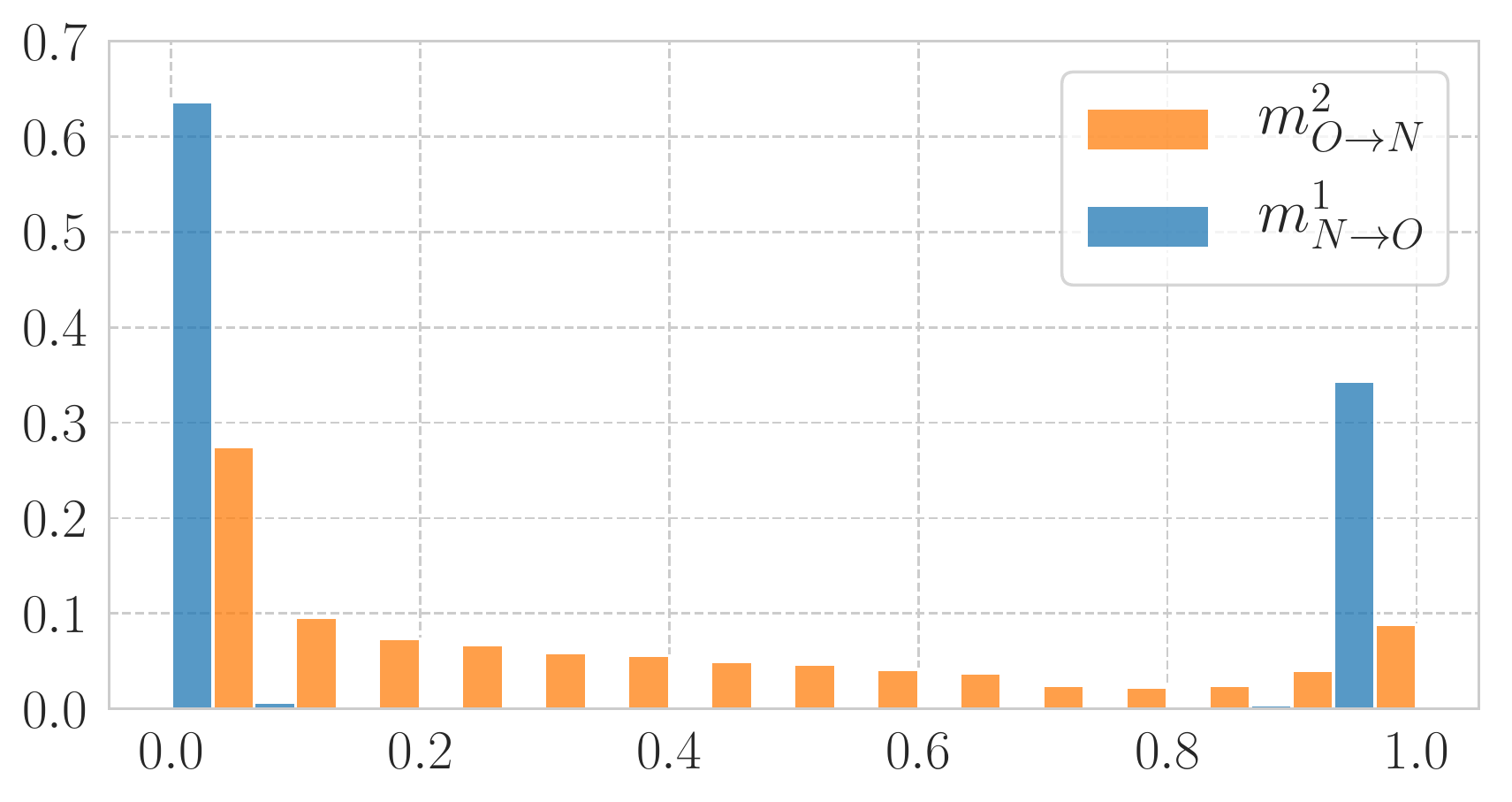}
\caption{
\textbf{Distribution of probability weight $p_1$ associated with $w_1$ in messages $m_{N \rightarrow O}^1$ and $m_{O \rightarrow N}^2$ for \DiscCom.}
The vocabulary consists of two words, $w_1$ and $w_2$. Since $p_1+p_2=1$, we only plot $p_1$ here. For $m_{N \rightarrow O}^1$, probabilities are concentrated at $p_1=0$ and $p_1=1$. For $m_{N \rightarrow O}^1$, distribution is comparatively uniform with higher probabilities at $p_1=0$ and $p_1=1$.}
\label{fig:probability-distributions}
\end{figure}

\xhdr{What does \atwo tell \aone in $m_{N \rightarrow O}^1$?}
We again hypothesize that \atwo uses $m_{N \rightarrow O}^1$ to communicate which goal it has to navigate to.
Since there are eight goal categories, \atwo needs to communicate which one is the current goal.
We observe that \atwo sends \symone when the goal object is a red, white or black, and sends \symtwo otherwise.
To quantify the correlation between the communication symbol and the current goal, we train a random forest classifier that predicts the communication symbol given the object category.
Here, we use random forest classifiers rather than linear probes to better handle the non-linear decision boundaries that emerge in the $m_{O \rightarrow N}^2$ interpretation as seen in \Cref{fig:m21_1ON}.
Note that to interpret \ContCom, we predict properties like goal category or goal direction using the messages. In contrast, to interpret \DiscCom, we predict communication symbols using properties.
In both cases, we predict a discrete variable like object category or goal direction in \ContCom and communication symbol in \DiscCom.
The classifier used here is trained using data from all the validation episodes.
The data is split into train and test sets and our classifier attains almost 100\% accuracy on the test set (see supplement).

\xhdr{What does \aone tell \atwo in $m_{O \rightarrow N}^2$?}
Similar to \ContCom, \aone utilizes $m_{O \rightarrow N}^2$ to communicate the goal location.
\Cref{fig:m21_1ON} shows the symbols sent by \aone against the relative location of the current object goal in the egocentric frame of \atwo when the message was sent (similar to \Cref{fig:cont_comm_len2_m_O_to_N}). 
Points are accumulated across 1000 validation episodes of \mon{1}.
We observe that \aone communicates \symone, \symtwo or \symthree depending on the position of the current target object with respect to \atwo.
To verify this observation, we train a random forest classifier to predict the communication symbol from the $(x,y)$ coordinate of the current target goal in \atwo's reference frame.
We observe an accuracy of about 89\% with high precision and recall for all three classes \symone, \symtwo and \symthree (details in supplement).
With a larger vocabulary of size 3 \aone can send even more fine-grained information about the location of the current goal (see supplement).
In both cases, we observe that the majority of symbols are associated with areas within the field of view of \atwo (delineated in red).
Thus, \aone uses a higher proportion of the communication bandwidth to communicate to \atwo the location of the current goal if it is in \atwo's field of view.
Possibly, it is more advantageous for \atwo to have precise information about the goal location when it is in front.
If the goal is in the field of view, \aone sends a different symbol depending on the distance of the current goal from \atwo.
Here also, messages sent by \aone and \atwo are dependent on their observations. Hence, both of them exhibit positive signaling.

\xhdr{Are \atwo's actions influenced by $m_{O \rightarrow N}^2$?}
Agents exhibit positive listening~\cite{lowe2019pitfalls} if the received messages influence the agent policy.
\Cref{tab:struccom_action_symbol} reports the percentage of each action taken for communication symbols in \DiscCom (with vocabulary size 2).
We observe that \atwo never calls \found when it receives \symthree.
This is intuitive as \symthree is communicated when the goal is far ahead of \atwo.
We also observe that \atwo is more likely to move forward when it receives \symthree as compared to \symone or \symtwo.
This is also intuitive as \atwo is more likely to move forward when the goal is far ahead.

\begin{table}
\centering
\vspace{2pt}
\resizebox{\columnwidth}{!}{
\begin{tabular}{@{}l cccc@{}}
\toprule
 & \textsc{Found} & \textsc{Forward} & \textsc{Turn Left} & \textsc{Turn Right} \\
\midrule
\symone & 0.8 & 43.9 & 24.7 & 30.6 \\
\symtwo & 0.3 & 52.2 & 28.7 & 18.8 \\
\symthree & 0.0 & 63.4 & 18.9 & 17.7\\
\bottomrule
\end{tabular}
}
\vspace{2pt}
\caption{
\textbf{Distribution over actions taken by \atwo upon receiving each \DiscCom communication symbol (vocabulary size 2).}
Values in each row report percentage out of all actions taken when that symbol is received.
Note that \symthree leads to a high percentage of \textsc{Forward} actions and no \textsc{Found} actions.
This is intuitive in light of the spatial distribution of goal positions relative to \atwo when \symthree is communicated, as visualized in \Cref{fig:m21_1ON}.
}
\label{tab:struccom_action_symbol}
\end{table}

\xhdr{What happens when the goal is in \atwos view?}
The distribution of the exchanged messages remains unchanged, but how \atwo acts based on the received messages is different.
We performed two experiments at evaluation time to study this case.
\textbf{1)} \aone sends random messages when the goal is visible to \atwo. 
We find this does not change the overall performance of \atwo.
\textbf{2)} We insert an incorrect goal in the scene while keeping \aones map unchanged.
\Progress and \PPL metrics drop to 29\% and 7\% respectively.
We conclude that when the goal is visible, \atwo ignores messages from \aone and relies on its perception to navigate.

\section{Conclusion}
\label{sec:conclusion}
We proposed the collaborative multi-object navigation task (CoMON) for studying the grounding of learned communication between heterogeneous agents.
Using this task, we investigated two families of communication mechanisms (structured and unstructured communication) between heterogeneous agents.
We analyzed the emergent communication patterns though an egocentric and spatially grounded lens.
We found the emergence of interpretable perception-specific messages such as `I am looking for X' and egocentric instructions such as `look behind' and `goal is close in front.'
We believe the \task task, along with the interpretation framework for communication between agents that we presented will help to enable systematic study of grounded communication for embodied AI navigation agents.

\xhdr{Acknowledgements}
We thank the anonymous reviewers for their suggestions and feedback.  
This work was funded in part by a Canada CIFAR AI Chair, a Canada Research Chair and NSERC Discovery Grant, and enabled in part by support provided by \href{www.westgrid.ca}{WestGrid} and \href{www.computecanada.ca}{Compute Canada}. This work is supported in part by NSF under Grant \#1718221, 2008387, 2045586.

{\small
\bibliographystyle{plainnat}
\setlength{\bibsep}{0pt}
\bibliography{multiON}
}

\clearpage
\appendix
\section{Supplementary material}
\label{sec:supp}

\begin{table*}[t]
\ra{1.3}
\centering
\resizebox{\linewidth}{!}{
\begin{tabular}{H{2cm}H{6cm}H{2cm}H{6cm}}
\toprule
Notation & Description & Notation & Description\\
\midrule
\mon{m}&Episode with $m$ ordered object goals&$\hat{b}_N$&Initial belief of \atwo\\
$G$&Sequence of goal objects&$b_O$&Final belief of \aone\\
\aone&Oracle agent&$b_N$&Final belief of \atwo\\
\atwo&Navigator agent&$v_a$&Embedding of previous action $a_{t-1}$\\
$M$&Oracle map in global frame&$s$&Final state representation\\
$o_t$&Egocentric RGBD frames&$m_{N \rightarrow O}^r$&Message sent by \aone to \atwo in round $r$\\
$g_t$&Current goal object one-hot vector&$m_{O \rightarrow N}^r$&Message sent by \atwo to \aone in round $r$\\
$a_t$&Action taken by the agent&$r_t$&Reward at time-step $t$\\
\ContCom&Unstructured communication&$r^{\text{goal}}$&Reward for finding a goal\\
\DiscCom&Structured communication&$r^{\text{closer}}$&Reward for moving closer to goal\\
$E$&Oracle map in egocentric frame&$r^{\text{time penalty}}$&Time penalty reward\\
$v_o$&RGBD features&$w_i$&Embedding of word $i$\\
$v_g$&Embedding of one-hot goal vector $g_t$&$p_i$&Probability for word $i$\\
$\hat{b}_O$&Initial belief of \aone&$\Delta_i$&Communication symbol $i$\\
\bottomrule
\end{tabular}
}
\vspace{1pt}
\caption{\textbf{Summary of notation.} Subscript $t$ denotes the corresponding notation at time step $t$}
\label{tab:notation}
\end{table*}

This supplemental document provides the following additional contents to support the main paper:
\begin{compactitem}
\item \ref{sec:notation} Overview of notation used in the paper
\item \ref{sec:full_eval} Additional quantitative evaluation
\item \ref{sec:m_O_N_comm_interpret} Interpretation of $m_{O \rightarrow N}^1$ for \mon{1}
\item \ref{sec:disc_comm_interpret} Interpretation of \DiscCom using vocabulary size 3
\item \ref{sec:comm_2ON} Interpretation of \ContCom and \DiscCom for \mon{2}
\item \ref{sec:additional_info} Additional analyses to check for information content of messages
\item \ref{sec:episode_viz} Episode map visualizations
\item \ref{sec:impl_details} Implementation details
\end{compactitem}

\subsection{Notation overview}
\label{sec:notation}

\Cref{tab:notation} provides a summary of the definitions of important notations used in the paper and this supplementary document.

\subsection{Additional quantitative evaluation}
\label{sec:full_eval}

We report only the \PPL and \Progress metrics in the main paper.
\Cref{tab:full-eval} summarizes the complete set of evaluation metrics we use: \Progress, \PPL, \Success and \SPL.
The trends in \Success and \SPL are similar to those in \Progress and \PPL.
We also perform generalization experiments by evaluating models trained on \mon{3} on the more difficult \mon{4} and \mon{5} tasks. \Cref{tab:full-eval} summarizes those results as well.
We observe that \DiscCom outperforms \ContCom in these generalization experiments as well.
\begin{table*}[h]
\ra{1.3}
\centering
\resizebox{\linewidth}{!}{
\begin{tabular}{@{}l rrrrr rrrrr rrrrr rrrrr@{}}
\toprule
 & \multicolumn{5}{c}{\Progress (\%)} & \multicolumn{5}{c}{\PPL (\%)} &  \multicolumn{5}{c}{\Success (\%)} & \multicolumn{5}{c}{\SPL (\%)}\\  \cmidrule(lr){2-6} \cmidrule(lr@{0mm}){7-11} \cmidrule(lr){12-16} \cmidrule(lr){17-21}
 & \mon{1} & \mon{2} & \mon{3} & \mon{4} & \mon{5} & \mon{1} & \mon{2} & \mon{3} & \mon{4} & \mon{5} & 
 \mon{1} & \mon{2} & \mon{3} & \mon{4} & \mon{5} & \mon{1} & \mon{2} & \mon{3} & \mon{4} & \mon{5}\\
\midrule
 \NoMap & 56&39&26&10&7 & 35&26&16&7&5 & 56&30&10&7&2 & 35&18&5&3&1  \\
 \RandContCom & 59&40&28&7&5 & 36&28&18&3&2 & 58&33&12&0&0 & 32&20&6&0&0  \\
 \RandDiscCom & 50&31&24&16&10 & 33&24&16&11&6 & 50&30&9&6&1 & 33&16&5&3&1 \\
 \ContCom & \textbf{87}&77&63&41&26 & 60&51&39&23&13 & \textbf{87}&57&53&23&7 &60&43&40&13&3\\
 \DiscCom & 85&\textbf{80}&\textbf{70}&\textbf{50}&\textbf{35} & \textbf{67}&\textbf{59}&\textbf{50}&\textbf{32}&\textbf{22} & 85&\textbf{65}&\textbf{58}&\textbf{32}&\textbf{14} & \textbf{67}&\textbf{46}&\textbf{45}&\textbf{20}&\textbf{9}\\
\midrule
 \OracleMap & 89&80&70&45&26 & 74&64&52&28&14 & 89&69&61&27&8 & 74&49&42&16&4 \\
\bottomrule
\end{tabular}
}
\vspace{1pt}
\caption{\textbf{Additional quantitative metrics on} \mon{1}, \mon{2} \textbf{and} \mon{3} \textbf{tasks and generalization to \mon{4} and \mon{5}.
}
All agents are trained on \mon{3} and evaluated on the task indicated in each column.
}
\label{tab:full-eval}
\end{table*}

\begin{table}
\ra{1.3}
\centering
\resizebox{\linewidth}{!}{
\begin{tabular}{@{}lc rrr rrr@{}}
\toprule
 & \multirow{2}{*}{Dim} & \multicolumn{3}{c}{\Progress (\%)} & \multicolumn{3}{c}{\PPL (\%)}\\ \cmidrule(lr){3-5} \cmidrule(lr@{0mm}){6-8}
 & & \mon{1} & \mon{2} & \mon{3} & \mon{1} & \mon{2} & \mon{3}\\
\midrule
 \multirow{3}{*}{\ContCom} & 2 & 87 & 77 & 63 & 60 & 51 & 39\\
 & 3 & 88 & 78 & 67 & 66 & 55 & 45\\
 & 4 & 88 & 79 & 68 & 66 & 57 & 46\\
 
\midrule
\multirow{3}{*}{\DiscCom} & 2 & 85 & \textbf{80} & \textbf{70} & 67 & 59 & 50\\
& 3 & 89 & 78 & \textbf{70} & \textbf{72} & 57 & 52\\
& 4 & \textbf{90} & \textbf{80} & \textbf{70} & 67 & \textbf{60} & \textbf{54}\\

\bottomrule
\end{tabular}
}
\vspace{1pt}
\caption{\textbf{Effect of message length on performance.} \DiscCom outperforms \ContCom on similar message length and there is a slight improvement with increasing message dimension.}
\label{tab:message-dimension}
\end{table}

\Cref{tab:message-dimension} shows the effect of increasing message length on the task performance.
2-dimensional message results are repeated from \Cref{tab:results}.
\DiscCom still outperforms \ContCom on equal message lengths.
Overall, there are small improvements with increasing message dimension.
We hypothesize that \aone can encode the goal location in 2-dimensional messages, thus higher-dimensional messages provide small improvements.

\subsection{Interpretation of $m_{O \rightarrow N}^1$ for \mon{1}}
\label{sec:m_O_N_comm_interpret}

In the \task task, \atwo has knowledge of the goal to visit while \aone has knowledge of the semantic map (with goal positions) as well as the position and orientation of \aone.  In the main paper, we showed that $m_{O \rightarrow N}^2$ is used to communicate the location of the goal.  Here we consider what information is in the initial message sent from \aone to \atwo ($m_{O \rightarrow N}^1$) for \mon{1}.
Note that in the case of \mon{1}, there is only one goal, so \aone can already send information about that goal without waiting for $m_{N \rightarrow O}^1$.
Using similar methods as we employed in the main paper for $m_{O \rightarrow N}^2$, we find that $m_{O \rightarrow N}^1$ is also used to communicate the location of the goal relative to \atwo for \ContCom and \DiscCom.

\subsubsection{Interpretation of $m_{O \rightarrow N}^1$ in \ContCom}
\label{sec:m_O_N_cont_comm_interpret}

\begin{figure*}
\resizebox{\linewidth}{!}{
\newcolumntype{C}{>{\centering\arraybackslash} m{8.8cm} }
\begin{tabular}{@{}CC@{}}
{$1^{st}$ element} & {$2^{nd}$ element} \\
    \includegraphics[width=1\linewidth]{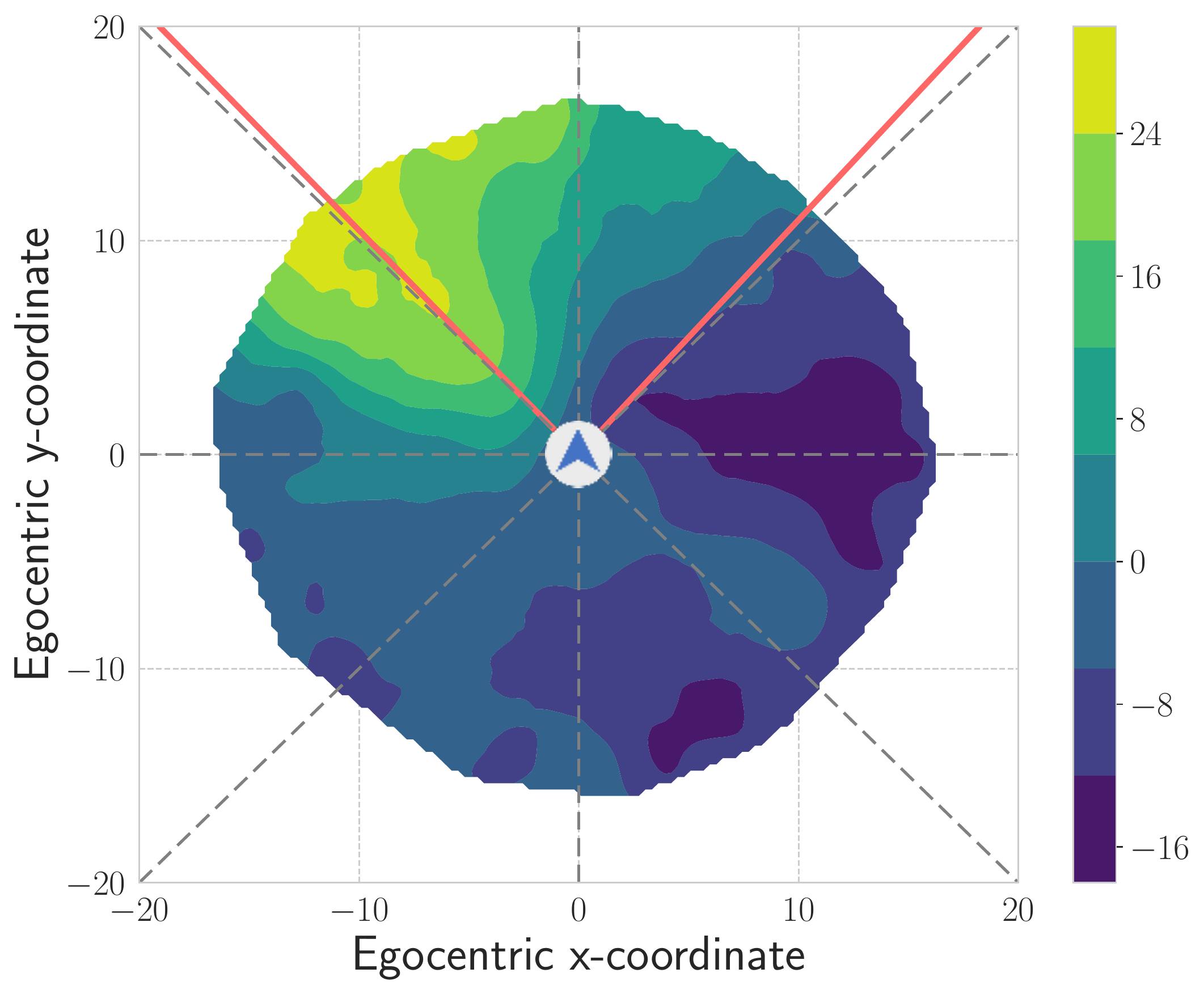}
    &
    \includegraphics[width=1\linewidth]{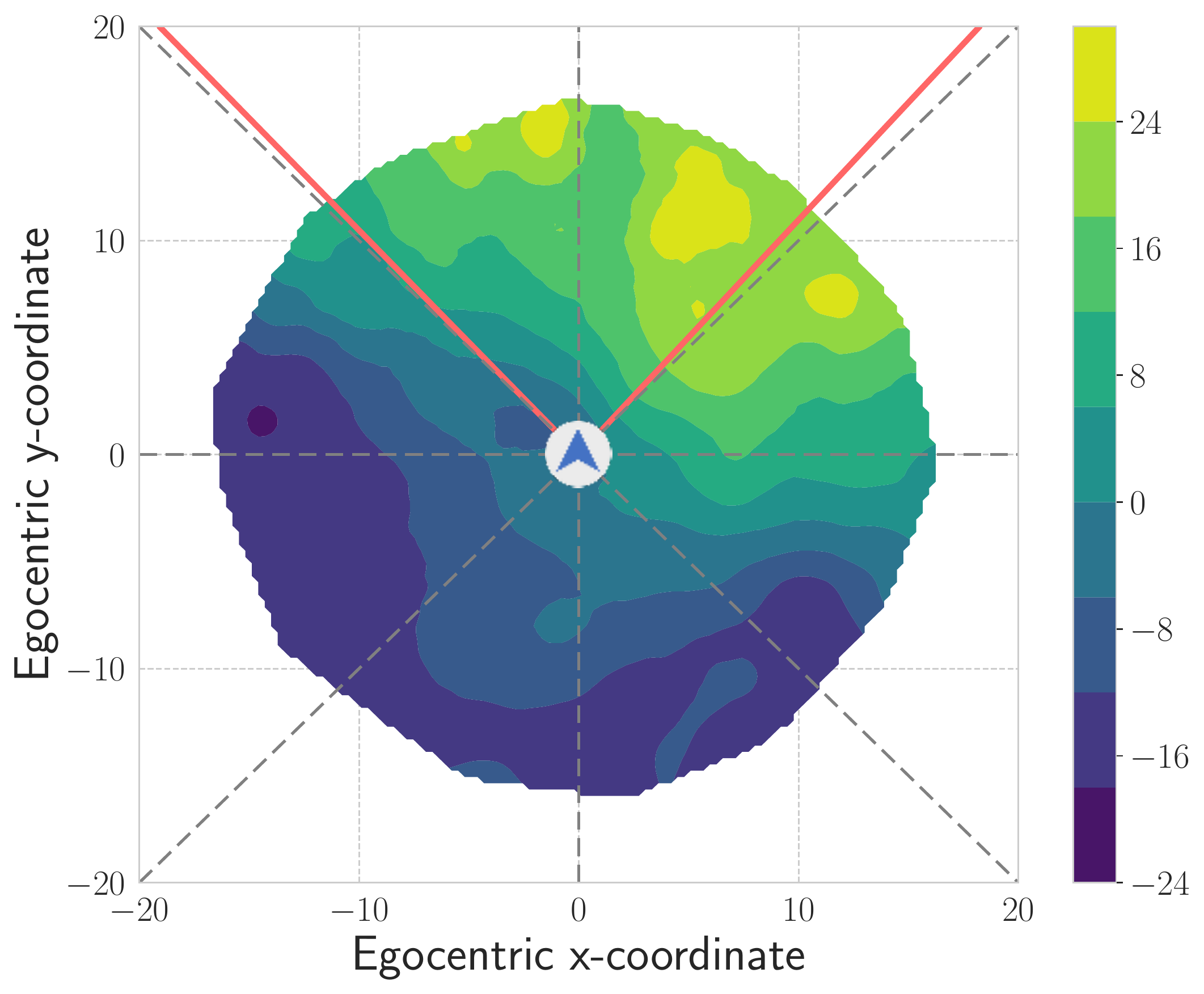}
\end{tabular}
}
\caption{
\textbf{Egocentric visualization of \ContCom communication symbol} $m_{O \rightarrow N}^1$.
The two plots visualize the value of the first and second element of the message plotted \wrt the relative coordinates of the goal object from \atwo.
The navigator agent \atwo is facing the +y axis and its field-of-view is marked with red lines.
The plot on the left corresponds to the $1^\text{st}$ dimension of the message, while the plot on the right corresponds to the $2^\text{nd}$ dimension.
The value of each dimension is indicated by the color hue.
}
\label{fig:cont_comm_len2_m_O_to_N_round_0}
\end{figure*}

\Cref{fig:cont_comm_len2_m_O_to_N_round_0} shows the distribution of $m_{O \rightarrow N}^1$ w.r.t. the relative coordinates of the goal object from \atwo, using a similar visualization as for $m_{O \rightarrow N}^2$ (Figure 5 in the main paper).
We note there is a correlation between $m_{O \rightarrow N}^1$ and the location of the current goal object: with the first element indicating whether the goal is to the left of the agent, and the second element whether the goal is to the right.
To quantify the relation, we again fit linear probes.
As the target of the linear probe, we bin the angles into 8 bins each of $45^\circ$ (see dashed lines in \Cref{fig:cont_comm_len2_m_O_to_N_round_0}). 
Our probe attains classification accuracy of 51\% (compared to chance accuracy of 12.5\%) supporting our hypothesis that $m_{O \rightarrow N}^1$ includes information about the location of the current goal object relative to \atwo.

\subsubsection{Interpretation of $m_{O \rightarrow N}^1$ in \DiscCom}
\label{sec:m_O_N_disc_comm_interpret}

\begin{figure*}
\resizebox{\linewidth}{!}{
\newcolumntype{C}{>{\centering\arraybackslash} m{8.8cm} }
\begin{tabular}{@{}CCCC@{}}
    \includegraphics[width=1\linewidth]{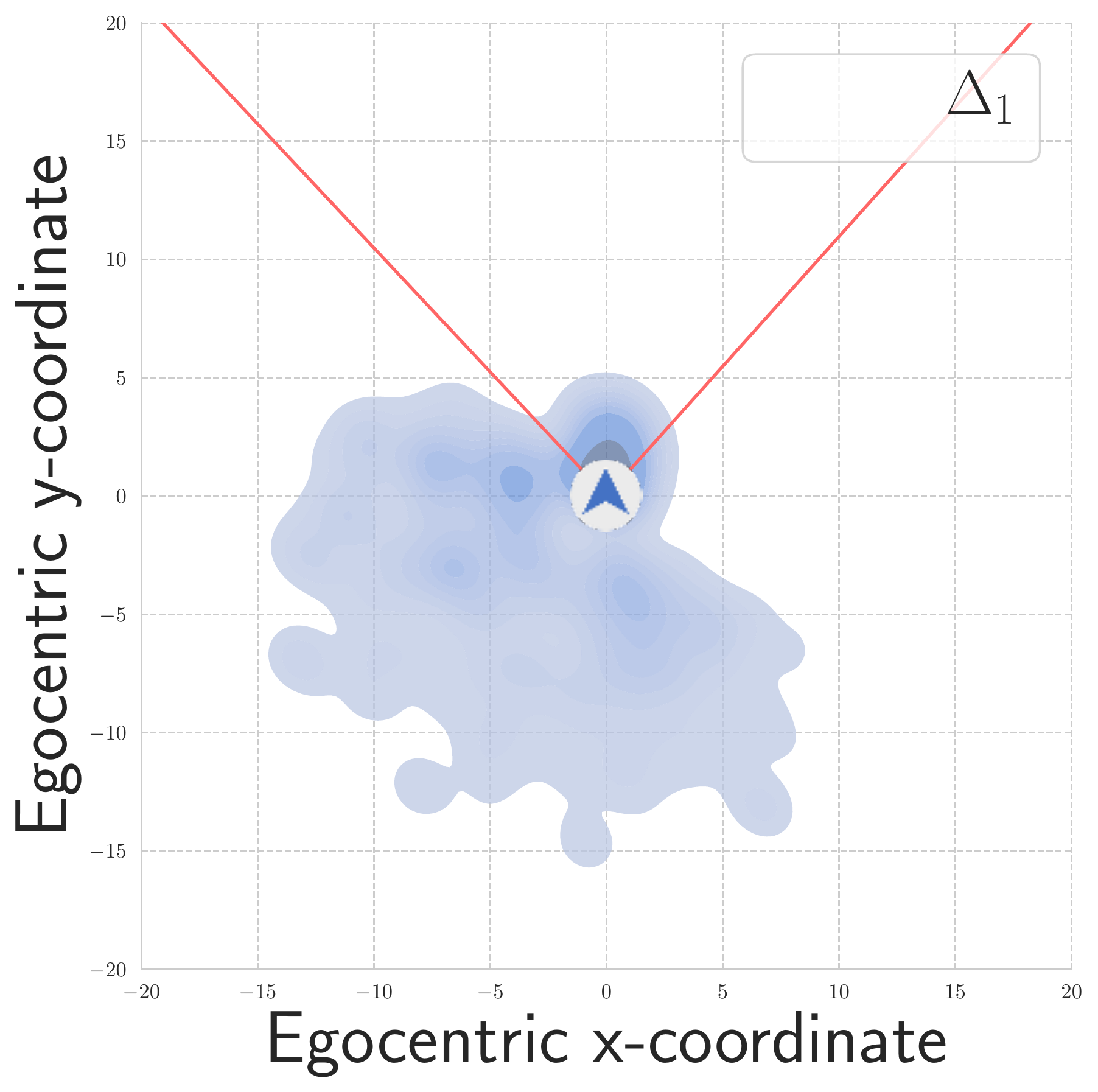} &  
    \includegraphics[width=1\linewidth]{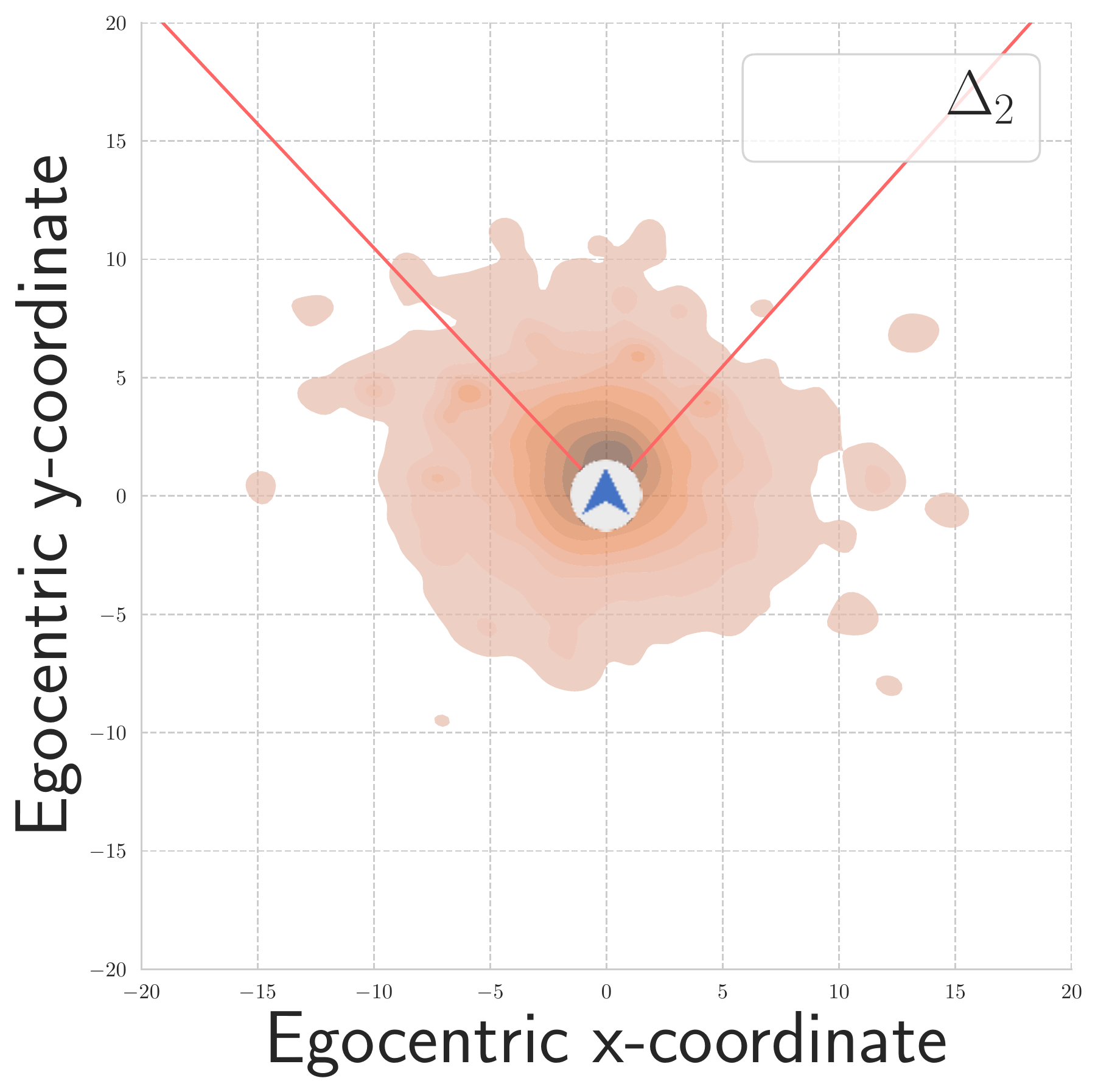} &  
    \includegraphics[width=1\linewidth]{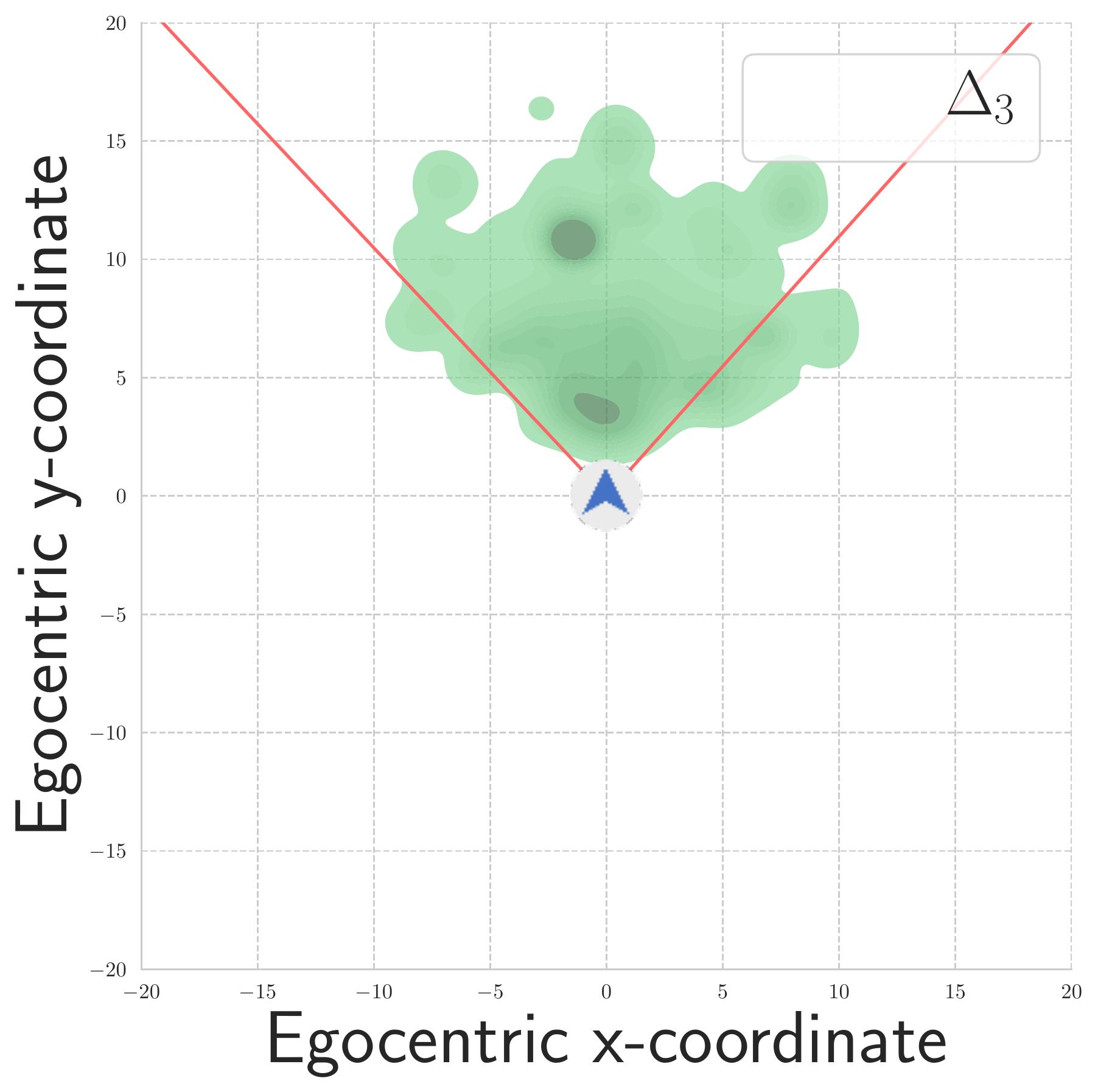} &  
    \includegraphics[width=1\linewidth]{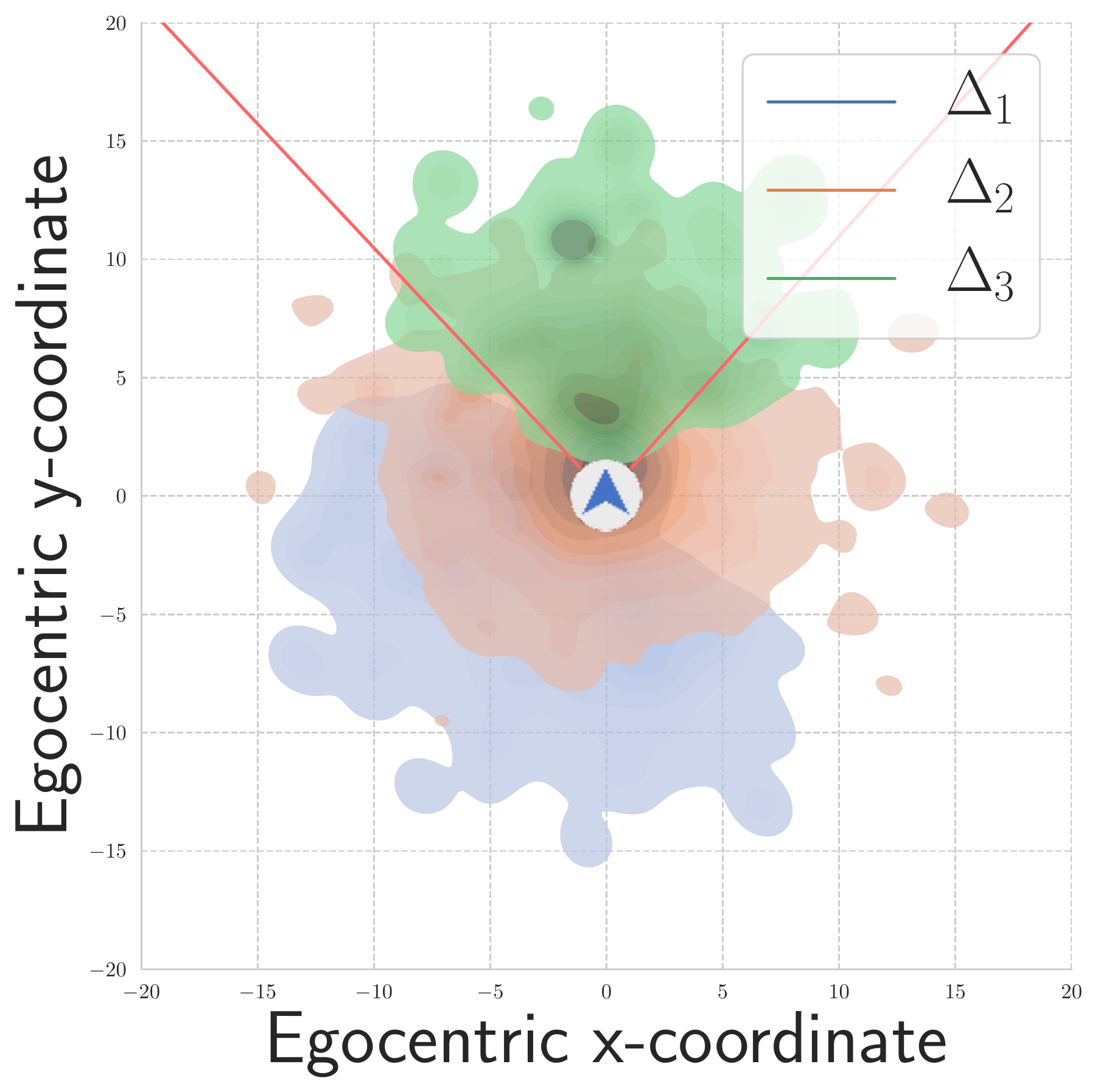}   
\end{tabular}
}
\caption{
\textbf{Egocentric visualization of \DiscCom communication symbol} $m_{O \rightarrow N}^1$.
The plots show the relative coordinates of the current goal object from \atwo's perspective when \aone communicates the symbol through \DiscCom with vocabulary size two.
The navigator agent (\atwo) is facing the +y axis and its field-of-view is marked with red lines.
Data points are accumulated across all validation episodes, and we plot contour lines of the bivariate density distribution.
Each data point is a message with $(x,y)$ coordinates determined from the coordinates of the current goal object in \atwo's egocentric reference frame when the message was sent.
The first three plots are for each communication symbol, and the right-most combines all symbols.
Note how each symbol represents distinct regions that are egocentrically organized around the agent: \symone captures `behind and not visible', \symtwo corresponds mostly to `close, in front', and \symthree is `farther in front'.
}
\label{fig:disc_comm_m_O_to_N_round0}
\end{figure*}

As in the main paper (Section 6.2), we perform a similar analysis for $m_{O \rightarrow N}^1$ as for $m_{O \rightarrow N}^2$, where we use thresholds to group the messages into \symone, \symtwo, and \symthree.  \Cref{fig:disc_comm_m_O_to_N_round0} plots the distribution of each symbol w.r.t. the relative location of the current goal relative to \atwo (similar to Figure 6 in the main paper).
We observe that $m_{O \rightarrow N}^1$ is again used to convey the goal object location, but the correlation between the communicated message and the goal object location is weaker than that of $m_{O \rightarrow N}^2$.  This is evident from the higher overlap of the regions corresponding to each symbol (compared to Figure 6 in the main paper). 
This observation is confirmed by the lower classification accuracy of $83\%$ (vs $89\%$ for $m_{O \rightarrow N}^2$) after training a random forest classifier to predict the communicated symbol from the $(x, y)$ coordinate of the current goal object. 

\begin{figure*}[b]
\resizebox{\linewidth}{!}{
\newcolumntype{C}{>{\centering\arraybackslash} m{8.8cm} }
\newcolumntype{A}{>{\centering\arraybackslash} m{17.6cm} }
\begin{tabular}{@{}CCCA@{}}
    \includegraphics[width=1\linewidth]{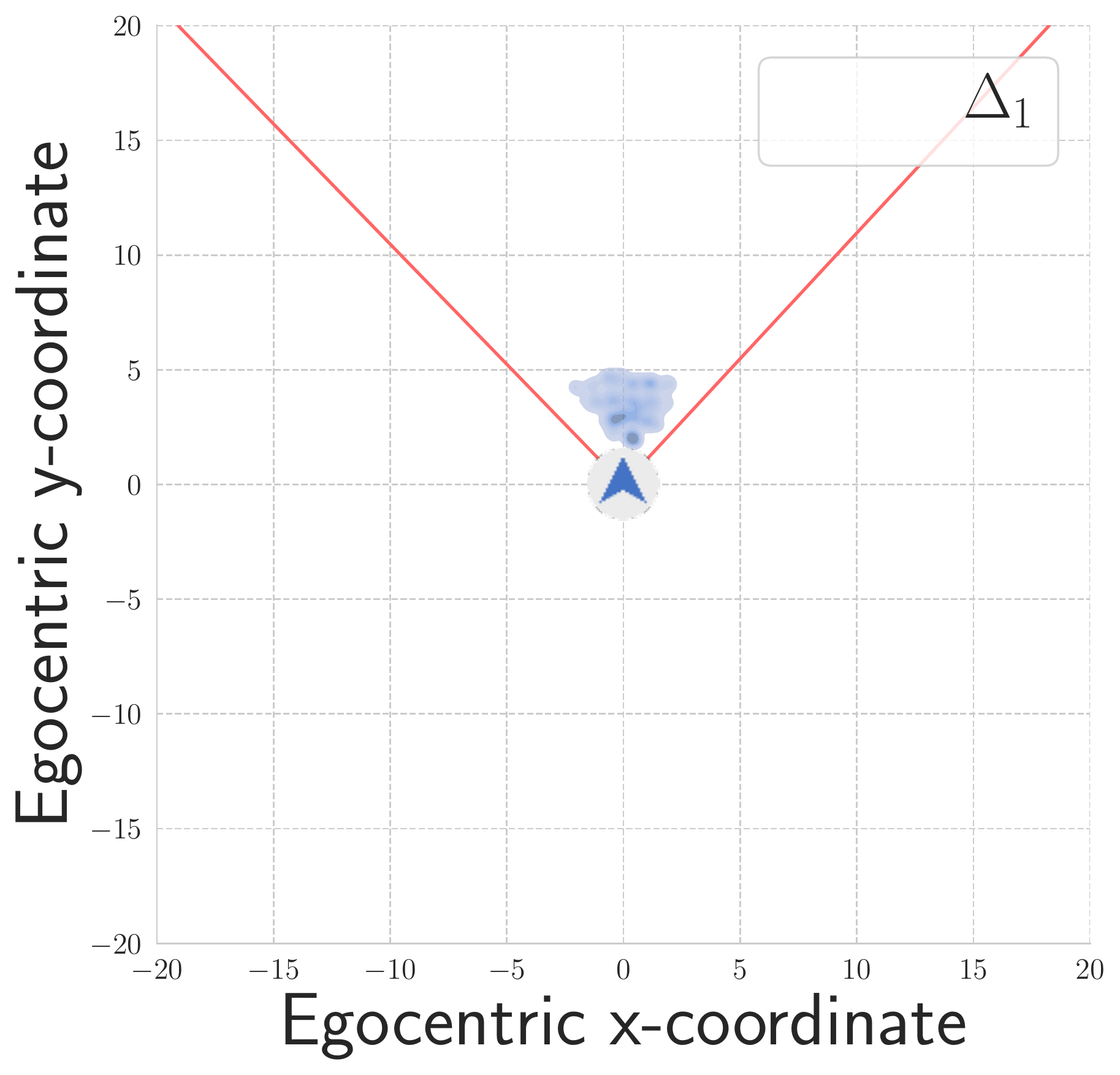} &  
    \includegraphics[width=1\linewidth]{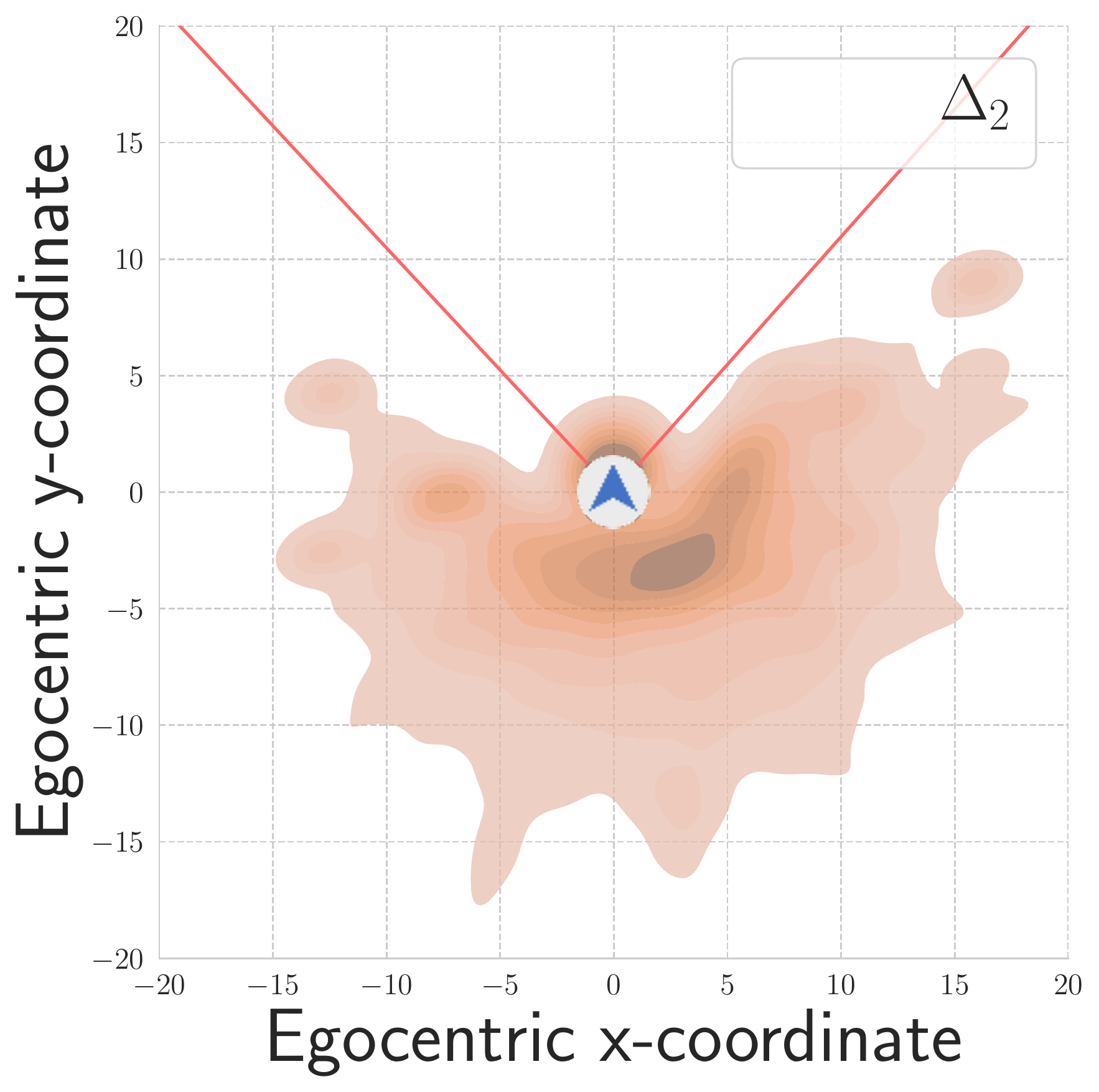} &  
    \includegraphics[width=1\linewidth]{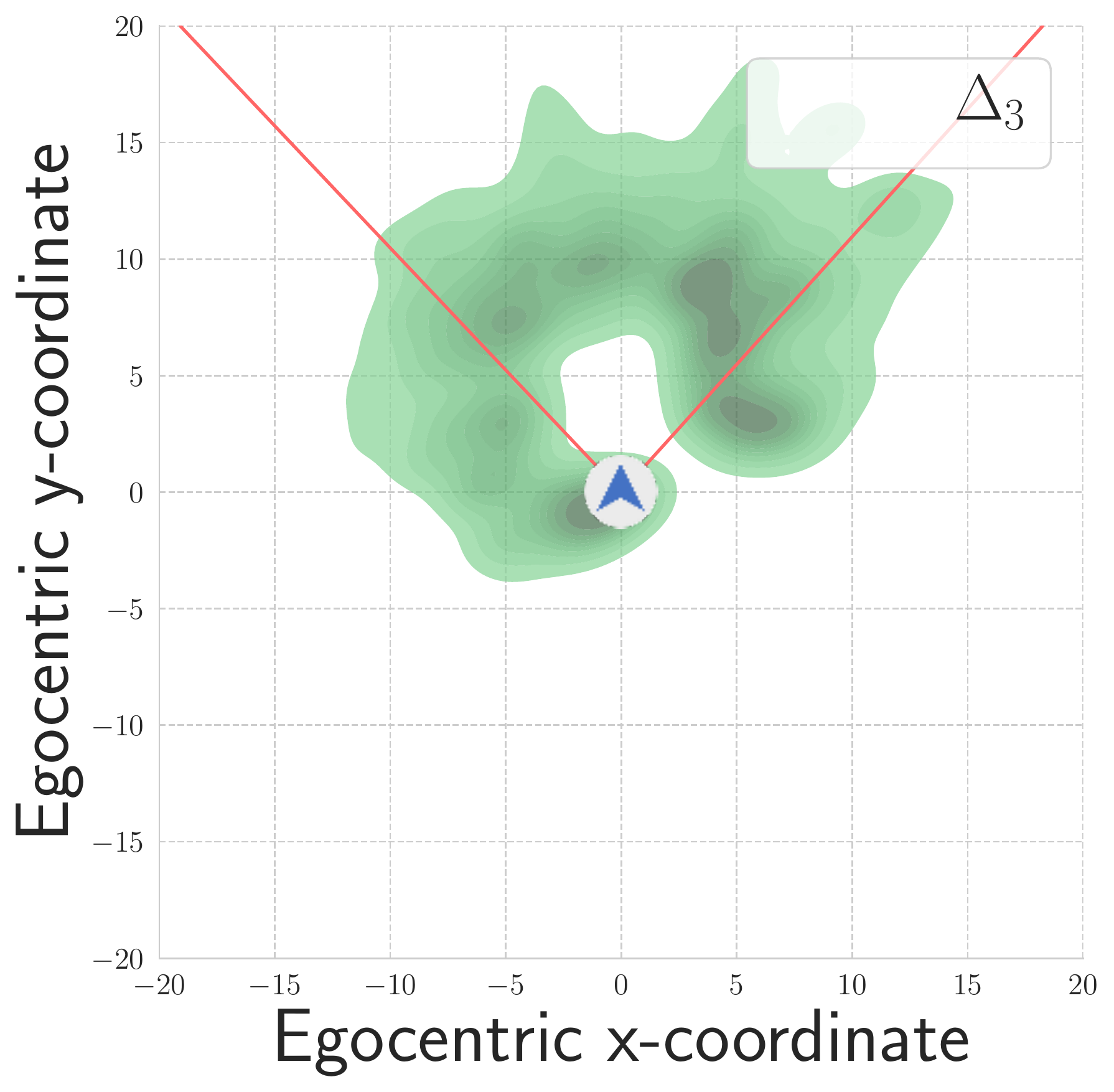} & 
    \multirow{2}{*}[1.5in]{\includegraphics[width=1\linewidth]{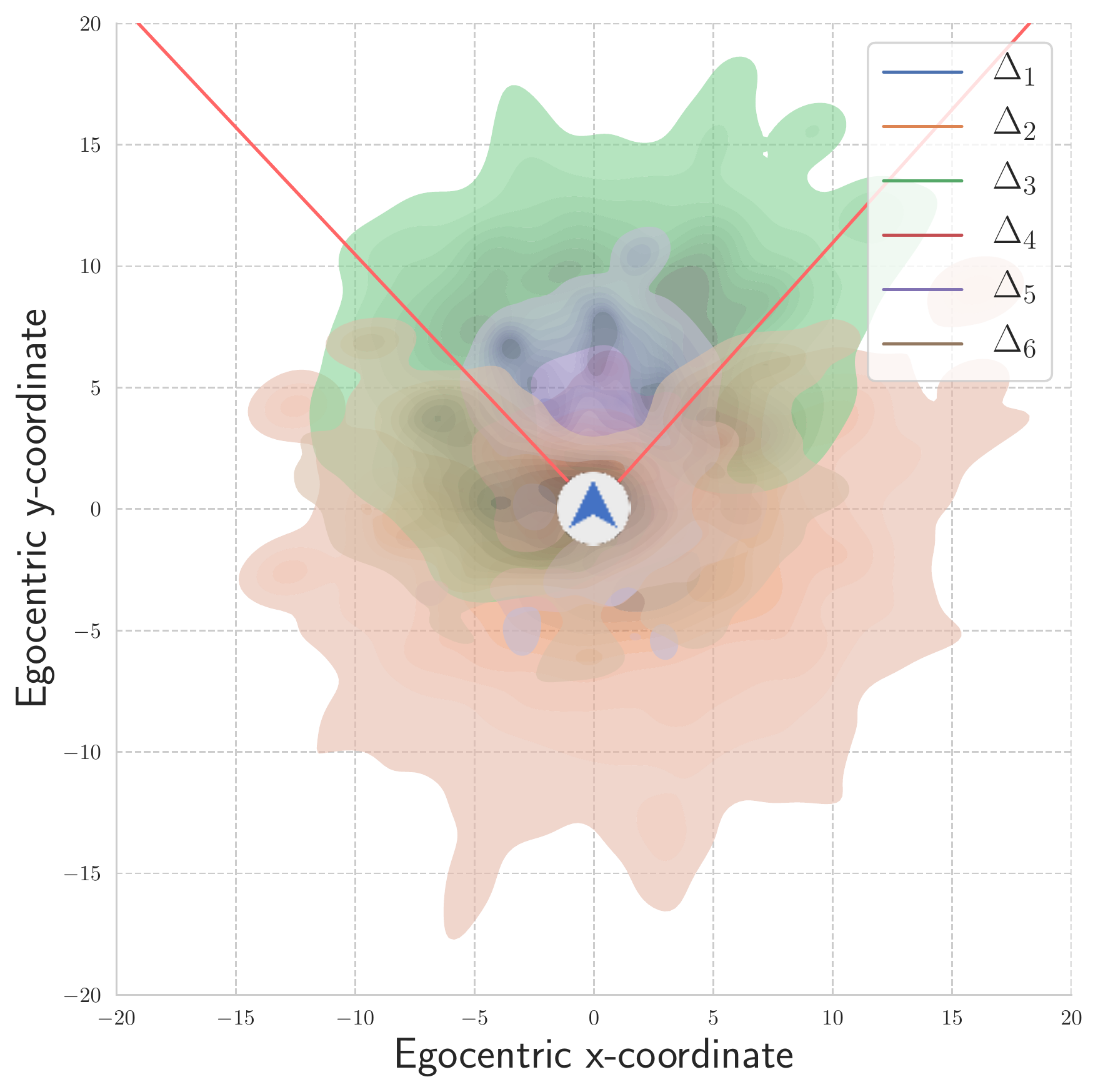}}\\
    \includegraphics[width=1\linewidth]{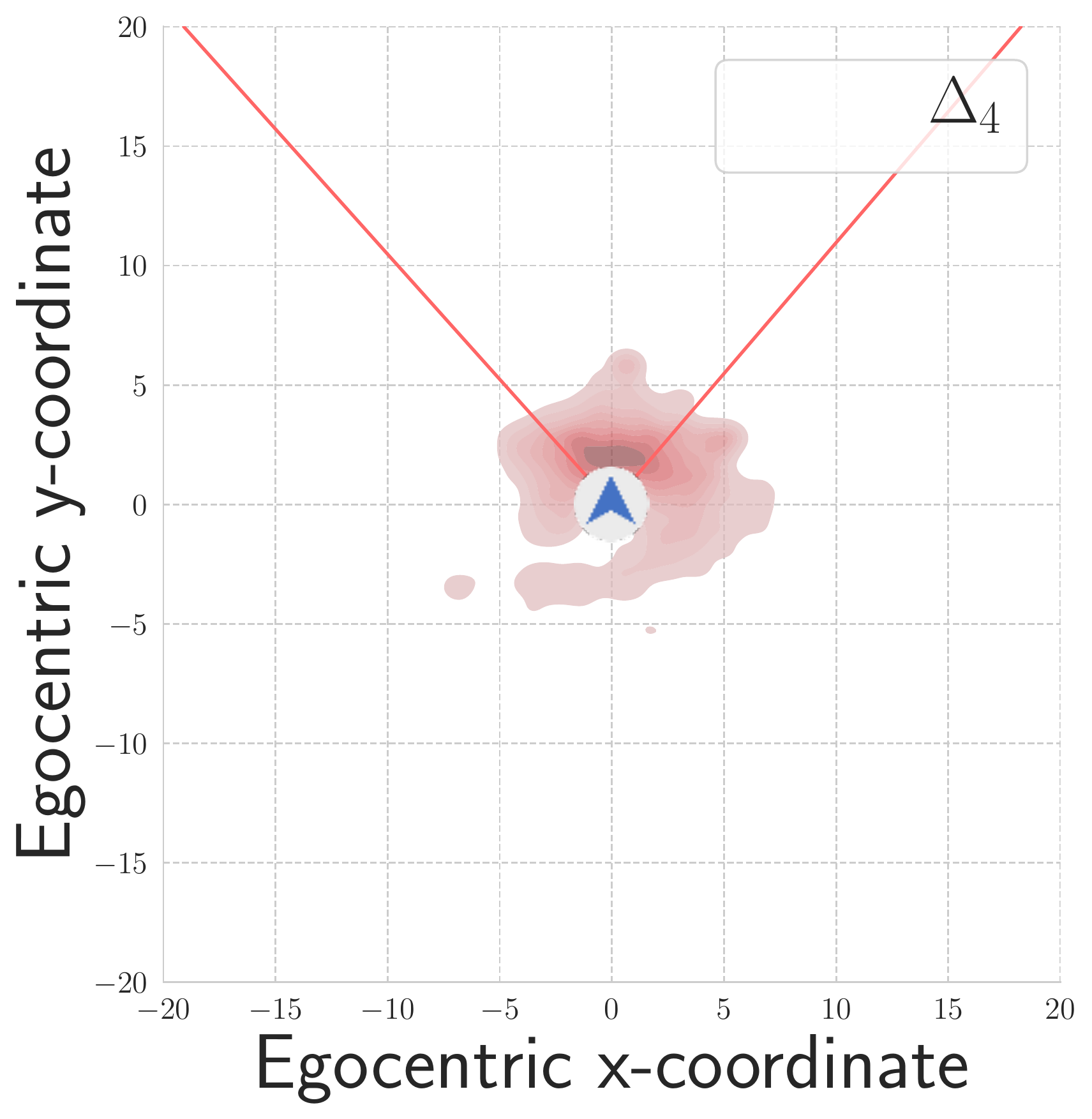} &  
    \includegraphics[width=1\linewidth]{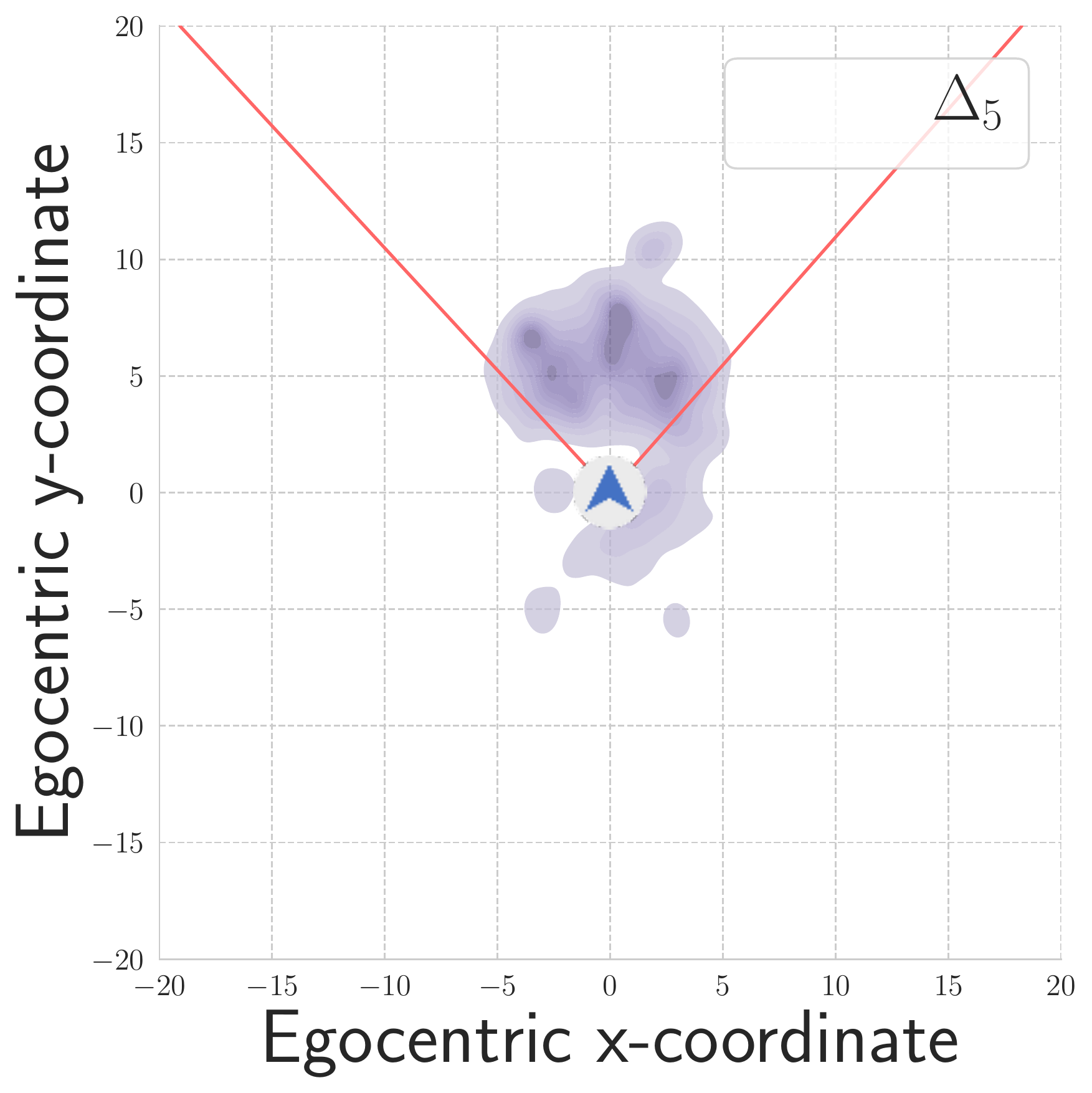} &  
    \includegraphics[width=1\linewidth]{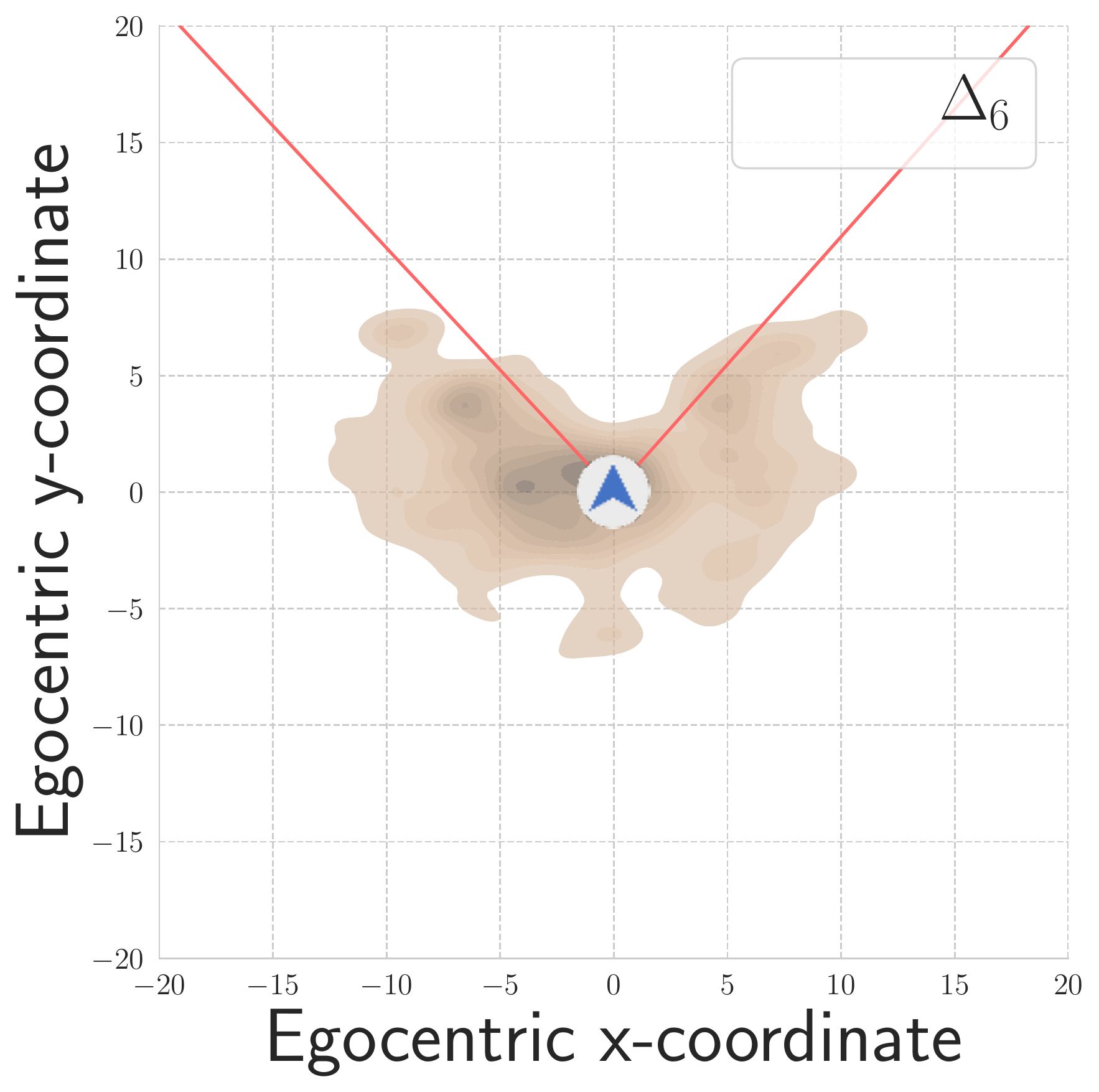}\\
    
\end{tabular}
}
\caption{
\textbf{Egocentric visualization of \DiscCom communication symbol $m_{O \rightarrow N}^2$ for vocabulary size 3.}
The plots show the relative coordinates of the current goal object from \atwo's perspective when \aone communicates the symbol through \DiscCom with vocabulary size three.
The navigator agent (\atwo) is facing the +y axis and its field-of-view is marked with red lines.
Data points are accumulated across all validation episodes, and we plot contour lines of the bivariate density distribution.
Each data point is a message with $(x,y)$ coordinates determined from the coordinates of the current goal object in \atwo's egocentric reference frame when the message was sent.
The six plots on the left are for each communication symbol, and the right-most combines all symbols.
Note how each symbol represents distinct regions that are egocentrically organized around the agent.
}
\label{fig:disc_comm_vocab3_m_O_to_N_exploded}
\end{figure*}

\begin{figure*}[b]
\resizebox{\linewidth}{!}{
\newcolumntype{C}{>{\centering\arraybackslash} m{8.8cm} }
\newcolumntype{A}{>{\centering\arraybackslash} m{17.6cm} }
\begin{tabular}{@{}CCCA@{}}
    \includegraphics[width=1\linewidth]{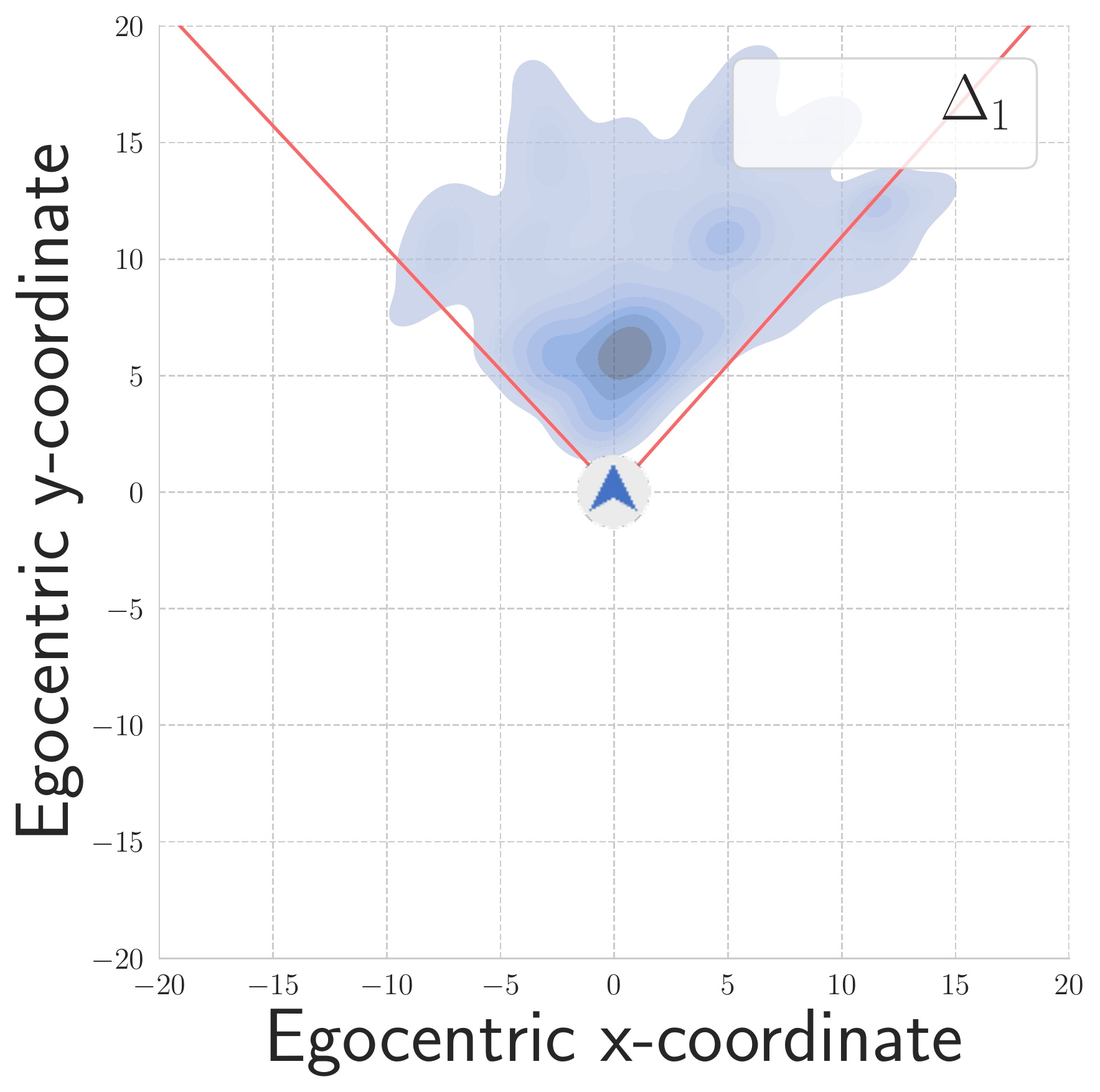} &  
    \includegraphics[width=1\linewidth]{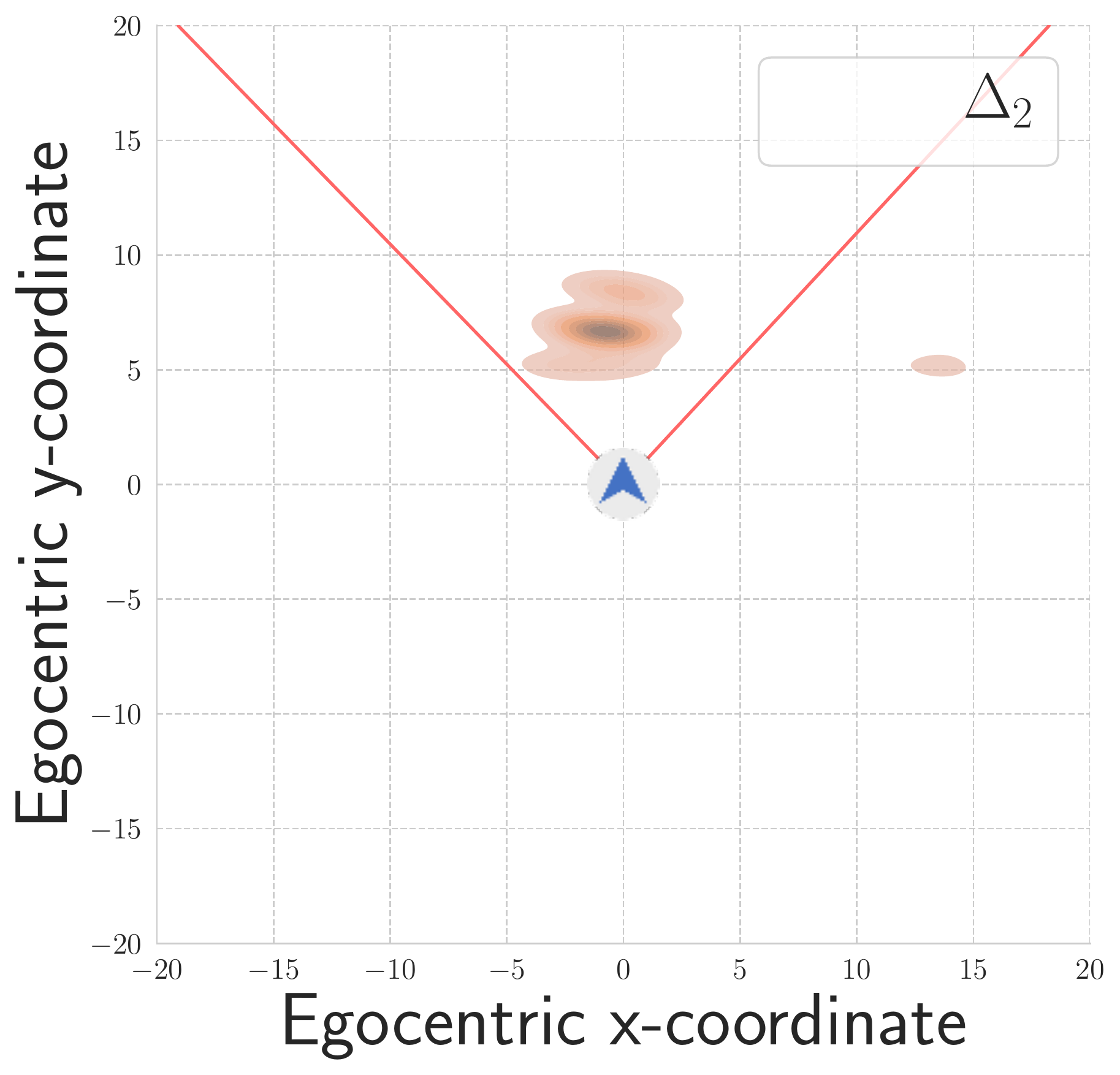} &  
    \includegraphics[width=1\linewidth]{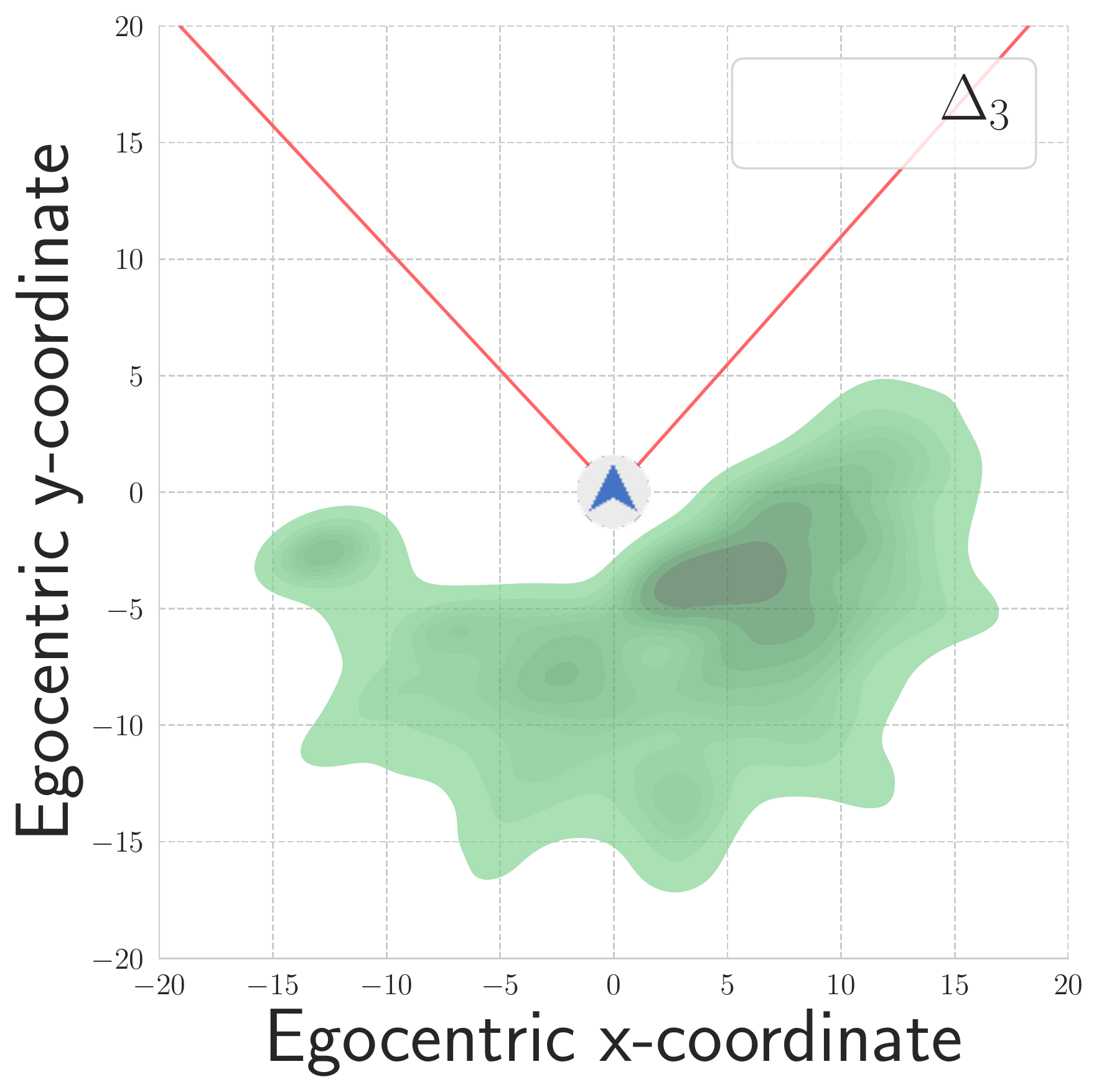} &
    \multirow{2}{*}[1.65in]{\includegraphics[width=1\linewidth]{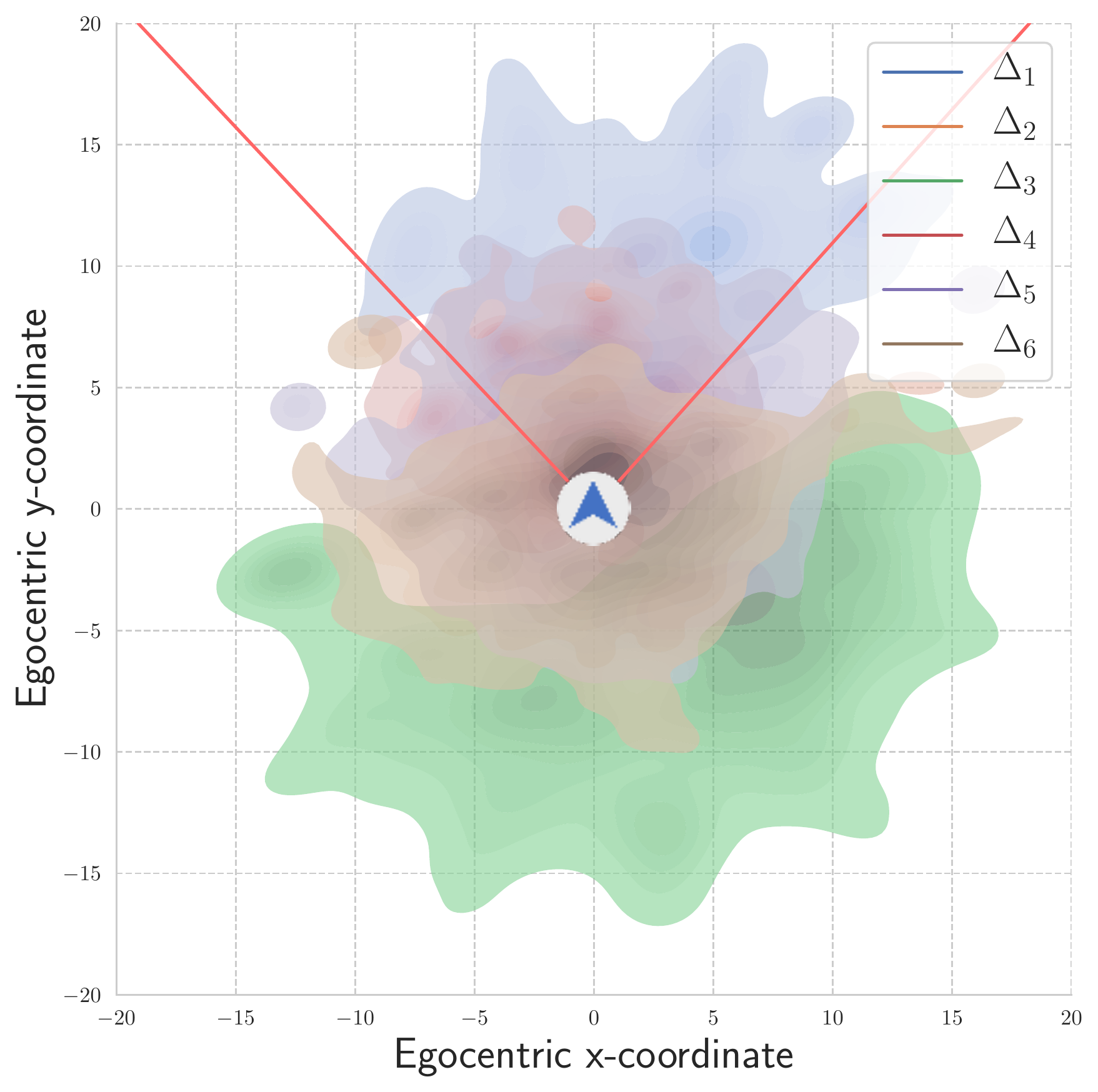}}\\
    \includegraphics[width=1\linewidth]{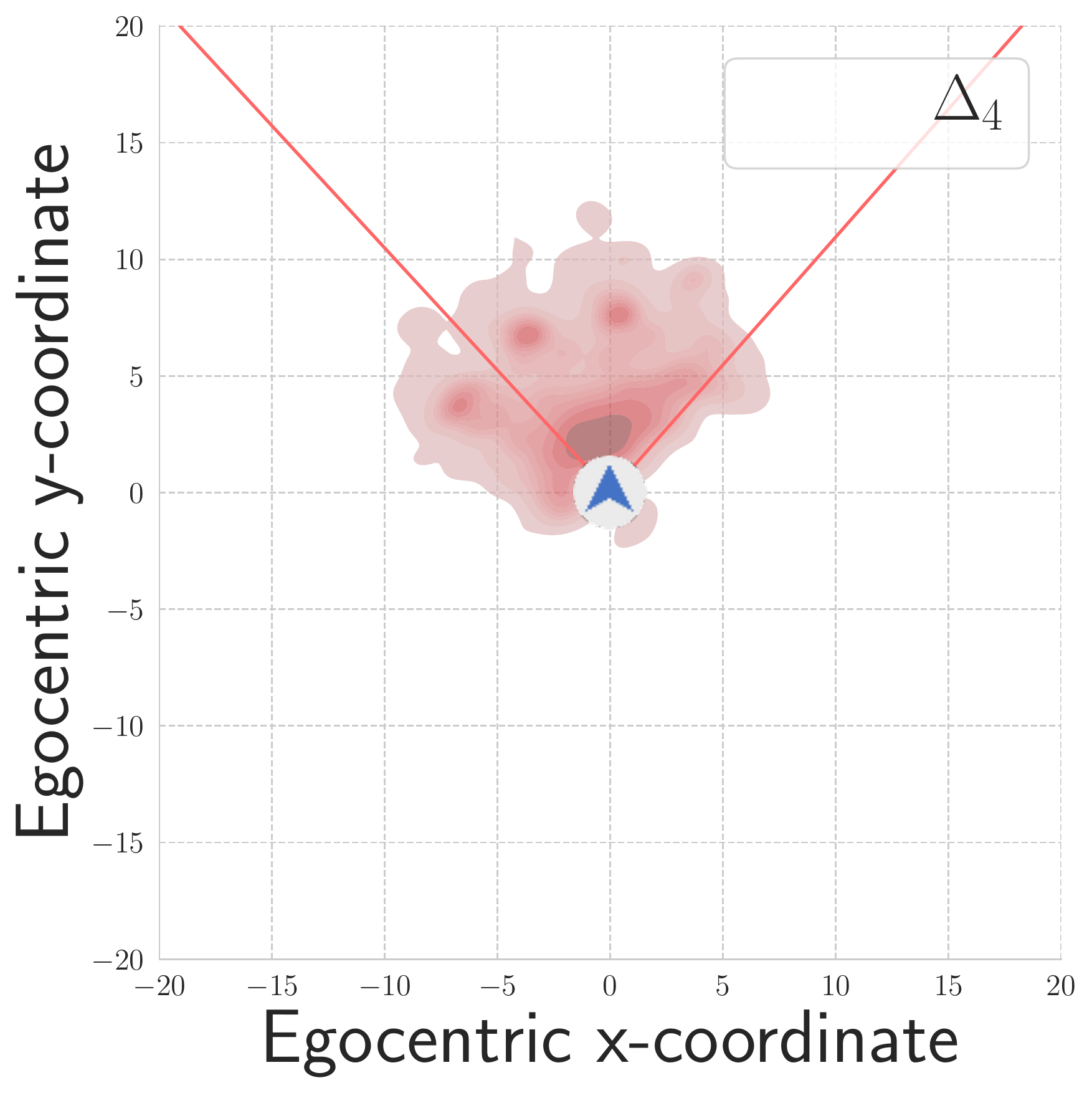} &  
    \includegraphics[width=1\linewidth]{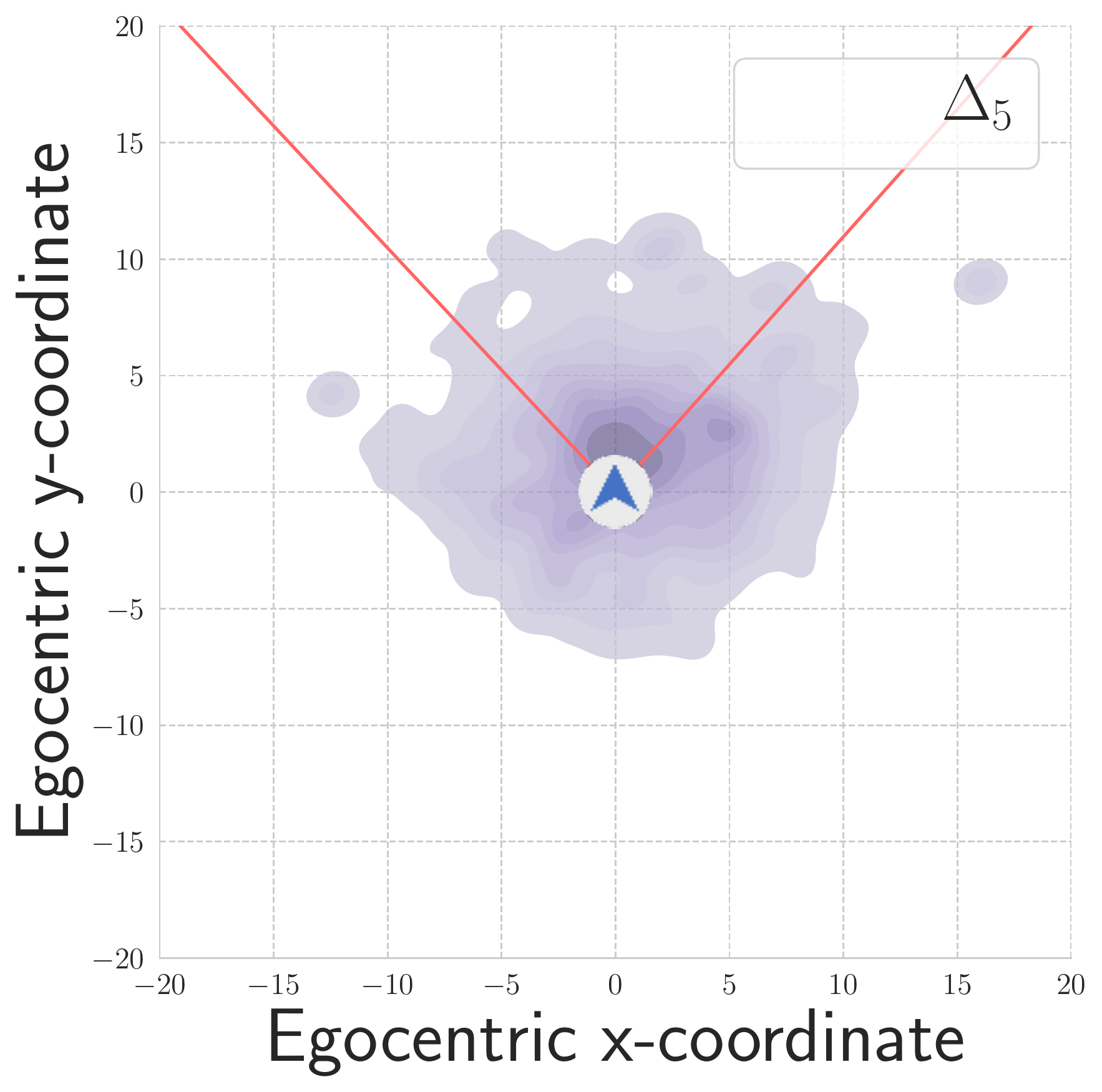} &  
    \includegraphics[width=1\linewidth]{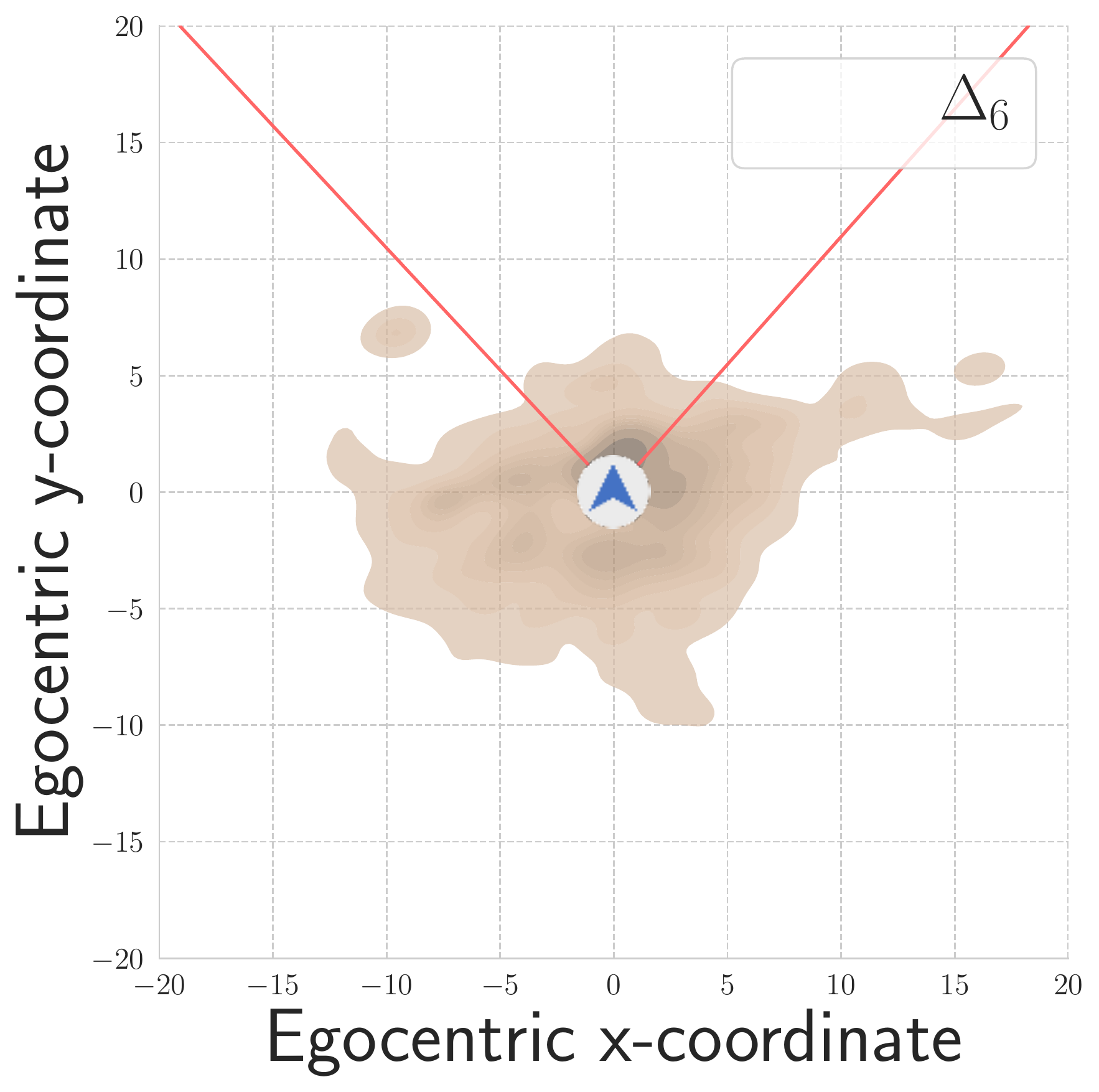}\\
    
\end{tabular}
}
\caption{
\textbf{Egocentric visualization of \DiscCom communication symbol $m_{O \rightarrow N}^1$ for vocabulary size 3.}
The plots show the relative coordinates of the current goal object from \atwo's perspective when \aone communicates the symbol through \DiscCom with vocabulary size two.
The navigator agent (\atwo) is facing the +y axis and its field-of-view is marked with red lines.
Data points are accumulated across all validation episodes, and we plot contour lines of the bivariate density distribution.
Each data point is a message with $(x,y)$ coordinates determined from the coordinates of the current goal object in \atwo's egocentric reference frame when the message was sent.
The first three plots are for each communication symbol, and the right-most combines all symbols.
}
\label{fig:disc_comm_vocab3_m_O_to_N_exploded_round0}
\end{figure*}

\subsection{Interpretation of \DiscCom for vocabulary size 3}
\label{sec:disc_comm_interpret}
We provide details of the analysis of \DiscCom with vocabulary of size 3.  Similar to our analysis for vocabulary size 2 (see section 6.2 in main paper), we bin the messages probabilities based on the observed probabilities. Due to the larger vocabulary size, we bin the messages into six classes (vs 3 classes for vocabulary size of 2): \symone, \symtwo, \symthree, \symfour, \symfive or \symsix. See \ref{sec:disc_com_details} for more details about the binning process. When we examine the messages, we see a consistent pattern as we observed for vocabulary size 2. 

\xhdr{What does \atwo tell \aone in $m_{N \rightarrow O}^1$?}
Here also, we observe that \atwo uses $m_{N \rightarrow O}^1$ to convey the goal object to \aone. \atwo send \symone when the goal object is a red, green, pink, or cyan cylinder.
It sends \symtwo for blue and yellow cylinders, and it sends \symthree otherwise.
We find that \symfour, \symfive, \symsix are not used for $m_{N \rightarrow O}^1$, and are only used in $m_{O \rightarrow N}^2$ and $m_{O \rightarrow N}^1$.

\xhdr{What does \aone tell \atwo in $m_{O \rightarrow N}^2$?}
We perform the same interpretation analysis as we did for vocabulary size 2 in the main paper.
We again observe that \aone utilizes $m_{O \rightarrow N}^2$ to convey the goal location to \atwo (see \Cref{fig:disc_comm_vocab3_m_O_to_N_exploded}).
Because of the availability of more communication symbols in vocabulary size 3, \aone send more fine-grained information about the regions.
Similar to vocabulary size 2 (main paper section 6.2), we observe that more symbols are allocated to the front of the agent than at its back. 

\xhdr{What does \aone tell \atwo in $m_{O \rightarrow N}^1$?}
For this message, our observations are again consistent with those of vocabulary size 2 (see \Cref{fig:disc_comm_vocab3_m_O_to_N_exploded_round0}).
\aone sends different symbols for different goal locations, but there is more overlap between the regions allocated to the symbols as compared to that in $m_{O \rightarrow N}^2$.

\subsection{Interpretation for \mon{2}}
\label{sec:comm_2ON}

Most of our analysis thus far has focused on \mon{1}.
Here we analyze what is communicated in \ContCom and \DiscCom for \mon{2} using the same methodology.

\subsubsection{Interpretation of \ContCom for \mon{2}}
\label{sec:u_comm_2ON}

\xhdr{What does \atwo tell \aone in $m_{N \rightarrow O}^1$?}
We observe that the distribution of $m_{N \rightarrow O}^1$ is similar to that in Figure 4 of the main paper.
This is expected as \atwo would send similar message irrespective of the number of goals in an episode.

\begin{figure*}
\resizebox{\linewidth}{!}{
\newcolumntype{C}{>{\centering\arraybackslash} m{8.8cm} }
\begin{tabular}{@{}CC@{}}
{$1^{st}$ element} & {$2^{nd}$ element} \\
    \includegraphics[width=1\linewidth]{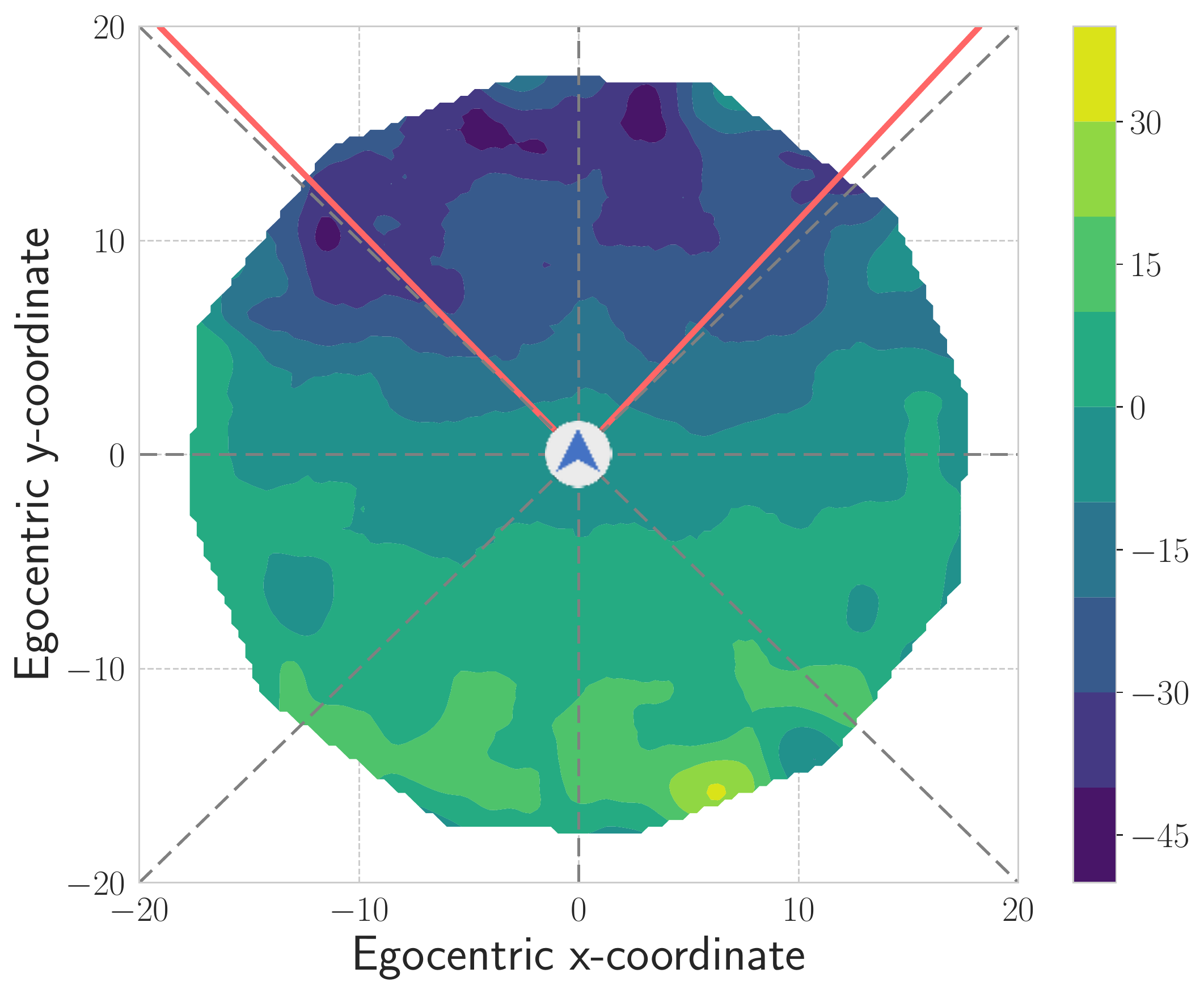}
    &
    \includegraphics[width=1\linewidth]{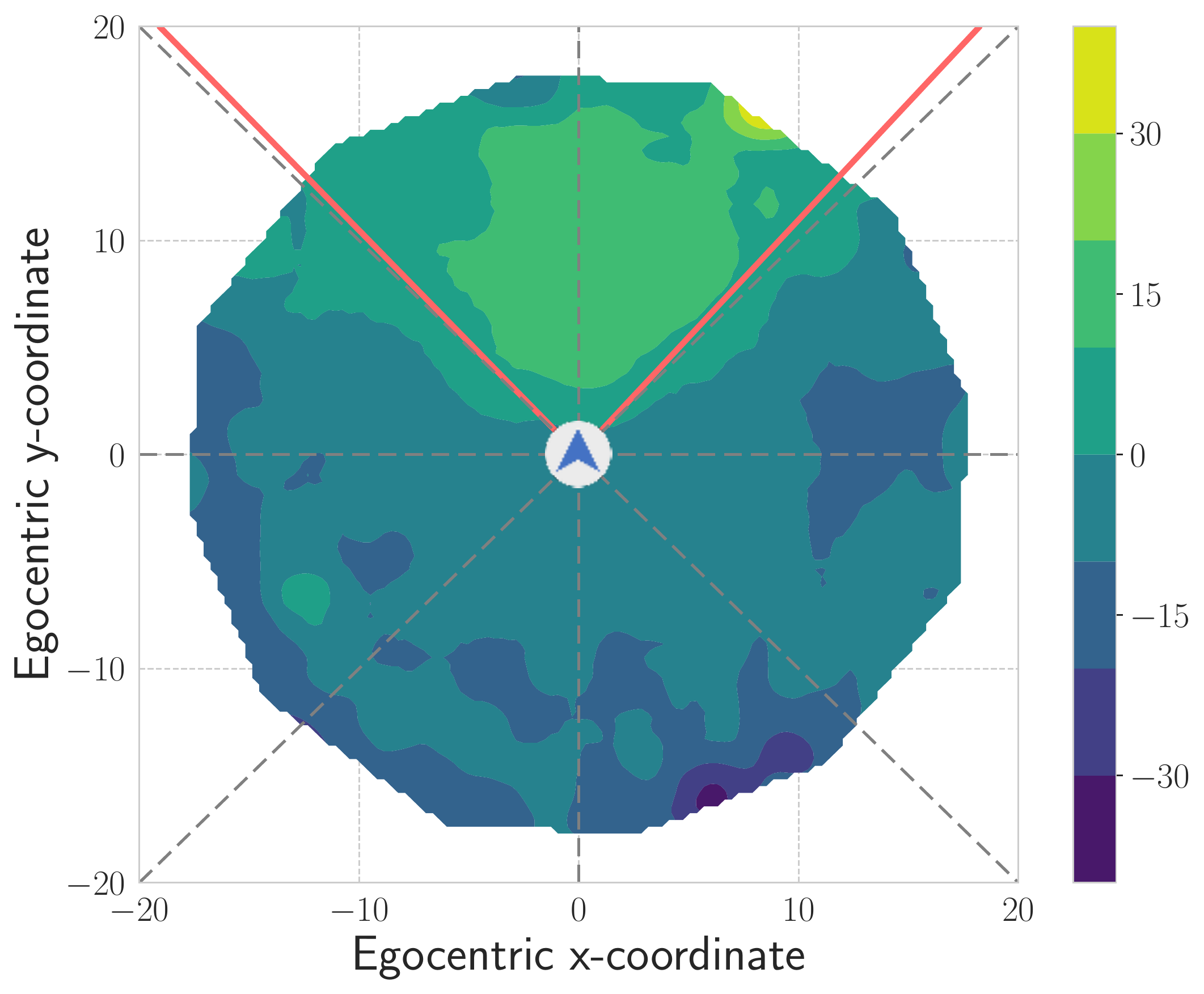}
\end{tabular}
}
\caption{
\textbf{Egocentric visualization of \ContCom communication symbol} $m_{O \rightarrow N}^2$ \textbf{for \mon{2}.}
The two plots visualize the value of the first and second element of the message plotted \wrt the relative coordinates of the goal object from \atwo.
The navigator agent \atwo is facing the +y axis and its field-of-view is marked with red lines.
The plot on the left corresponds to the $1^\text{st}$ dimension of the message, while the plot on the right corresponds to the $2^\text{nd}$ dimension.
The value of each dimension is indicated by the color hue.
}
\label{fig:cont_comm_2ONm_O_to_N_round_1}
\end{figure*}

\xhdr{What does \aone tell \atwo in $m_{O \rightarrow N}^2$?}
In \Cref{fig:cont_comm_2ONm_O_to_N_round_1}, we show show the distribution of $m_{O \rightarrow N}^2$ against the current object goal in the spatial reference frame defined by the position and orientation of \atwo (egocentric frame) at the environment step when the message was sent. Our observations are consistent with what we observed for \mon{1}. $m_{O \rightarrow N}^2$ is used to convey the goal location to \atwo.

\begin{figure*}
\resizebox{\linewidth}{!}{
\newcolumntype{C}{>{\centering\arraybackslash} m{8.8cm} }
\newcolumntype{D}{>{\centering\arraybackslash} m{1cm} }
\begin{tabular}{@{}DCCCC@{}}
    \rotatebox{90}{\LARGE{Distinguishable goals}} &
    \includegraphics[width=1\linewidth]{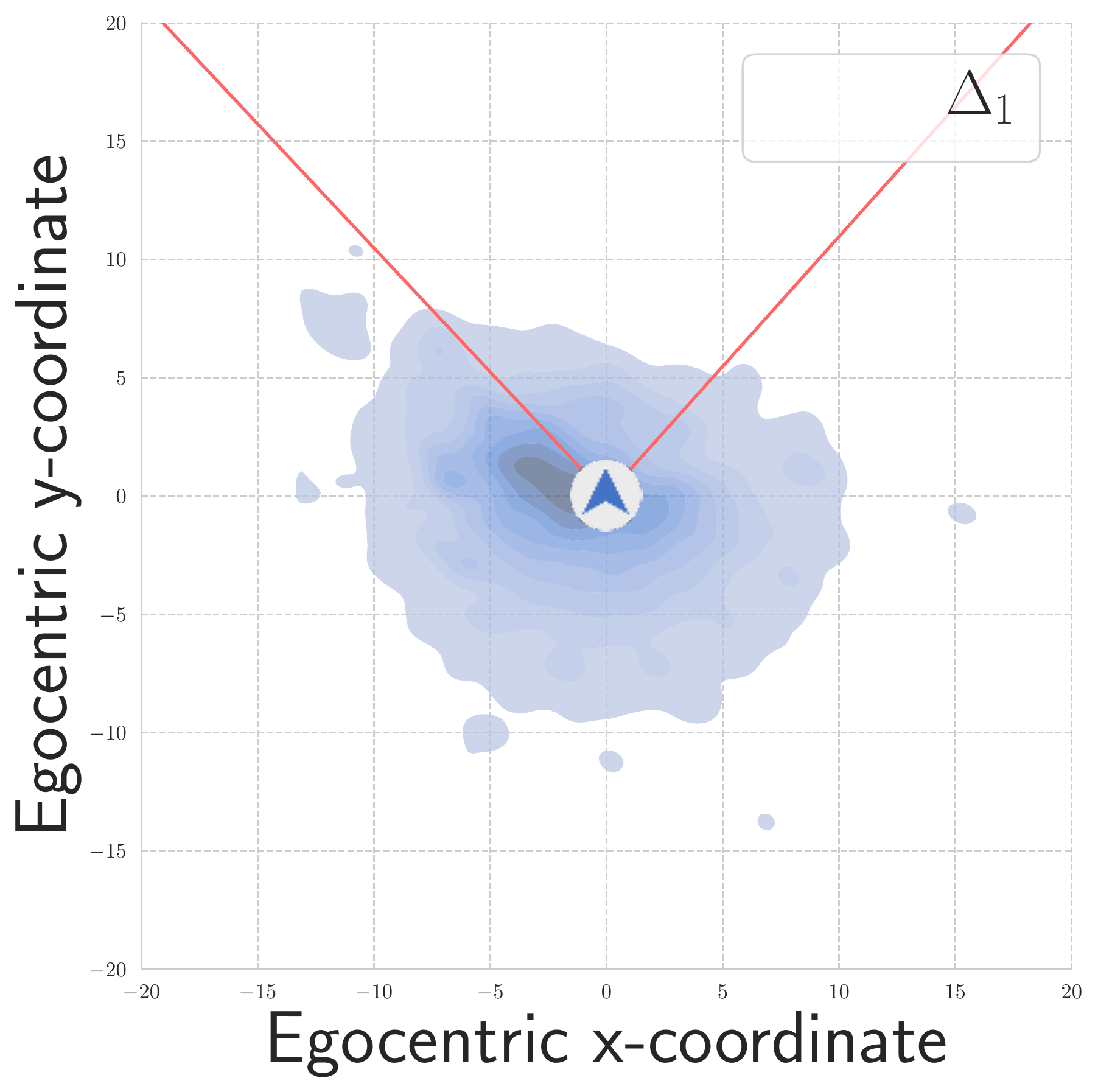} &  
    \includegraphics[width=1\linewidth]{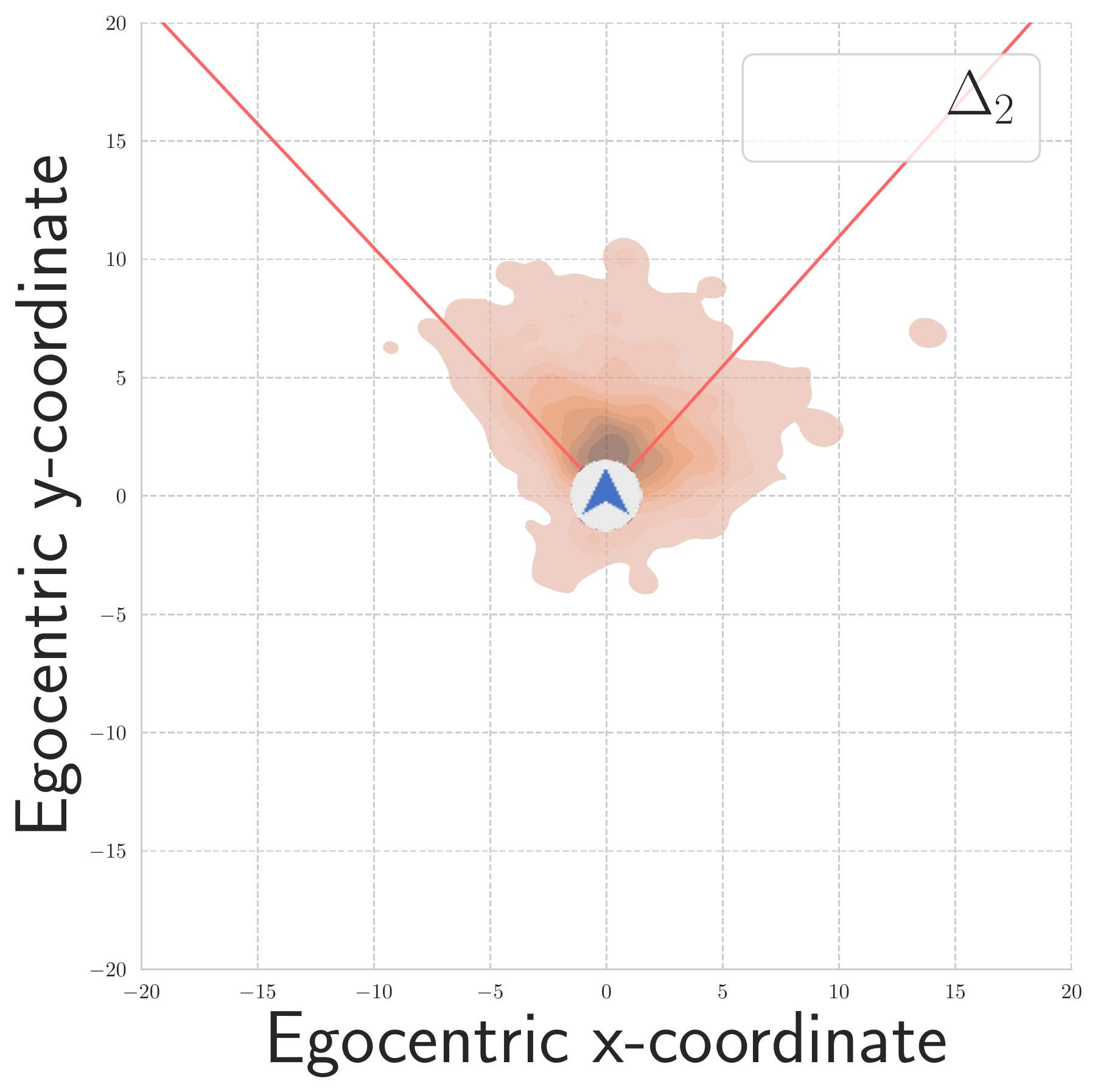} &  
    \includegraphics[width=1\linewidth]{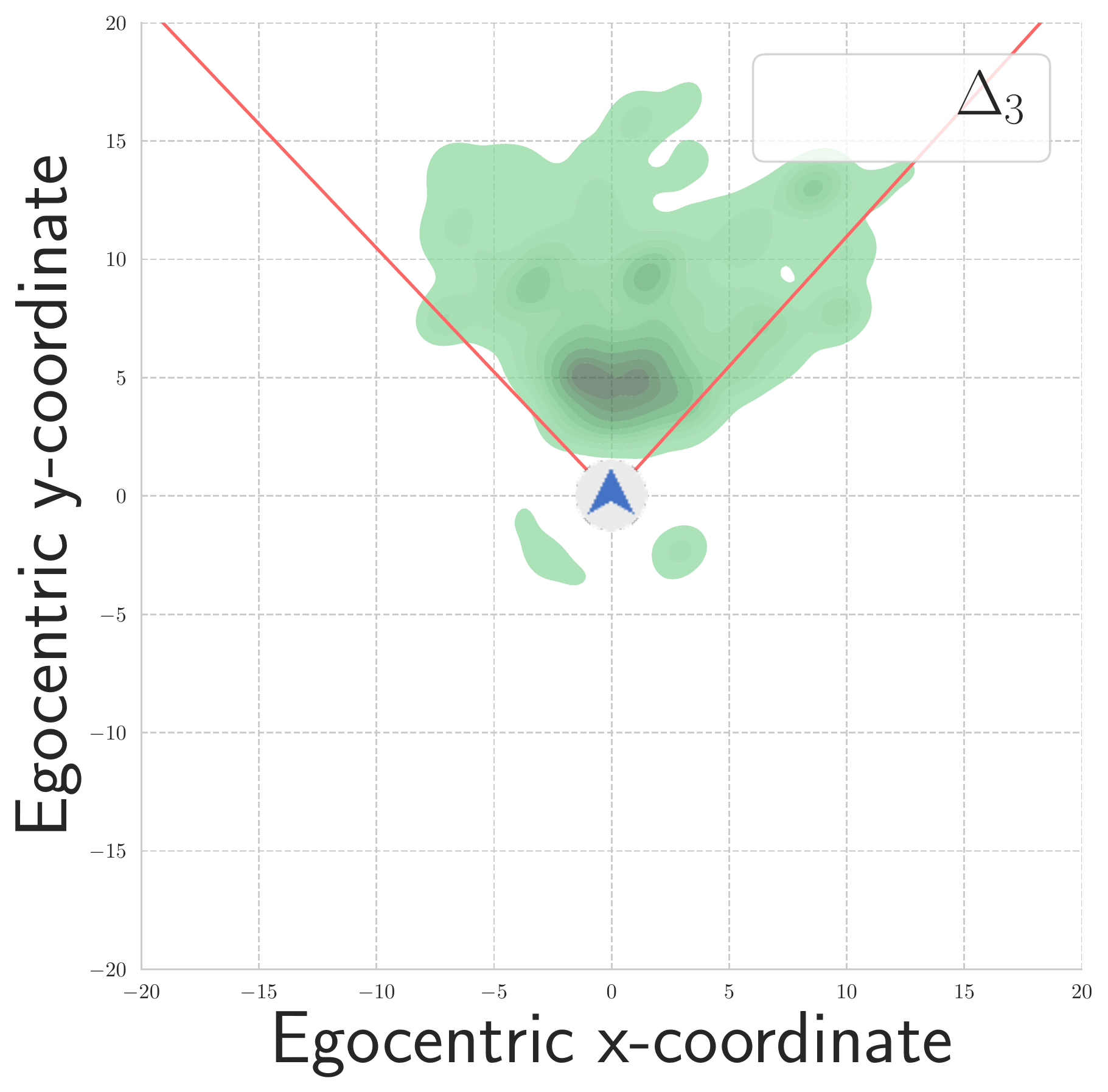} &  
    \includegraphics[width=1\linewidth]{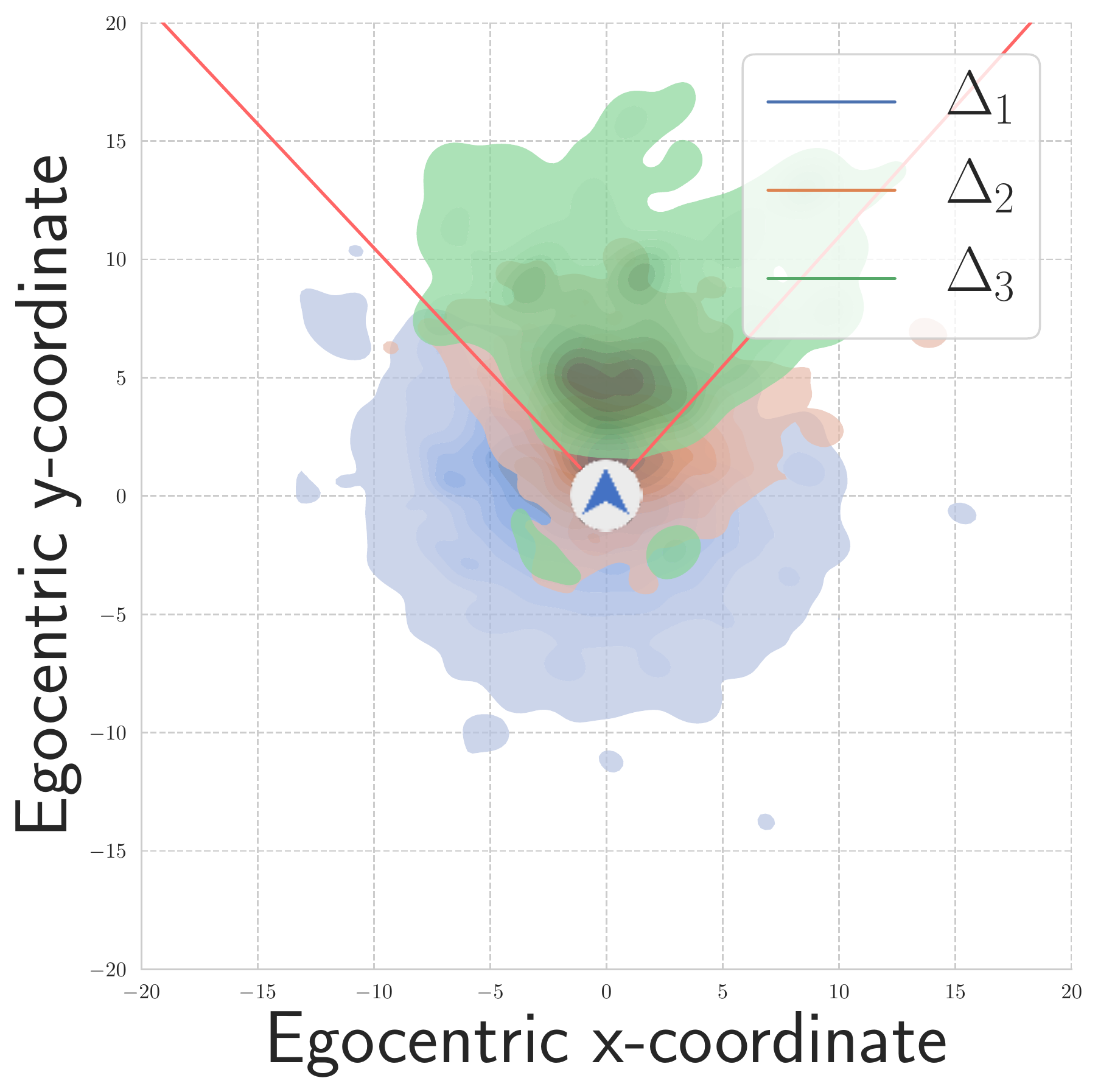}   \\
    \rotatebox{90}{\LARGE{Indistinguishable Goals}} &
    \includegraphics[width=1\linewidth]{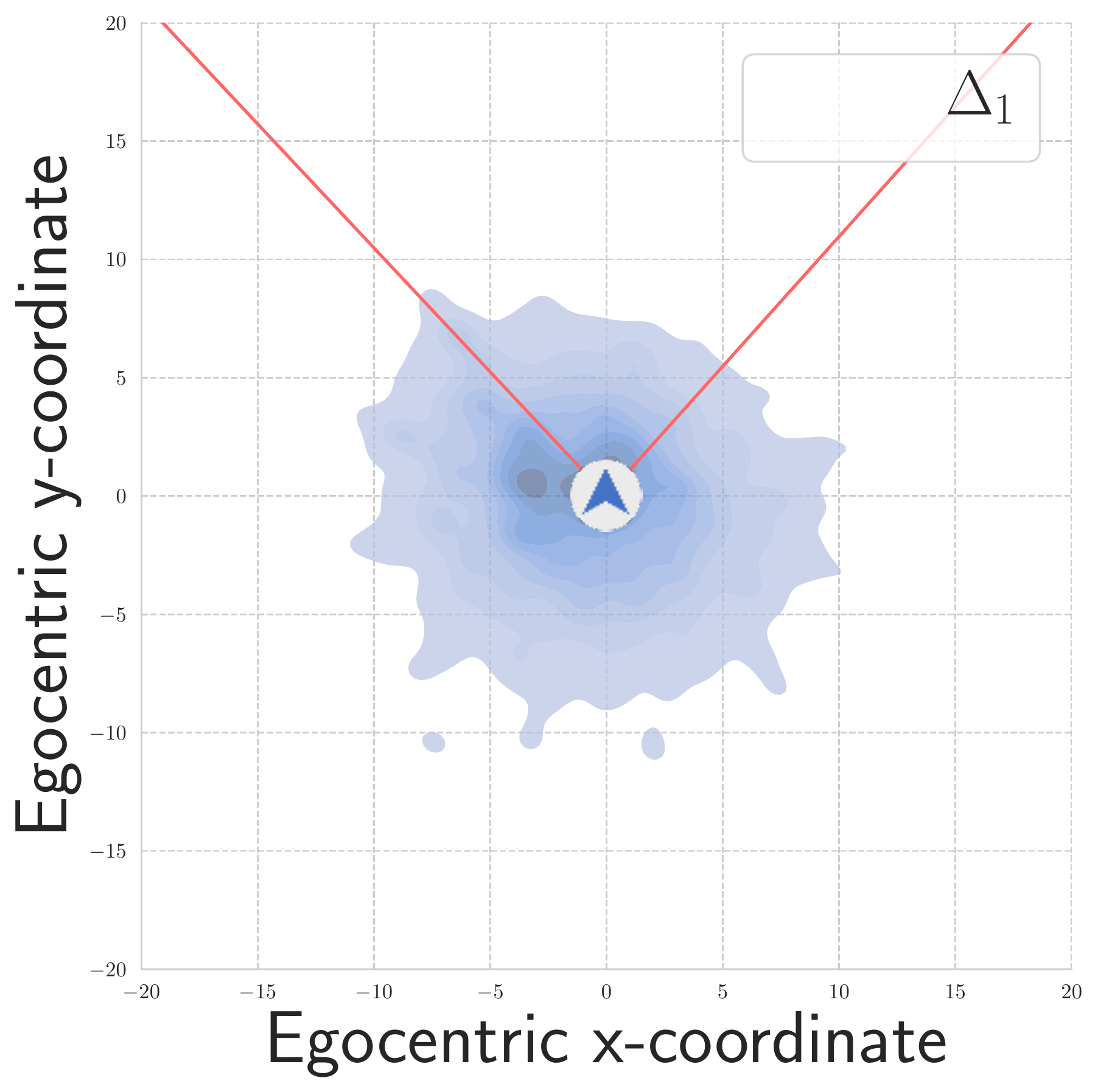} &  
    \includegraphics[width=1\linewidth]{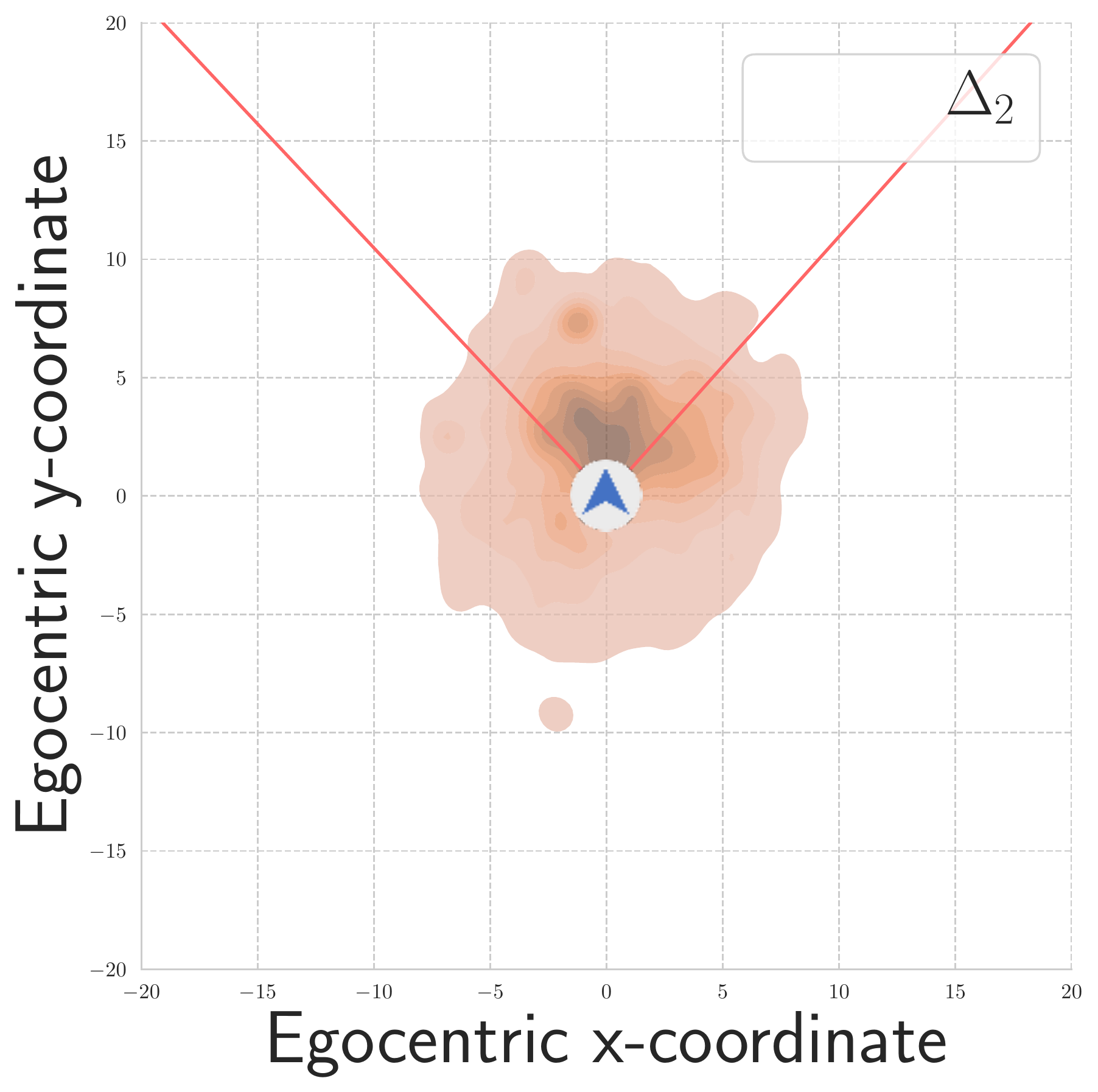} &  
    \includegraphics[width=1\linewidth]{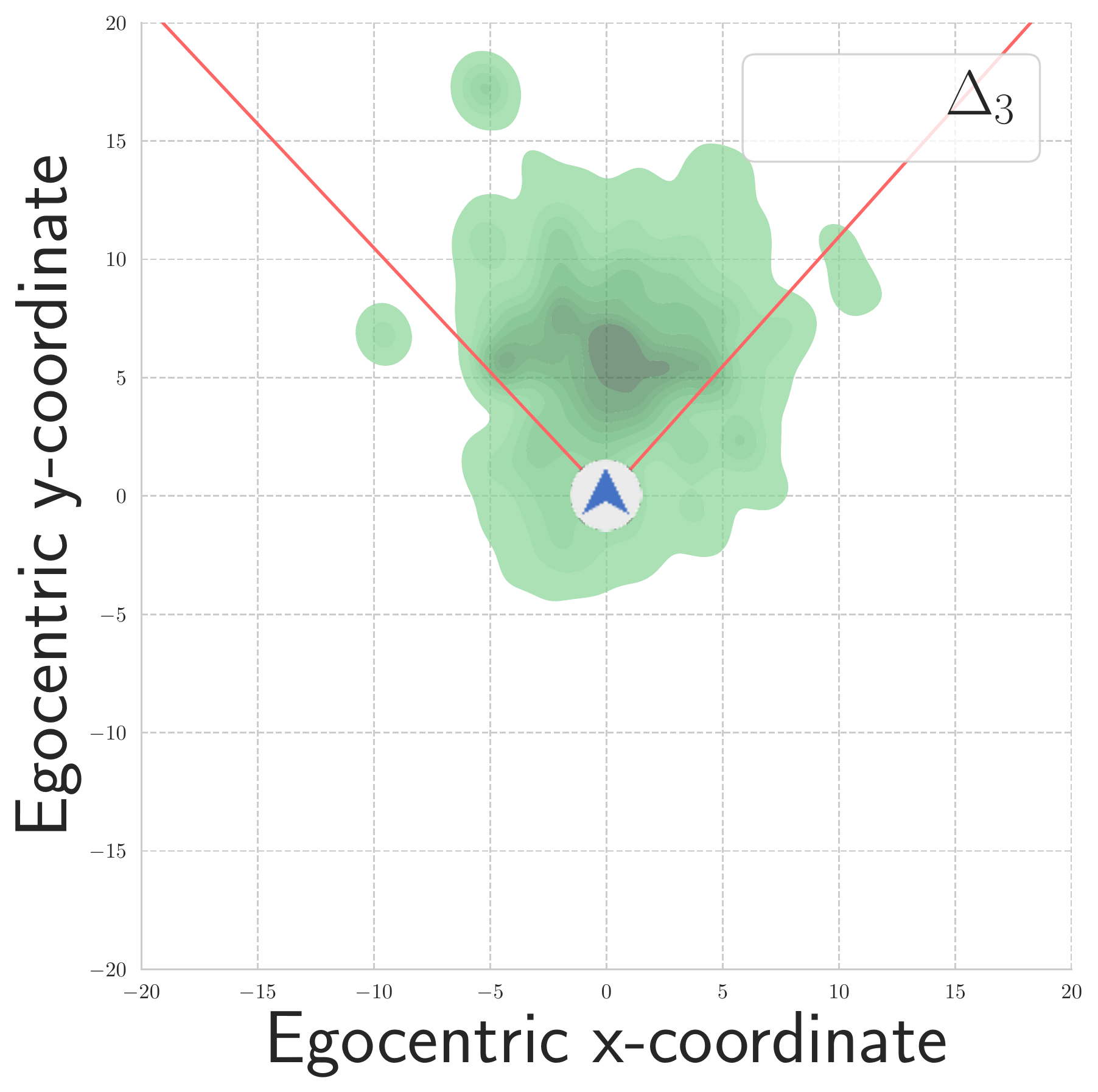} &  
    \includegraphics[width=1\linewidth]{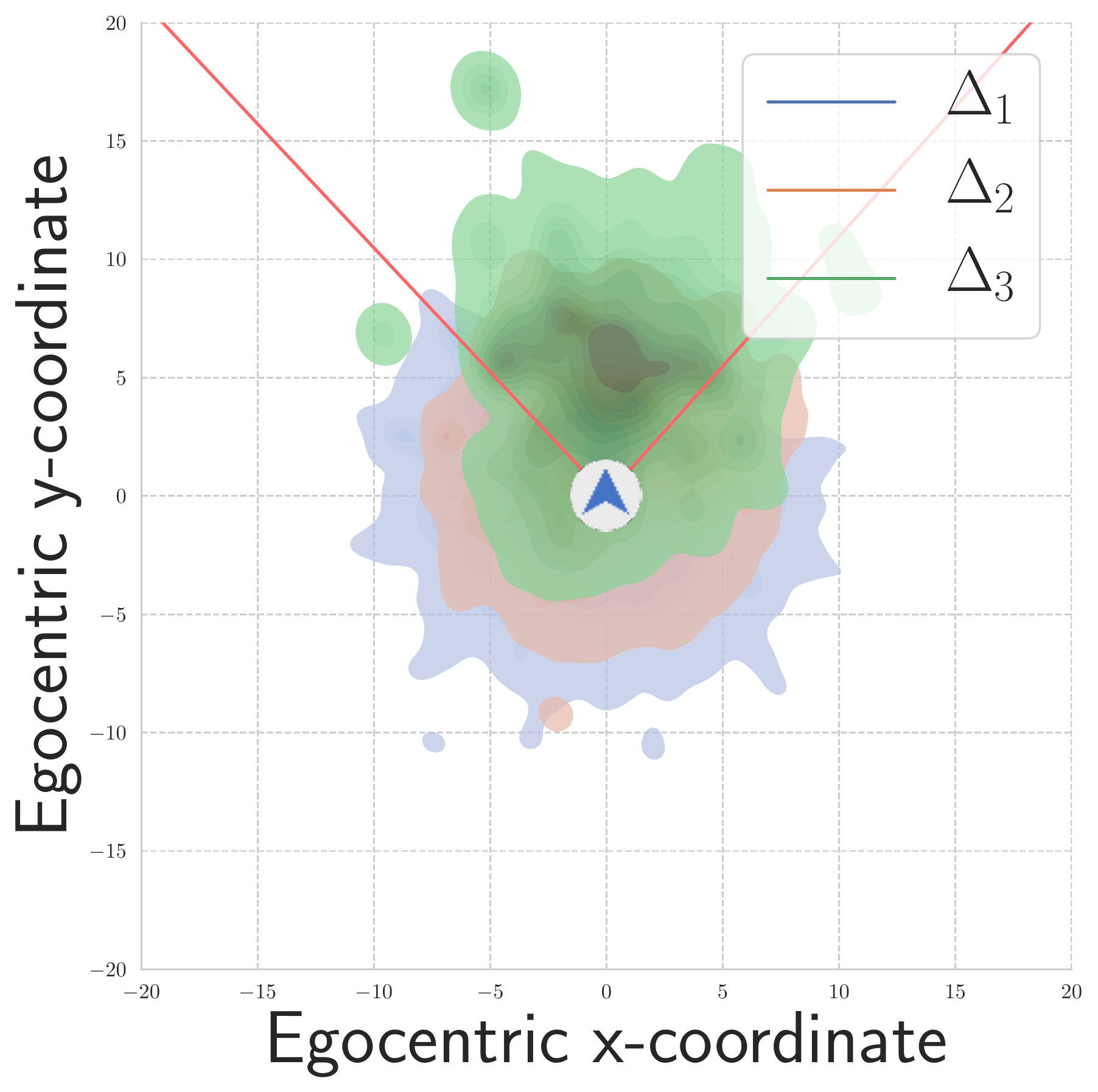}   
\end{tabular}
}
\caption{
\textbf{Egocentric visualization of \DiscCom communication symbol $m_{O \rightarrow N}^2$.} First and second row show the case when the goals are distinguishable and indistinguishable by \aone respectively.
The plots show the relative coordinates of the current goal object from \atwo's perspective when \aone communicates the symbol through \DiscCom with vocabulary size two.
The navigator agent (\atwo) is facing the +y axis and its field-of-view is marked with red lines.
Data points are accumulated across all validation episodes, and we plot contour lines of the bivariate density distribution.
Each data point is a message with $(x,y)$ coordinates determined from the coordinates of the current goal object in \atwo's egocentric reference frame when the message was sent.
The first three plots are for each communication symbol, and the right-most combines all symbols. Notice that first row symbols have lesser overlap than the second row symbols.
}
\label{fig:disc_comm_2ON_m_O_to_N}
\end{figure*}

\subsubsection{Interpretation of \DiscCom for \mon{2}}
\label{sec:s_comm_2ON}

In this setting we use a vocabulary size of 2 and group the messages into three symbols as in the main paper.
Because the number of symbols is less than the number of possible goals, we observe that the agents use a partitioning scheme when sending messages.
This phenomenon has been observed in Kottur et al. and is consistent with game theory results.

\xhdr{What does \atwo tell \aone in $m_{N \rightarrow O}^1$?}
Here also, \aone sends similar $m_{N \rightarrow O}^1$ as in \mon{1}.
That is, \aone sends \symone when the goal object is a red, white or black cylinder, and sends \symtwo otherwise.
\atwo partitions the goal objects into two sets: $P_1$ with $3$ categories and $P_2$ with $5$ categories.

\xhdr{What does \aone tell \atwo in $m_{O \rightarrow N}^2$?}
As $m_{N \rightarrow O}^1$ only sends \symone or \symtwo, \aone cannot infer the precise current goal object.
If both of these objects lie in $P_2$, \atwo would send \symtwo to \aone throughout the episode.
Therefore, \aone would not know which of the $2$ objects \atwo is looking for at the moment.
Instead, if one of the target objects lies in $P_1$ and the other in $P_2$, \aone can infer the current target object \atwo is looking for.
We plot the message $m_{O \rightarrow N}^2$ for the two cases separately.
In \Cref{fig:disc_comm_2ON_m_O_to_N}, first row represents the case when the current goal can be distinguished from $m_{N \rightarrow O}^1$.
Note that the current goal is said to be \textit{distinguishable} from $m_{N \rightarrow O}^1$ if the two goals for the episode lie in separate partitions $P_1$ and $P_2$.
$m_{O \rightarrow N}^2$ correlates more strongly with the location of the current goal in the former case, where \aone could infer the current goal before sending $m_{O \rightarrow N}^2$.
This is reflected in the symbols being more well separated in the first row of \Cref{fig:disc_comm_2ON_m_O_to_N} than in the second row.
This can be observed by the overlaps between symbols.
Distinguishable goals have less overlap between symbol regions as compared to indistinguishable goals.
To quantify the separation of symbols in the two plots, we also train a random forest classifier to predict the communication symbol given the x,y coordinates of the symbol in the plots as input.
The prediction accuracy for distinguishable goals is 84\% and for indistinguishable goals it is 76\%.

\begin{figure*}
\vspace{1cm}
\includegraphics[width=\textwidth]{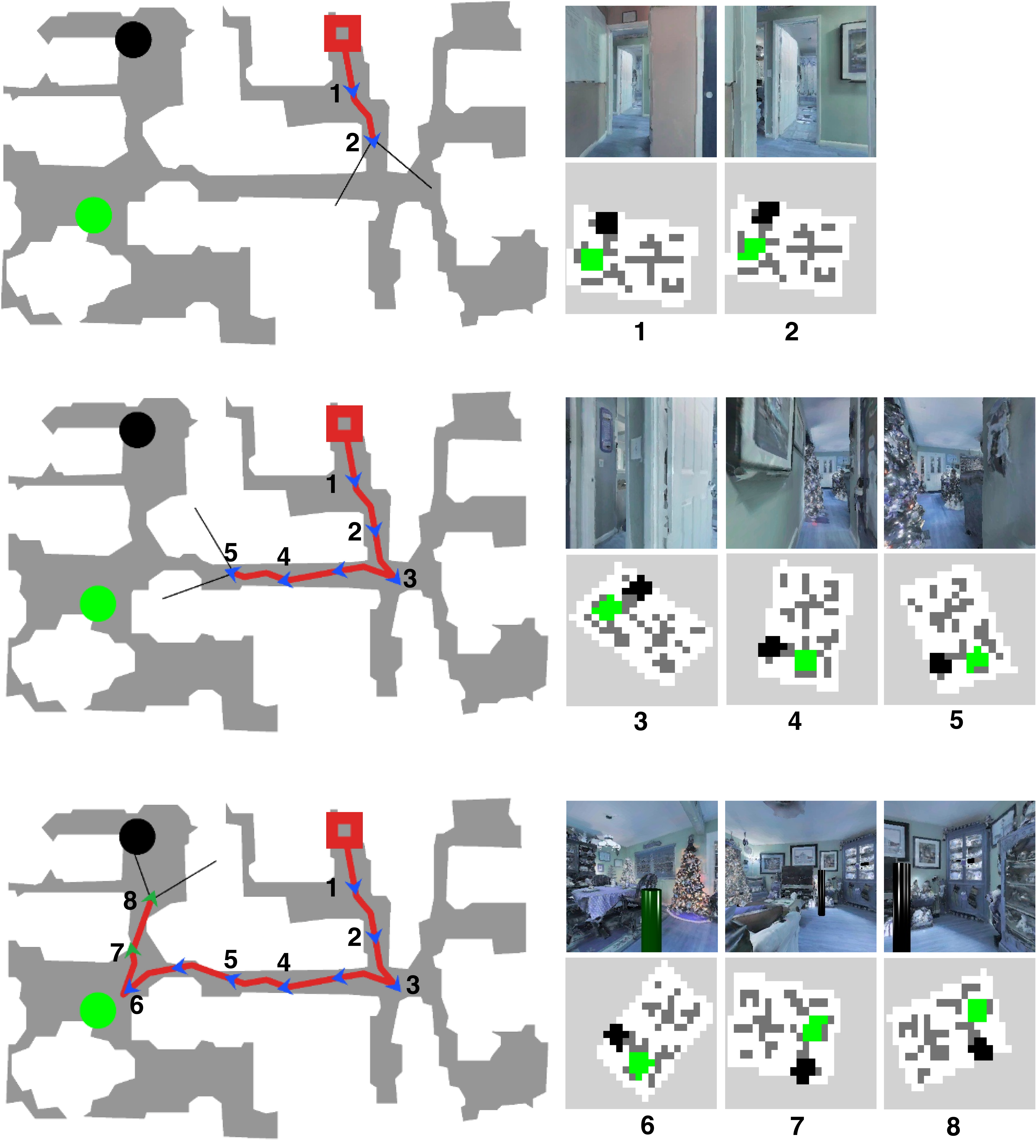}
\caption{\textbf{Example navigation episode with communication message $m_{N \rightarrow O}$ on the agent trajectory for \DiscCom.}
The message $m_{N \rightarrow O}$ is depicted by the color of the arrow symbol at various points on \atwo's trajectory on the top-down map.
The sequence of maps from top-to-bottom visualizes the trajectory at different points in time.
Egocentric observations and map representations at specific agent positions are given to the right of each map.
Note the changed communication symbol (from blue agent symbol to green) after the first green goal is found and the agent proceeds to the next black goal.
}
\label{fig:full_vis_N_to_O}
\end{figure*}
\begin{figure*}
\vspace{0.5cm}
\includegraphics[width=\textwidth]{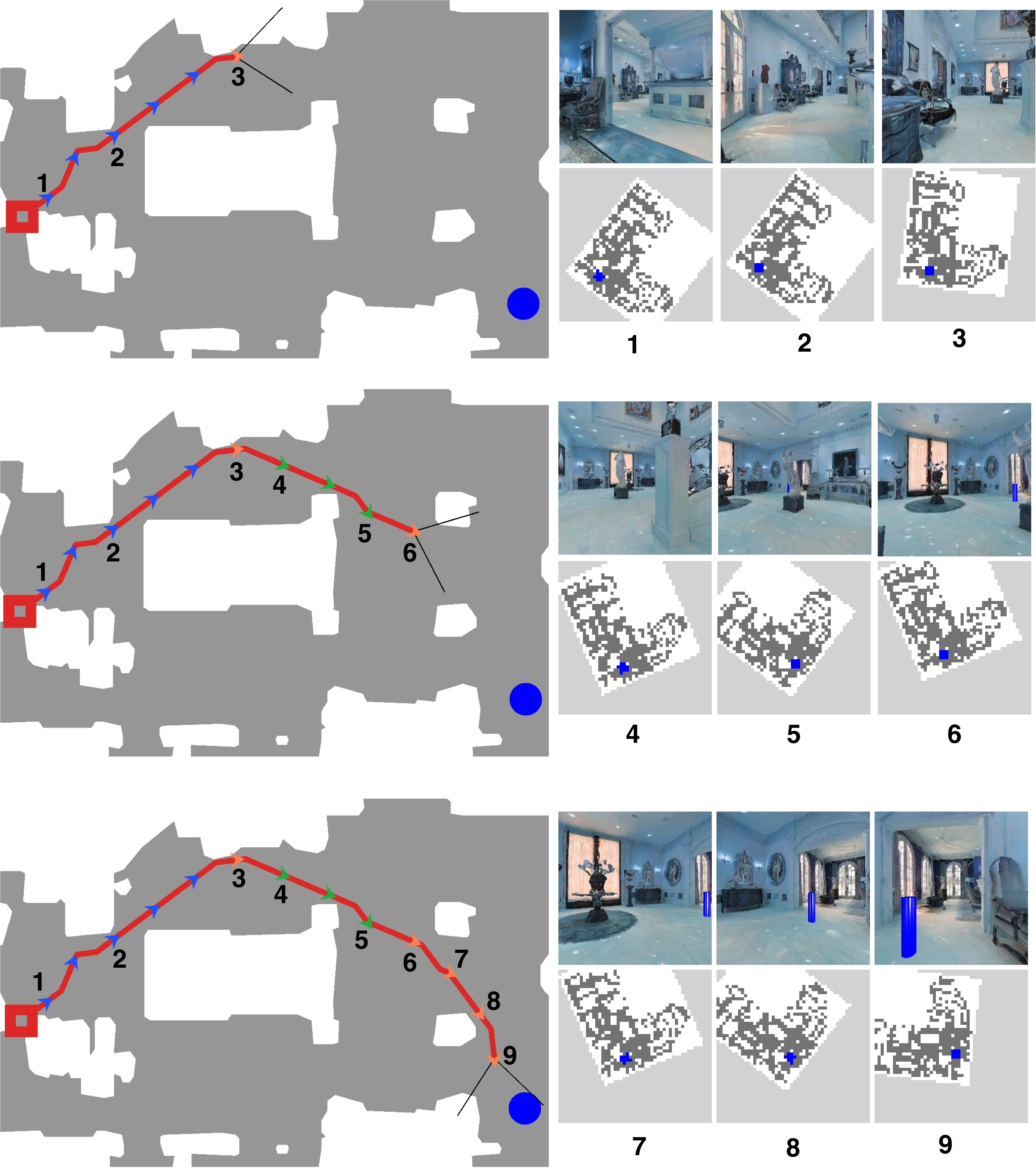}
\caption{\textbf{Example navigation episode with communication message $m_{O \rightarrow N}$ on the agent trajectory for \DiscCom.}
The message $m_{O \rightarrow N}$ is depicted by the color of the arrow symbol at various points on \atwo's trajectory on the top-down map.
The sequence of maps from top-to-bottom visualizes the trajectory at different points in time.
Egocentric observations and map representations at specific agent positions are given to the right of each map.
Note how the communication symbol changes as the relative location of the goal object with respect to the agent changes: when the goal is not ahead of the agent, it is blue; when the goal is ahead of the agent but far away, it is green; and when the goal is in front of the agent, it is orange.  
}
\label{fig:full_vis_O_to_N}
\end{figure*}

\subsection{Episode map visualizations}
\label{sec:episode_viz}

In \Cref{fig:full_vis_N_to_O}, we provide a visualization of egocentric observations and map state for \DiscCom at several points on the trajectory to show correlations between the communication symbols for $m_{N \rightarrow O}$ (shown on the trajectory) and what the agent observes at each position. Similarly in \Cref{fig:full_vis_O_to_N}, we show the correlations  for $m_{O \rightarrow N}$.

\subsection{Are messages conveying other information?}
\label{sec:additional_info}

We investigated other information that the messages might be conveying, but did not find a strong signal.   
We checked if $m_{O \rightarrow N}^1$ or $m_{O \rightarrow N}^2$ conveys the optimal action and if $m_{N \rightarrow O}^1$ conveys whether the current goal is in \atwo's view.
We also checked whether messages from \atwo to \aone contain direction, and messages from \aone to \atwo contain color, and did not find any correlations.

\section{Implementation details}
\label{sec:impl_details}

\subsubsection{Architecture details}
\label{sec:arch_details}
Here we report the architectural details. \aone has an oracle map $M$ of spatial dimension $300 \times 300$.
This contains occupancy and goal object information. $M$ is converted to egocentric map $E$ of spatial dimension $45 \times 45$.
Each of the occupancy and goal object information is converted to $16$ dimensional embeddings for each grid location so the map is of dimension $45 \times 45 \times 32$.
This is passed through a map encoder comprising of a two layered CNN and a linear layer to obtain $256-$dimensional belief $\hat{b}_O$. $b_O$ is a $256-$dimensional vector as well.

RGBD observations of \aone are passed through an image encoder.
It consists of three CNN layers and a linear layer to obtain an image embedding $v_o$ of shape $512$.
The current goal embedding $v_g$ and previous action embedding $v_a$ are both $16-$dimensional vectors.
The belief $\hat{b}_N$ and $b_O$ are of shape $512$.
The state representation vector $s$ is of shape $528$.

\subsubsection{Details about random baselines}
\label{sec:random_baselines}

Here, we present the implementation details for \RandContCom and \RandDiscCom.
In \RandContCom, we replace the message by a random vector sampled from a multi-variate gaussian distribution with mean and variance equal to the mean and variance of the corresponding message sent in the validation set. For \RandDiscCom, we replace the message by random probabilities sampled from a random multinomial probability vectors and these probabilities sum up to $1$.

\subsubsection{\textbf{\DiscCom} classifier implementation}
\label{sec:classifiers}
To establish the existence of various correlations between the communication symbols exchanged between the agents in \DiscCom, we train random forest classifiers that predict the communication symbol given a quantity $Q$ as input.
We report the classification accuracy for $m_{N \rightarrow O}^1$ and $m_{O \rightarrow N}^2$ in Section 6.2 and for $m_{O \rightarrow N}^1$ above.
For all of these, the data for training/evaluating the classifier is obtained by evaluating the model on the validation set of 1,000 episodes and accumulating the relevant metrics at each environment step across the 1,000 episodes.
At each environment step, we log the following: \{$m_{N \rightarrow O}^1$, $m_{O \rightarrow N}^1$, $m_{O \rightarrow N}^2$, current object goal category, relative location of current goal in \atwo's egocentric reference frame.\} We first balance the dataset such that each symbol $\Delta_i$ has equal number of training examples.
The collected data, where each data point corresponds to an environment step, is divided into train and val sets in 3/1 ratio.
The classifier is then trained to predict the communication symbol $\Delta_i$ from quantity $q$ using the train set.
We report the classification accuracy on the val set.

\subsubsection{Binning of probabilities in \textbf{\DiscCom}}
\label{sec:disc_com_details}

Here, we describe the implementation of binning to create the discrete symbols used in our interpretation of \DiscCom.

\xhdr{Vocabulary size 2.}
Let the probability vector output by the final softmax layer of communication module be $\boldsymbol{p} = [p_1, p_2]$ and let the binned vector be $\boldsymbol{d}$.
If $p_1 < 0.2$, $\boldsymbol{d}=[0,1]$; if $p_1>0.8$, $\boldsymbol{d}= [1,0]$; and if $0.2 \le p_1 \le 0.8 $, $\boldsymbol{d} = [0.5,0.5]$.
As such, each agent sends one of the three categorical vectors during each round of communication.

\xhdr{Vocabulary size 3.} 
Here, the model outputs a probability vector $\boldsymbol{p}$ of length 3: $[p_1, p_2, p_3]$.
The procedure for obtaining the binned vector $\boldsymbol{d}$ is described below:
\begin{equation*}
\boldsymbol{d} =
\begin{cases*}
  [1,0,0] & if $p_1 > 0.75$ \\
  [0,1,0] & if $p_2 > 0.75$ \\
  [0,0,1] & if $p_3 > 0.75$ \\
  [0,0.5,0.5] & if $\max(p_1, p_2, p_3) < 0.75$ \\ & and $p_1 < p_2,p_3$ \\
  [0.5,0,0.5] & if $\max(p_1, p_2, p_3) < 0.75$ \\ & and $p_2 < p_1,p_3$ \\
  [0.5,0.5,0] & if $\max(p_1, p_2, p_3) < 0.75$ \\ & and $p_3 < p_2,p_1$ \\
\end{cases*}
\end{equation*}
Under this formulation, each agent can be considered to send only a discrete symbol to the other agent during communication.

\end{document}